\documentclass[10pt,journal,compsoc]{IEEEtran}

\usepackage{color}
\usepackage[pagebackref=true,breaklinks=true,letterpaper=true,colorlinks,bookmarks=false]{hyperref}
\usepackage{multirow}
\usepackage{array}
\usepackage{amsmath}
\usepackage[flushleft]{threeparttable}
\usepackage{tabularx} 
\usepackage{soul}

\ifCLASSOPTIONcompsoc
  \usepackage[nocompress]{cite}
\else
  \usepackage{cite}
\fi
\ifCLASSINFOpdf
  \usepackage[pdftex]{graphicx}
  \graphicspath{{images/}}
  \DeclareGraphicsExtensions{.pdf,.jpeg,.jpg}
  \usepackage{epstopdf}
\else
\fi
\usepackage{rotating}
\usepackage{amsmath}
\usepackage{algorithmic}
\ifCLASSOPTIONcompsoc
  \usepackage[caption=false,font=footnotesize,labelfont=sf,textfont=sf]{subfig}
\else
  \usepackage[caption=false,font=footnotesize]{subfig}
\fi
\usepackage{stackengine}

\usepackage{stfloats}
\usepackage{url}

\usepackage{float}

\makeatletter
\newcommand{\thickhline}{%
    \noalign {\ifnum 0=`}\fi \hrule height 1pt
    \futurelet \reserved@a \@xhline
}
\newcolumntype{?}{!{\vrule width 1.2pt}}

\begin{document}
%
\title{AffectNet: A Database for Facial Expression, Valence, and Arousal Computing in the Wild}
%
%
%
%

\author{Ali~Mollahosseini,~\IEEEmembership{Student Member,~IEEE,}
        Behzad~Hasani,~\IEEEmembership{Student Member,~IEEE,}
        and~Mohammad~H.~Mahoor,~\IEEEmembership{Senior Member,~IEEE}
\IEEEcompsocitemizethanks{\IEEEcompsocthanksitem Authors are with the Department
of Electrical and Computer Engineering, University of Denver, Denver,
CO, 80210.\protect\\
E-mail: {amollah, bhasani, mmahoor}@du.edu
}
}

\markboth{IEEE Transactions on Affective Computing}%
{Mollahosseini \MakeLowercase{\textit{et al.}}: AffectNet: A Database for Facial Expression, Valence, and Arousal Computing in the Wild}
%



\IEEEtitleabstractindextext{%
\begin{abstract}
Automated affective computing in the wild setting is a challenging problem in computer vision. Existing annotated databases of facial expressions in the wild are small and mostly cover discrete emotions (aka the categorical model). There are very limited annotated facial databases for affective computing in the continuous dimensional model (e.g., valence and arousal). To meet this need, we collected, annotated, and prepared for public distribution a new database of facial emotions in the wild (called \textit{AffectNet}). \textit{AffectNet} contains more than 1,000,000 facial images from the Internet by querying three major search engines using 1250 emotion related keywords in six different languages. About half of the retrieved images were manually annotated for the presence of seven discrete facial expressions and the intensity of valence and arousal. \textit{AffectNet} is by far the largest database of facial expression, valence, and arousal in the wild enabling research in automated facial expression recognition in two different emotion models. Two baseline deep neural networks are used to classify images in the categorical model and predict the intensity of valence and arousal. Various evaluation metrics show that our deep neural network baselines can perform better than conventional machine learning methods and off-the-shelf facial expression recognition systems.
\end{abstract}

}

\maketitle

\IEEEdisplaynontitleabstractindextext

%
\IEEEpeerreviewmaketitle

\IEEEraisesectionheading{\section{Introduction}\label{sec:introduction}}

\IEEEPARstart{A}{ffect} is a psychological term used to describe the outward expression of emotion and feelings. Affective computing seeks to develop systems and devices that can recognize, interpret, and simulate human affects through various channels such as face, voice, and biological signals~\cite{tao2005affective}. Face and facial expressions are undoubtedly one of the most important nonverbal channels used by the human being to convey internal emotion. 

There have been tremendous efforts to develop reliable automated Facial Expression Recognition (FER) systems for use in affect-aware machines and devices. Such systems can understand human emotion and interact with users more naturally. However, current systems have yet to reach the full emotional and social capabilities necessary for building rich and robust Human Machine Interaction (HMI). This is mainly due to the fact that HMI systems need to interact with humans in an uncontrolled environment (aka wild setting) where the scene lighting, camera view, image resolution, background, user’s head pose, gender, and ethnicity can vary significantly. More importantly, the data that drives the development of affective computing systems and particularly FER systems lack sufficient variations and annotated samples that can be used in building such systems.

\begin{figure}
	\centering
	\includegraphics[width=3.5in]{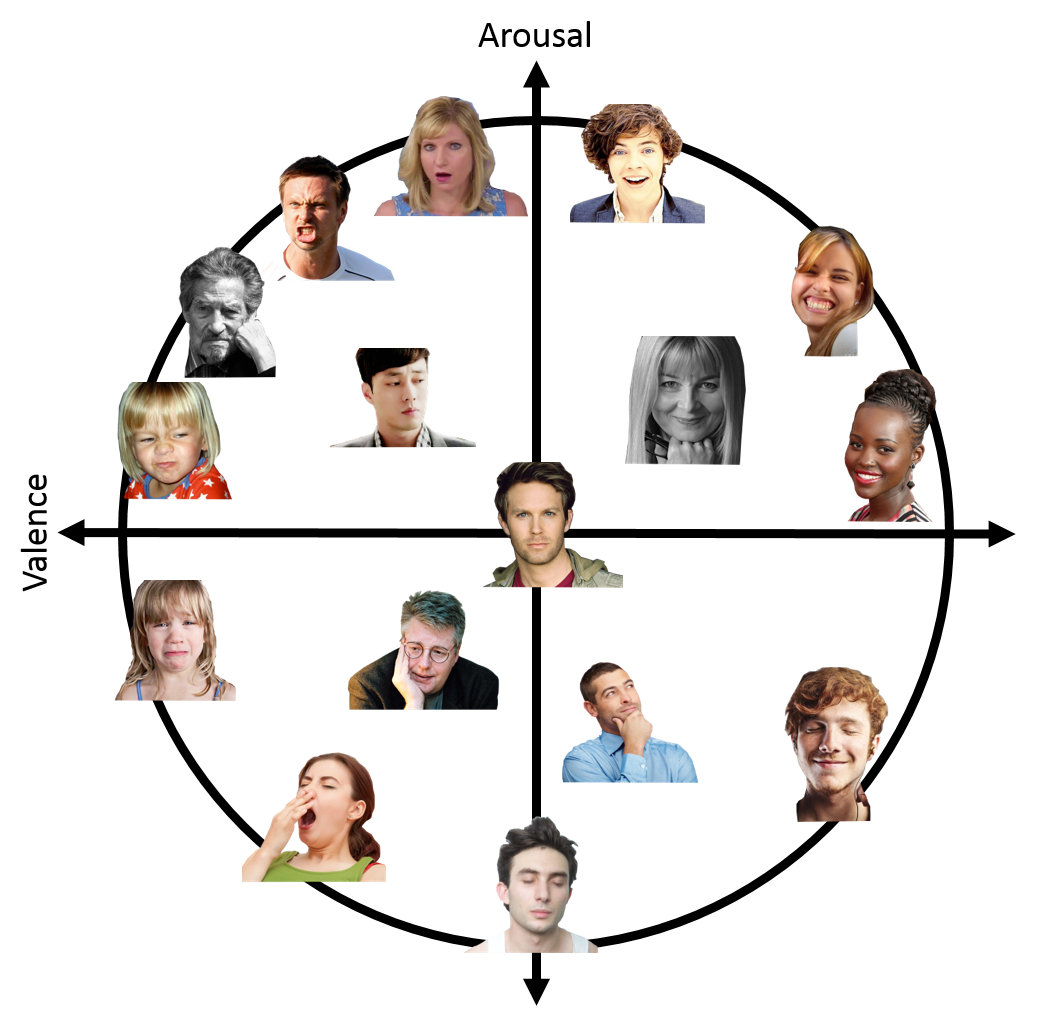}
	\vspace{-6mm}
	\caption{Sample images in Valence Arousal circumplex}
	\label{fig:valenceArousal}
\end{figure}

There are several models in the literature to quantify affective facial behaviors: 1) categorical model, where the emotion/affect is chosen from a list of affective-related categories such as six basic emotions defined by Ekman \textit{et al.}~\cite{ekman1971constants}, 2) dimensional model, where a value is chosen over a continuous emotional scale, such as valence and arousal~\cite{russell1980circumplex} and 3) Facial Action Coding System (FACS) model, where all possible facial actions are described in terms of Action Units (AUs)~\cite{ekman1977facial}. FACS model explains facial movements and does not describe the affective state directly. There are several methods to convert AUs to affect space (e.g., EMFACS~\cite{friesen1983emfacs} states that the occurrence of AU6 and AU12 is a sign of happiness). In the categorical model, mixed emotions cannot adequately be transcribed into a limited set of words. Some researchers tried to define multiple distinct compound emotion categories (e.g., happily surprised, sadly fearful)~\cite{du2014compound} to overcome this limitation. However, still the set is limited, and the intensity of the emotion cannot be defined in the categorical model. In contrast, the dimensional model of affect can distinguish between subtly different displays of affect and encode small changes in the intensity of each emotion on a continuous scale, such as valence and arousal. Valence refers to how positive or negative an event is, and arousal reflects whether an event is exciting/agitating or calm/soothing~\cite{russell1980circumplex}. Figure~\ref{fig:valenceArousal} shows samples of facial expressions represented in the 2D space of valence and arousal. As it is shown, there are several different kinds of affect and small changes in the same emotion that cannot be easily mapped into a limited set of terms existing in the categorical model.

The dimensional model of affect covers both intensity and different emotion categories in the continuous domain. Nevertheless, there are relatively fewer studies on developing automated algorithms in measuring affect using the continuous dimensional model (e.g., valence and arousal). One of the main reasons is that creating a large database to cover the entire continuous space of valence and arousal is expensive and there are very limited annotated face databases in the continuous domain. This paper contributes to the field of affective computing by providing a large annotated face database of the dimensional as well as the categorical models of affect.

The majority of the techniques for automated affective computing and FER are based on supervised machine learning methodologies. These systems require annotated image samples for training. Researchers have created databases of human actors/subjects portraying basic emotions~\cite{lyons1998coding, tian2001recognizing, lucey2010extended, pantic2005web, gross2010multiPie}. Most of these databases mainly contain posed expressions acquired in a controlled lab environment. However, studies show that posed expressions can be different from unposed expressions in configuration, intensity, and timing~\cite{cohn2004timing, valstar2006spontaneous}. Some researchers captured unposed facial behavior while the subject is watching a short video~\cite{mavadati2013disfa, mcduff2013affectiva}, engaged in laboratory-based emotion inducing tasks~\cite{sneddon2012belfast}, or interacted with a computer-mediated tutoring system~\cite{grafsgaard2013automatically}. Although a large number of frames can be obtained by these approaches, the diversity of these databases is limited due to the number of subjects, head position, and environmental conditions.

Recently, databases of facial expression and affect in the wild received much attention. These databases are either captured from movies or the Internet, and annotated with categorical model~\cite{dhall2013emotion, FER2013, Mollahosseini2016FacialWild}, dimensional model~\cite{zafeiriou_in_the_wild_2016}, and FACS model~\cite{benitez2016emotionet}. However, they only cover one model of affect, have a limited number of subjects, or contain few samples of certain emotions such as disgust. Therefore, a large database, with a large amount of subject variations in the wild condition that covers multiple models of affect (especially the dimensional model) is a need.


To address this need, we created a database of facial \textbf{Affect} from the Inter\textbf{Net} (called \textit{AffectNet}) by querying different search engines (Google, Bing, and Yahoo) using 1250 emotion related tags in six different languages (English, Spanish, Portuguese, German, Arabic, and Farsi). \textit{AffectNet} contains more than one million images with faces and extracted facial landmark points. Twelve human experts manually annotated 450,000 of these images in both categorical and dimensional (valence and arousal) models and tagged the images that have any occlusion on the face. Figure~\ref{fig:valenceArousal} shows sample images from \textit{AffectNet} and their valence and arousal annotations.

To calculate the agreement level between the human labelers, 36,000 images were annotated by two human labelers. \textit{AffectNet} is by far the largest database of facial affect in still images which covers both categorical and dimensional models. The cropped region of the facial images, the facial landmark points, and the affect labels will be publicly available to the research community\footnote{Interested researcher can download a copy of AffectNet from: \url{http://mohammadmahoor.com/databases-codes/}}. Considering the lack of in-the-wild large facial expressions datasets and more specifically annotated face datasets in the continuous domain of valence and arousal, \textit{AffectNet} is a great resource which will enable further progress in developing automated methods for facial behavior computing in both the categorical and continuous dimensional spaces.
 
The rest of this paper is organized as follows. Section~\ref{sec:databaseOverviews} reviews the existing databases and state-of-the-art methods for facial expression recognition with emphasis on the dimensional model and in the wild setting databases. Section~\ref{sec:AffectNet} explains the process of collecting \textit{AffectNet} images from the Internet and annotating the categorical and dimensional models. Section~\ref{sec:baseline} presents two different baselines for automatic recognition of categorical emotions and prediction of dimensional valence and arousal in the continuous space using \textit{AffecNet} images. Finally Section ~\ref{sec:conclusion} concludes the paper. 
 
\section{Related Work}
\label{sec:databaseOverviews}

\subsection{Existing databases}
\label{Sec:ExistingDatabases}

\begin{table*}[]
	\centering
	\caption{The Summary and Characteristics of Reviewed Databases in Affect Recognition}
	\label{tab:datasetsOverview}
	\vspace{-2mm}
\begin{tabular}{|l|l|l|l|l|}
\hline
\textbf{Database}                                                         & \textbf{Database information}                                                                                                                                  & \textbf{\# of Subjects}                                                           & \textbf{Condition}                                                      & \textbf{Affect Modeling}                                                                                                  \\ \hline \hline
CK+~\cite{lucey2010extended}                                              & - Frontal and 30 degree images                                                                                                                                 & - 123                                                                             & \begin{tabular}[c]{@{}l@{}}- Controlled\\ - Posed\end{tabular}                    & \begin{tabular}[c]{@{}l@{}}- 30 AUs\\ - 7 emotion categories\end{tabular}                                 \\ \hline
MultiPie~\cite{gross2010multiPie}                                                                  & \begin{tabular}[c]{@{}l@{}}- Around 750,000 images \\ - Under multiple viewpoints and illuminations\end{tabular}                                                                                                                                 & - 337                                                                             & \begin{tabular}[c]{@{}l@{}}- Controlled\\ - Posed\end{tabular}                    & - 7 emotion categories                              \\ \hline
MMI~\cite{pantic2005web}                                                  & \begin{tabular}[c]{@{}l@{}}- Subjects portrayed 79 series of facial expressions \\ - Image sequence of frontal and side view are captured\end{tabular}         & - 25                                                                              & \begin{tabular}[c]{@{}l@{}}- Controlled \\ - Posed \\ \& Spontaneous\end{tabular} & \begin{tabular}[c]{@{}l@{}}- 31 AUs\\ - Six basic expression\end{tabular}                                                 \\ \hline
DISFA~\cite{mavadati2013disfa}                                            & \begin{tabular}[c]{@{}l@{}}- Video of subjects while watching a four minutes video \\ - Clip are recorded by a stereo camera \end{tabular}                     & - 27                                                                              & \begin{tabular}[c]{@{}l@{}}- Controlled \\ - Spontaneous\end{tabular}             & - 12 AUs                                                                                                                  \\ \hline
SALDB~\cite{nicolaou2010audio,nicolaou2011continuous} & \begin{tabular}[c]{@{}l@{}}- SAL\\ - Audiovisual (facial expression,shoulder, audiocues)\\ - 20 facial feature points, 5 shoulder points for video\end{tabular} & \begin{tabular}[c]{@{}l@{}}- 4 \end{tabular} & \begin{tabular}[c]{@{}l@{}}- Controlled \\ - Spontaneous\end{tabular}             & \begin{tabular}[c]{@{}l@{}}- Valence \\ - Quantized~\cite{nicolaou2010audio}\\ - Continuous~\cite{nicolaou2011continuous}\end{tabular} \\ \hline
RELOCA~\cite{ringeval2013introducing}                                     & - Multi-modal audio, video, ECG and EDA                                                                                                                        & - 46                                                                              & \begin{tabular}[c]{@{}l@{}}- Controlled \\ - Spontaneous\end{tabular}             & \begin{tabular}[c]{@{}l@{}}- Valence and arousal \\ (continuous)\\ - Self assessment\end{tabular}                         \\ \hline
AM-FED~\cite{mcduff2013affectiva}                                         & - 242 facial videos                                                                                                                                            & \begin{tabular}[c]{@{}l@{}}- 242\end{tabular}                                     & \begin{tabular}[c]{@{}l@{}} - Spontaneous \end{tabular}       & - 14 AUs                                                                                                                   \\ \hline
DEAP~\cite{koelstra2012deap}                                                     & \begin{tabular}[c]{@{}l@{}}- 40 one-minute long videos shown to subjects \\ - EEG signals recorded \end{tabular}                                               & - 32                                                                  & \begin{tabular}[c]{@{}l@{}} - Controlled \\ - Spontaneous \end{tabular}												                       & \begin{tabular}[c]{@{}l@{}} - Valence and arousal \\ (continuous) \\ - Self assessment\end{tabular}                                           \\ \hline
AFEW~\cite{dhall2013emotion}                                              & - Videos                                                                                                                                                       & - 330                                                                             & - Wild                                                                            & - 7 emotion categories                                                                                                    \\ \hline
FER-2013~\cite{FER2013}                                                   & - Images queried from web                                                                                                                                      & - $\sim$35,887                                                                    & - Wild                                                                            & - 7 emotion categories                                                                                                    \\ \hline
EmotioNet~\cite{benitez2016emotionet}                                     & \begin{tabular}[c]{@{}l@{}}- Images queried from web\\ - 100,000 images annotated manually\\ - 900,000 images annotated automatically\end{tabular}             & - $\sim$100,000                                                                   & - Wild                                                                            & \begin{tabular}[c]{@{}l@{}}- 12 AUs annotated\\ - 23 emotion categories\\ based on AUs\end{tabular}                       \\ \hline
Aff-Wild~\cite{zafeiriou_in_the_wild_2016}                                & - 500 videos from YouTube                                                                                                                                      & - 500                                                                             & - Wild																	           & \begin{tabular}[c]{@{}l@{}}- Valence and arousal \\ (continuous)\end{tabular}                                             \\ \hline
FER-Wild~\cite{Mollahosseini2016FacialWild}                         & - 24,000 images from web                                                                                                                                       & - $\sim$24,000                                                                    & - Wild																	           & - 7 emotion categories                                             \\ \hline

\begin{tabular}[c]{@{}l@{}}\textit{\textbf{AffectNet}}\\ \textbf{(This work)}\end{tabular} & \begin{tabular}[c]{@{}l@{}}\textbf{- 1,000,000 images with 
facial landmarks}\\ \textbf{- 450,000 images annotated manually}\end{tabular}                                                & \textbf{- 
$\sim$450,000 }                                                                  & \textbf{- 
Wild}                                                                            & \begin{tabular}[c]{@{}l@{}}\textbf{- 8 
emotion categories}\\ \textbf{- Valence and arousal}\\ \textbf{(continuous)}\end{tabular}                     \\ \hline
\end{tabular}
\end{table*}

Early databases of facial expressions such as JAFFE~\cite{lyons1998coding}, Cohn-Kanade~\cite{tian2001recognizing, lucey2010extended}, MMI~\cite{pantic2005web}, and MultiPie~\cite{gross2010multiPie} were captured in a lab-controlled environment where the subjects portrayed different facial expressions. This approach resulted in a clean and high-quality database of posed facial expressions. However, posed expressions may differ from daily life unposed (aka spontaneous) facial expressions. Thus, capturing spontaneous expression became a trend in the affective computing community. Examples of these environments are recording the responses of participants' faces while watching a stimuli (e.g., DISFA~\cite{mavadati2013disfa}, AM-FED~\cite{mcduff2013affectiva}) or performing laboratory-based emotion inducing tasks (e.g., Belfast~\cite{sneddon2012belfast}). These databases often capture multi-modal affects such as voice, biological signals, etc. and usually a series of frames are captured that enable researchers to work on temporal and dynamic aspects of expressions. However, the diversity of these databases is limited due to the number of subjects, head pose variation, and environmental conditions.

Hence there is a demand to develop systems that are based on natural, unposed facial expressions. To address this demand, recently researchers paid attention to databases in the wild. Dhall~\emph{et al.}~\cite{dhall2013emotion} released \textit{Acted Facial Expressions in the Wild (AFEW)} from 54 movies by a recommender system based on subtitles. The video clips were annotated with six basic expressions plus neutral. AFEW contains 330 subjects aged 1-77 years and addresses the issue of temporal facial expressions in the wild. A static subset (SFEW~\cite{dhall2011static}) is created by selecting some frames of AFEW. SFEW covers unconstrained facial expressions, different head poses, age range, occlusions, and close to real world illuminations. However, it contains only 700 images, and there are only 95 subjects in the database. 

The \textit{Facial Expression Recognition 2013 (FER-2013)} database was introduced in the ICML 2013 Challenges in Representation Learning~\cite{FER2013}. The database was created using the Google image search API that matched a set of 184 emotion-related keywords to capture the six basic expressions as well as the neutral expression. Images were resized to 48x48 pixels and converted to grayscale. Human labelers rejected incorrectly labeled images, corrected the cropping if necessary, and filtered out some duplicate images. The resulting database contains 35,887 images most of which are in the wild settings. FER-2013 is currently the biggest publicly available facial expression database in the wild settings, enabling many researchers to train machine learning methods such as Deep Neural Networks (DNNs) where large amounts of data are needed. In FER-2013, the faces are not registered, a small number of images portray disgust (547 images), and unfortunately most of facial landmark detectors fail to extract facial landmarks at this resolution and quality. In addition, only the categorical model of affect is provided with FER-2013.

The \textit{Affectiva-MIT Facial Expression Dataset (AM-FED)} database~\cite{mcduff2013affectiva} contains 242 facial videos (160K frames) of people watching Super Bowl commercials using their webcam. The recording conditions were arbitrary with different illumination and contrast. The database was annotated frame-by-frame for the presence of 14 FACS action units, head movements, and automatically detected landmark points. AM-FED is a great resource to learn AUs in the wild. However, there is not a huge variance in head pose (limited profiles), and there are only a few subjects in the database.

The FER-Wild~\cite{Mollahosseini2016FacialWild} database contains 24,000 images that are obtained by querying emotion-related terms from three search engines. The OpenCV face recognition was used to detect faces in the images, and 66 landmark points were found using Active Appearance Model (AAM)~\cite{mollahosseini2013bidirectional} and a face alignment algorithm via regression local binary features~\cite{ren2014face, LequanYu2016}. Two human labelers annotated the images into six basic expressions and neutral. Comparing with FER-2013, FER-Wild images have a higher resolution with facial landmark points necessary to register the images. However, still a few samples portray some expressions such as disgust and fear and only the categorical model of affect is provided with FER-Wild.

The \textit{EmotioNet}~\cite{benitez2016emotionet} consists of one million images of facial expressions downloaded from the Internet by selecting all the words derived from the word ``feeling'' in WordNet~\cite{miller1995wordnet}. Face detector~\cite{viola2004robust} was used to detect faces in these images and the authors visually inspected the resultant images. These images were then automatically annotated with AUs and AU intensities by an approach based on Kernel Subclass Discriminant Analysis (KSDA)~\cite{you2011kernel}. The KSDA-based approach was trained with Gabor features centered on facial landmark with a Radial Basis Function (RBF) kernel. Images were labeled as one of the 23 (basic or compound) emotion categories defined in~\cite{du2014compound} based on AUs. For example, if an image has been annotated as having AUs 1, 2, 12 and 25, it is labeled as happily surprised. A total of 100,000 images (10\% of the database) were manually annotated with AUs by experienced coders. The proposed AU detection approach was trained on CK+~\cite{lucey2010extended}, DISFA~\cite{mavadati2013disfa}, and CFEE~\cite{lucey2011painful} databases, and the accuracy of the automated annotated AUs was reported about 80\% on the manually annotated set. EmotioNet is a novel resource of FACS model in the wild with a large amount of subject variation. However, it lacks the dimensional model of affect, and the emotion categories are defined based on annotated AUs and not manually labeled.

On the other hand, some researchers developed databases of the dimensional model in the continuous domain. These databases, however, are limited since the annotation of continuous dimensions is more expensive and necessitate trained annotators. Examples of these databases are \textit{Belfast}~\cite{sneddon2012belfast}, \textit{RECOLA}~\cite{ringeval2013introducing}, \textit{Affectiva-MIT Facial Expression Dataset (AM-FED)}~\cite{mcduff2013affectiva}, and recently published \textit{Aff-Wild Database}~\cite{zafeiriou_in_the_wild_2016} which is the only database of dimensional model in the wild.

The Belfast database~\cite{sneddon2012belfast} contains recordings (5s to 60s in length) of mild to moderate emotional responses of 60 participants to a series of laboratory-based emotion inducing tasks (e.g., surprise response by setting off a loud noise when the participant is asked to find something in a black box). The recordings were labeled by information on self-report of emotion, the gender of the participant/experimenter, and the valence in the continuous domain. The arousal dimension was not annotated in Belfast database. While the portrayed emotions are natural and spontaneous, the tasks have taken place in a relatively artificial setting of a laboratory where there was a control on lighting conditions, head poses, etc.

The \textit{Database for Emotion Analysis using Physiological Signals (DEAP)}~\cite{koelstra2012deap} consists of spontaneous reactions of 32 participants in response to one-minute long music video clip. The EEG, peripheral physiological signals, and frontal face videos of participants were recorded, and the participants rated each video in terms of valence, arousal, like/dislike, dominance, and familiarity. Correlations between the EEG signal frequencies and the participants’ ratings were investigated, and three different modalities, i.e., EEG signals, peripheral physiological signals, and multimedia features on video clips (such as lighting key, color variance, etc.) were used for binary classification of low/high arousal, valence, and liking. DEAP is a great database to study the relation of biological signals and dimensional affect, however, it has only a few subjects and the videos are captured in lab controlled settings.

The RECOLA benchmark~\cite{ringeval2013introducing} contains videos of 23 dyadic teams (46 participants) that participated in a video conference completing a task which required collaboration. Different multi-modal data of the first five minutes of interaction, i.e., audio, video, ECG and EDA) were recorded continuously and synchronously. Six annotators measured arousal and valence. The participants reported their arousal and valence through the Self-Assessment Manikin (SAM)~\cite{bradley1994measuring} questionnaire before and after the task. RECOLA is a great database of the dimensional model with multiple cues and modalities, however, it contains only 46 subjects and the videos were captured in the lab controlled settings.

Audio-Visual Emotion recognition Challenge (AVEC) series of competitions ~\cite{schuller2011avec, schuller2012avec, valstar2013avec, valstar2014avec, ringeval2015avec, valstar2016avec} provided a benchmark of automatic audio, video and audiovisual emotion analysis in continuous affect recognition. AVEC 2011, 2012, 2013, and 2014 used videos from the SEMAINE~\cite{mckeown2012semaine} database videos. Each video is annotated by a single rater for every dimension using a two-axis joystick. AVEC 2015 and 2016 used the RECOLA benchmark in their competitions. Various continuous affect recognition dimensions were explored in each challenge year such as valence, arousal, expectation, power, and dominance, where the prediction of valence and arousal are studied in all challenges. 

The \textit{Aff-Wild} Database~\cite{zafeiriou_in_the_wild_2016} is by far the largest database for measuring continuous affect in the valence-arousal space ``in-the-wild''. More than 500 videos from YouTube were collected. Subjects in the videos displayed a number of spontaneous emotions while watching a particular video, performing an activity, and reacting to a practical joke. The videos have been annotated frame-by-frame by three human raters, utilizing a joystick-based tool to rate valence and arousal. \textit{Aff-Wild} is a great database of dimensional modeling in the wild that considers the temporal changes of the affect, however, it has a small subject variance, i.e., it only contains 500 subjects.

Table~\ref{tab:datasetsOverview} summarizes the characteristics of the reviewed databases in all three models of affect, i.e., categorical model, dimensional model, and Facial Action Coding System (FACS).

\begin{table*}[htbp]
	\centering
	\caption{State-of-the-art Algorithms and Their Performance on the Databases Listed in Table~\ref{tab:datasetsOverview}.}
	\label{Tab:ExisitngAlgorithms}
	\vspace{-2mm}
\begin{tabular}{|l|l|l|l|}
\hline
\multicolumn{1}{|c|}{\textbf{Work}}                                       & \multicolumn{1}{c|}{\textbf{Database}}                    & \multicolumn{1}{c|}{\textbf{Method}}                                                                                                                                                                                 & \multicolumn{1}{c|}{\textbf{Results}}                                                                                                                                   \\ \hline \hline
\begin{tabular}[c]{@{}l@{}}Mollahosseini \\ \textit{et al.}~\cite{mollahosseini2015going}\end{tabular}  & \begin{tabular}[c]{@{}l@{}} CK+ \\ MultiPie\end{tabular} & \begin{tabular}[c]{@{}l@{}}- Inception based Convolutional Neural Network (CNN)\\ - Subject-independent and cross-database experiments\end{tabular}                                                                                                 & \begin{tabular}[c]{@{}l@{}}- 93.2\% accuracy on CK+\\ - 94.7\% accuracy on MultiPie\end{tabular}                                   \\ \hline
\begin{tabular}[c]{@{}l@{}}Shan \textit{et al.}~\cite{shan2009facial}\end{tabular}  & \begin{tabular}[c]{@{}l@{}} MMI\end{tabular} & \begin{tabular}[c]{@{}l@{}}-  Different SVM kernels trained with LBP features \\- Subject-independent and cross-database experiments\end{tabular}                                                                                                 & \begin{tabular}[c]{@{}l@{}}- 86.9\% accuracy on MMI\end{tabular}                                   \\ \hline
\begin{tabular}[c]{@{}l@{}}Zhang \textit{et al.}~\cite{zhang2015facial}\end{tabular} & \begin{tabular}[c]{@{}l@{}}DISFA\end{tabular} & \begin{tabular}[c]{@{}l@{}}- $l_p norm$ multi-task multiple kernel learning\\ - learning shared kernels from a given set of base kernels\end{tabular}         & \begin{tabular}[c]{@{}l@{}}- 0.70 F1-score on DISFA \\ - 0.93 recognition rate on DISFA\end{tabular}                                                                                                                                         \\ \hline
\begin{tabular}[c]{@{}l@{}}Nicolaou \\ \textit{et al.}~\cite{nicolaou2011continuous}\end{tabular}       & SALDB                                            & \begin{tabular}[c]{@{}l@{}}- Bidirectional LSTM\\ - Trained on multiple engineered features extracted \\ ~ from audio, facial geometry , and shoulder\end{tabular}                                                     & \begin{tabular}[c]{@{}l@{}}- Leave-one-sequence-out\\ - BLSTM-NN outperform SVR \\ - Valence (RMSE=0.15 and CC=0.796)\\ - Arousal (RMSE=0.21 and CC=0.642)\end{tabular} \\ \hline
He \textit{et al.}~\cite{he2015multimodal}                                                        & RECOLA                                                    & \begin{tabular}[c]{@{}l@{}}- Multiple stack of bidirectional LSTM (DBLSTM-RNN)\\ - Trained on engineered features extracted from audio (LLDs), \\ ~ video (LPQ-TOP), 52 ECG features, and 22 EDA features\end{tabular} & \begin{tabular}[c]{@{}l@{}}- Winner of AVEC 2015 challenge \\ - Valence (RMSE=0.104 and CC=0.616) \\ - Arousal (RMSE=0.121 and CC=0.753)\end{tabular}                     \\ \hline
McDuff \textit{et al.}~\cite{mcduff2013affectiva}                                                      & AM-FED                                                    & \begin{tabular}[c]{@{}l@{}}- HOG features extracted\\ - SVM with RBF kernel\end{tabular}                                                                                                                                  & \begin{tabular}[c]{@{}l@{}}- AUC 0.90, 0.72 and 0.70 for smile, \\ AU2 and AU4 respectively\end{tabular}                                        \\ \hline
Koelstra \textit{et al.}~\cite{koelstra2012deap}                                                       & DEAP                                                    & \begin{tabular}[c]{@{}l@{}}- Gaussian naive Bayes classifier \\ - EEG, physiological signals, and multimedia features \\ - Binary classification of low/high arousal, valence, and liking \end{tabular}                                                                                                                                  & \begin{tabular}[c]{@{}l@{}}- 0.39 F1-score on Arousal \\ - 0.37 F1-score on Valence \\ - 0.40 F1-score on Liking \end{tabular}                                        \\ \hline
Fan \textit{et al.}~\cite{fan2016video}                                                       & AFEW                                                      & \begin{tabular}[c]{@{}l@{}}- Trained on both video and audio.\\ - VGG network are followed by LSTMs and combined with \\ ~  3D convolution\end{tabular}                                               & \begin{tabular}[c]{@{}l@{}}- Winner of EmotiW 2016 challenge\\ - 56.16\% accuracy on AFEW\end{tabular} \\ \hline
Tang \textit{et al.}~\cite{tang2013deep}                                                        & FER-2013                                                  & - CNN with linear one-vs-all SVM at the top                                                                                                                                                                        & \begin{tabular}[c]{@{}l@{}}- Winner of the FER challenge\\ - 71.2\% accuracy on test set\end{tabular}                                                                   \\ \hline
\begin{tabular}[c]{@{}l@{}}Benitez-Quiroz \\ \textit{et al.}~\cite{benitez2016emotionet}\end{tabular} & \begin{tabular}[c]{@{}l@{}} EmotioNet\end{tabular} & \begin{tabular}[c]{@{}l@{}}- New face feature extraction method using Gabor filters\\ - KSDA classification\\ - Subject-independent and cross-database experiments\end{tabular}                                      & - $\sim$80\% AU detection on EmotioNet                                                                                                                                          \\ \hline
\begin{tabular}[c]{@{}l@{}}Mollahosseini \\ \textit{et al.}~\cite{Mollahosseini2016FacialWild}\end{tabular}  & \begin{tabular}[c]{@{}l@{}} FER-Wild\end{tabular} & \begin{tabular}[c]{@{}l@{}}- Trained on AlexNet \\ -  Noise estimation methods used \end{tabular}                                                                                                 & \begin{tabular}[c]{@{}l@{}}- 82.12\% accuracy on FER-Wild\end{tabular}                                   \\ \hline
\end{tabular}
\end{table*}

\subsection{Evaluation Metrics}
\label{Sec:EvaluationMetrics}
There are various evaluation metrics in the literature to measure the reliability of annotation and automated affective computing systems. Accuracy, F1-score~\cite{sokolova2006beyond}, Cohen’s kappa~\cite{cohen1960}, Krippendorf’s Alpha~\cite{krippendorff1970estimating}, ICC~\cite{shrout1979intraclass}, area under the ROC curve (AUC), and area under Precision-Recall curve (AUC-PR)~\cite{jeni2013facing} are well-defined widely used metrics for evaluation of the categorical and FACS-based models. Since, the dimensional model of affect is usually evaluated in a continuous domain, different evaluation metrics are necessary. In the following, we review several metrics that are used in the literature for evaluation of dimensional model.

Root Mean Square Error (RMSE) is the most common evaluation metric in a continuous domain which is defined as:
\begin{equation}
RMSE = \sqrt{\frac{1}{n} \sum_{i=1}^{n}(\hat{\theta}_i-\theta_i)^2}
\end{equation}
where $\hat{\theta}_i$ and $\theta_i$ are the prediction and the ground truth of $i^{\text{th}}$ sample, and $n$ is the number of samples in the evaluation set. RMSE-based evaluation can heavily weigh the outliers~\cite{bermejo2001oriented}, and it is not able to provide the covariance of prediction and ground-truth to show how they change with respect to each other. Pearson’s correlation coefficient is therefore proposed in some literature~\cite{nicolaou2011continuous, schuller2011avec, schuller2012avec} to overcome this limitation:
\begin{equation}
CC = \frac{COV\{\hat{\theta}, \theta\}}{\sigma_{\hat{\theta}}\sigma_{\theta}} = 
\frac{E [(\hat{\theta}-\mu_{\hat{\theta}})(\theta-\mu_{\theta})]}{\sigma_{\hat{\theta}}\sigma_{\theta}}
\end{equation}

Concordance Correlation Coefficient (CCC) is another metric~\cite{ringeval2015avec, valstar2016avec} which combines the Pearson’s correlation coefficient (CC) with the square difference between the means of two compared time series:
\begin{equation}
\rho_c = \frac{2\rho \sigma_{\hat{\theta}} \sigma_{\theta}}{\sigma_{\hat{\theta}}^2 + \sigma_{\theta}^2 + (\mu_{\hat{\theta}} - \mu_\theta)^2}
\end{equation}
where $\rho$ is the Pearson correlation coefficient (CC) between two time-series (e.g., prediction and ground-truth), $\sigma_{\hat{\theta}}^2$ and $\sigma_{\theta}^2$ are the variance of each time series, and $\mu_{\hat{\theta}}$ and $\mu_{\theta}$ are the mean value of each. Unlike CC, the predictions that are well correlated with the ground-truth but shifted in value are penalized in proportion to the deviation in CCC. 

The value of valence and arousal are [-1,+1] and their signs are essential in many emotion-prediction applications. For example, if the ground-truth valence is +0.3, prediction of +0.7 is far better than prediction of -0.1, since +0.7 indicates a positive emotion similar to the ground-truth (despite both predictions have the same RMSE). Sign Agreement Metric (SAGR) is another metric that is proposed in~\cite{nicolaou2011continuous} to evaluate the performance of a valence and arousal prediction system. SAGR is defined as:
\begin{equation}
SAGR = \frac{1}{n}\sum_{i=1}^{n}\delta{(sign(\hat{\theta}_i),sign(\theta_i))}
\end{equation}
where $\delta$ is the Kronecker delta function, defined as:
\begin{equation}
\delta{(a,b)} = 
\begin{cases}
1,				& a = b\\
0,              & a \neq b
\end{cases}
\end{equation}   

The above discussed metrics are used to evaluate the categorical and dimensional baselines on \textit{AffectNet} in Sec.~\ref{sec:baseline}.

\subsection{Existing Algorithms}

Affective computing is now a well-established field, and there are many algorithms and databases for developing automated affect perception systems. Since it is not possible to include all those great works, we only give a brief overview and cover the state-of-the-art methods that are applied on the databases explained in Sec.~\ref{Sec:ExistingDatabases}.

Conventional algorithms of affective computing from faces use hand-crafted features such as pixel intensities~\cite{Mohammadi2014PCA_based}, Gabor filters~\cite{liu2002gabor}, Local Binary Patterns (LBP)~\cite{shan2009facial}, and Histogram of Oriented Gradients (HOG)~\cite{mavadati2013disfa}. These hand-crafted features often lack enough generalizability in the wild settings where there is a high variation in scene lighting, camera view, image resolution, background, subjects head pose and ethnicity.  

An alternative approach is to use Deep Neural Networks (DNN) to learn the most appropriate feature abstractions directly from the data and handle the limitations of hand-crafted features. DNNs have been a recent successful approach in visual object recognition~\cite{krizhevsky2012imagenet}, human pose estimation~\cite{toshev2014deeppose}, face verification~\cite{taigman2014deepface} and many more. This success is mainly due to the availability of computing power and existing big databases that allow DNNs to extract highly discriminative features from the data samples. There have been enormous attempts on using DNNs in automated facial expression recognition and affective computing~\cite{mollahosseini2015going, Mollahosseini2016FacialWild, tang2013deep, fan2016video, he2015multimodal} that are especially very successful in the wild settings.

Table~\ref{Tab:ExisitngAlgorithms} shows a list of the state-of-the-art algorithms and their performance on the databases listed in Table~\ref{tab:datasetsOverview}. As shown in the table, the majority of these approaches have used DNNs to learn a better representation of affect, especially in the wild settings. Even some of the approaches, such as the winner of the AVEC 2015 challenge~\cite{he2015multimodal}, trained a DNN with hand-crafted features and still could improve the prediction accuracy.

\section{AffectNet}
\label{sec:AffectNet}

\textbf{AffectNet} (\textbf{Affect} from the Inter\textbf{Net}) is the largest database of the categorical and dimensional models of affect in the wild (as shown in Table~\ref{tab:datasetsOverview}). The database is created by querying emotion related keywords from three search engines and annotated by expert human labelers. In this section, the process of querying the Internet, processing facial images and extracting facial landmarks, and annotating facial expression, valence, and arousal of affect are discussed.

\subsection{Facial Images from the Web}

\begin{figure*}[]
	\centering
	\includegraphics[width=\textwidth-4cm]{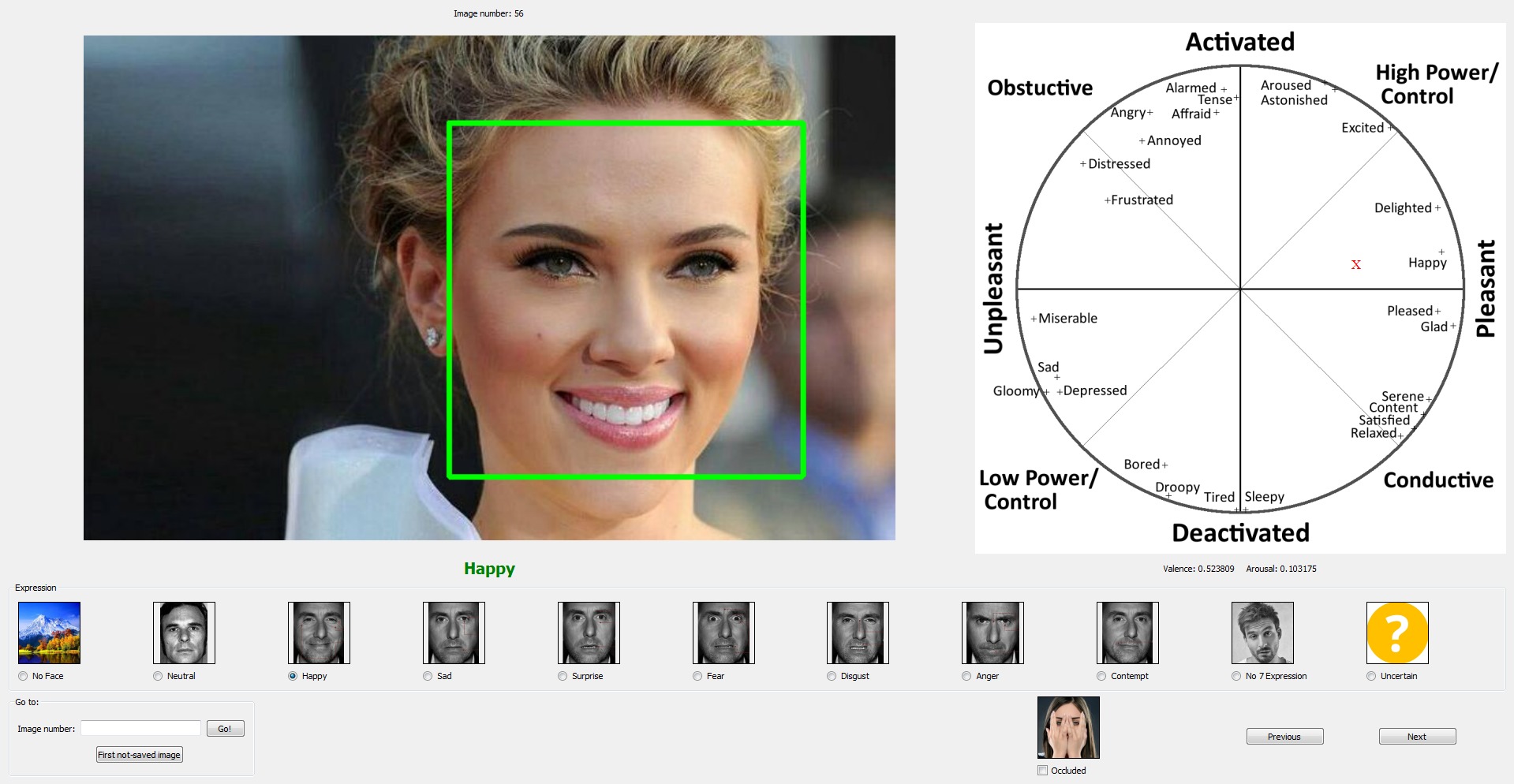}
	\vspace{-1mm}
	\caption{A screen-shot of the software application used to annotate categorical and dimensional (valence and arousal) 
	models of affect and the osculation tag if existing. Only one detected face in each image is annotated (shown in the 
	green bounding box).}
	\vspace{-3mm}
	\label{fig:annotationApScreenShot}
\end{figure*}

Emotion-related keywords were combined with words related to gender, age, or ethnicity, to obtain nearly 362 strings in the English language such as ``joyful girl'', ``blissful Spanish man'', ``furious young lady'', ``astonished senior''. These keywords are then translated into five other languages: Spanish, Portuguese, German, Arabic and Farsi. The direct translation of queries in English to other languages did not accurately result in the intended emotions since each language and culture has differing words and expressions for different emotions. Therefore, the list of English queries was provided to native non-English speakers who were proficient in English, and they created a list of queries for each emotion in their native language and inspected the quality of the results visually. The criteria for high-quality queries were those that returned a high percentage of human faces showing the intended queried emotions rather than drawings, graphics, or non-human objects. A total of 1250 search queries were compiled and used to crawl the search engines in our database. Since a high percentage of results returned by our query terms already contained neutral facial images, no individual query was performed to obtain additional neutral face. 

Three search engines (Google, Bing, and Yahoo) were queried with these 1250 emotion related tags. Other search engines such as Baidu and Yandex were considered. However, they either did not produce a large number of facial images with intended expressions or they did not have available APIs for automatically querying and pulling image URLs into the database. Additionally, queries were combined with negative terms (e.g., ``drawing'', ``cartoon'', ``animation'', ``birthday'', etc.) to avoid non-human objects as much as possible. Furthermore, since the images of stock photo websites are posed unnaturally and contain watermarks mostly, a list of popular stock photo websites was compiled  and the results returned from the stock photo websites were filtered out.

A total of $\sim$1,800,000 distinct URLs returned for each query were stored in the database. The OpenCV face recognition was used to obtain bounding boxes around each face. A face alignment algorithm via regression local binary features~\cite{ren2014face, LequanYu2016} was used to extract 66 facial landmark points. The facial landmark localization technique was trained using the annotations provided from the 300W competition~\cite{sagonas2015300}. More than 1M images containing at least one face with extracted facial landmark points were kept for further processing.

The average image resolution of faces in AffectNet are $425\times425$ with STD of $349\times349$ pixels. We used Microsoft cognitive face API to extract these facial attributes on 50,000 randomly selected images from the database. According to MS face API, 49\% of the faces are men. The average estimated age of the faces is 33.01 years with the standard deviation of 16.96 years. In particular, 10.85, 3.9, 30.19, 26.86, 14.46, and 13.75 percent of the faces are in age ranges [0, 10), [10, 20), [20, 30), [30, 40), [40, 50) and [50, -), respectively. MS face API detected forehead, mouth, and eye occlusions in 4.5, 1.08, and 0.49 percent of the images, respectively. Also, 9.63\% of the faces wear glasses, 51.07 and 41.4\% of the faces have eye and lip make-ups, respectively. In terms of head pose, the average estimated pitch, yaw, roll are 0.0,-0.7, and -1.19 degrees, respectively.

\subsection{Annotation}
\label{Sec:Annoation}

Crowd-sourcing services like Amazon Mechanical Turk are fast, cheap and easy approaches for labeling large databases. The quality of labels obtained from crowd-sourcing services, however, varies considerably among the annotators. Due to these issues and the fact that annotating the valence and arousal requires a deep understanding of the concept, we avoided crowd-sourcing facilities and instead hired 12 full-time and part-time annotators at the University of Denver to label the database. A total of 450,000 images were given to these expert annotators to label the face in the images into both discrete categorical and continuous dimensional (valence and arousal) models. Due to time and budget constraints each image was annotated by one annotator.

A software application was developed to annotate the categorical and dimensional (valence and arousal) models of affect. Figure~\ref{fig:annotationApScreenShot} shows a screen-shot of the annotation application. A comprehensive tutorial including the definition of the categorical and dimensional models of affect with some examples of each category, valence and arousal was given to the annotators. Three training sessions were provided to each annotator, in which the annotator labeled the emotion category, valence and arousal of 200 images and the results were reviewed with the annotators. Necessary feedback was given on both the categorical and dimensional labels. In addition, the annotators tagged the images that have any occlusion on the face. The occlusion criterion was defined as if any part of the face was not visible. If the person in the images wore glasses, but the eyes were visible without any shadow, it was not considered as occlusion. 

\subsubsection{Categorical Model Annotation}
\label{Sec:CategoricalModelAnnotation}
Eleven discrete categories were defined in the categorical model of \textit{AffectNet} as: Neutral, Happy, Sad, Surprise, Fear, Anger, Disgust, Contempt, None, Uncertain, and Non-face. The \emph{None} (``None of the eight emotions'') category is the type of expression/emotions (such as sleepy, bored, tired, seducing, confuse, shame, focused, etc.) that could not be assigned by annotators to any of the six basic emotions, contempt or neutral. However, valence and arousal could be assigned to these images. The \emph{Non-face} category was defined as images that: 1) Do not contain a face in the image; 2) Contain a watermark on the face; 3) The face detection algorithm fails and the bounding box is not around the face; 4) The face is a drawing, animation, or painted; and 5) The face is distorted beyond a natural or normal shape, even if an expression could be inferred. If the annotators were uncertain about any of the facial expressions, images were tagged as \emph{uncertain}. When an image was annotated as \emph{Non-face} or \textit{uncertain}, valence and arousal were not assigned to the image.

\begin{figure}
	\centering
	\subfloat
	{
		\centering
		\stackunder{\includegraphics[width=20mm]{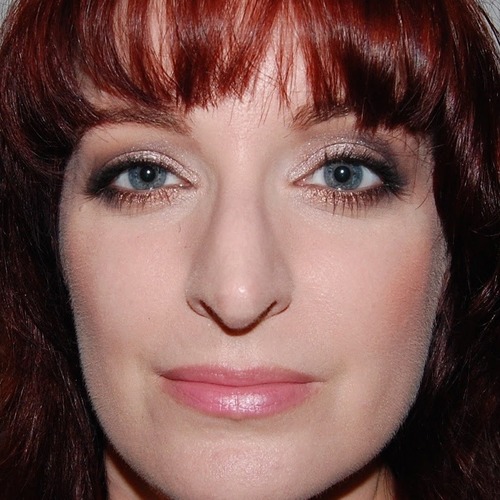}}{\scriptsize{\emph{Neutral}
				(Angry)}}%
	}
	\subfloat
	{
		\centering
		\stackunder{\includegraphics[width=20mm]{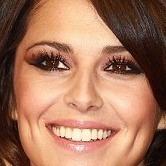}}{\scriptsize{\emph{Happy} 
				(Happy)}}%
	}
	\subfloat
	{
		\centering
		\stackunder{\includegraphics[width=20mm]{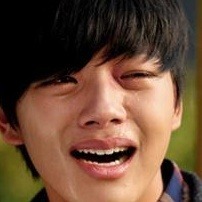}}{\scriptsize{\emph{Sad} 
				(Angry)}}%
	}
	\subfloat
	{
		\centering
		\stackunder{\includegraphics[width=20mm]{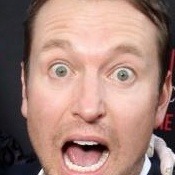}}{\scriptsize{\emph{Surprise} 
				(Fear)}}%
	}
	\vspace{-0.2cm}
	\subfloat
	{
		\centering
		\stackunder{\includegraphics[width=20mm]{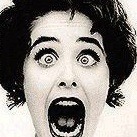}}{\scriptsize{\emph{Fear} 
				(Fear)}}%
	}	
	\subfloat
	{
		\centering
		\stackunder{\includegraphics[width=20mm]{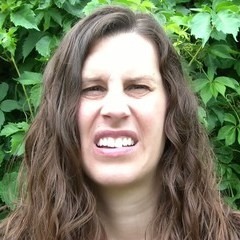}}{\scriptsize{\emph{Disgust}
				(Disgust)}}%
	}
	\subfloat
	{
		\centering
		\stackunder{\includegraphics[width=20mm]{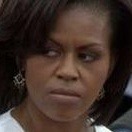}}{\scriptsize{\emph{Angry} 
				(Angry)}}%
	}
	\subfloat
	{
		\centering
		\stackunder{\includegraphics[width=20mm]{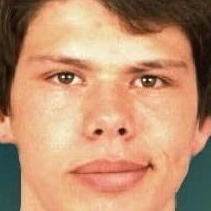}}{\scriptsize{\emph{Contempt}
				(Happy)}}%
	}
	\vspace{-0.2cm}
	\subfloat
	{
		\centering
		\stackunder{\includegraphics[width=20mm]{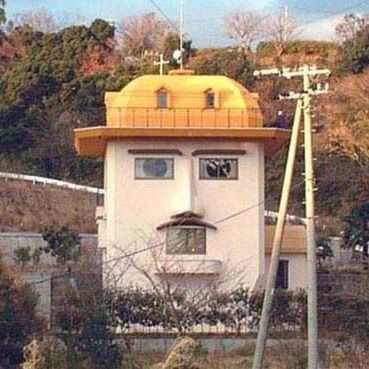}}{\scriptsize{\emph{Non-face} 
		(Surprise)}}%
	}
	\subfloat
	{
		\centering
		\stackunder{\includegraphics[width=20mm]{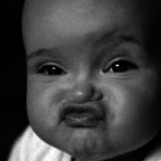}}{\scriptsize{\emph{Uncertain}
				(Sad)}}%
	}
	\subfloat
	{
		\centering
		\stackunder{\includegraphics[width=20mm]{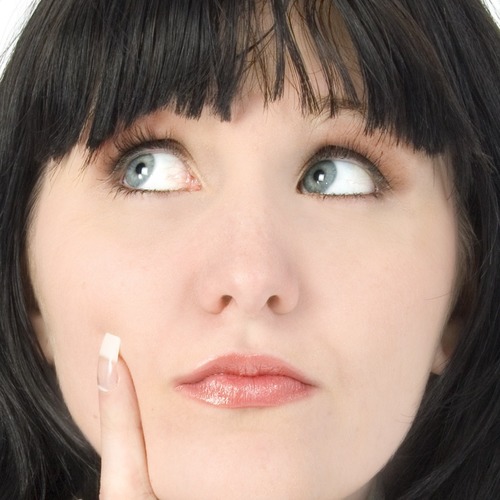}}{\scriptsize{\emph{None}
				(Fear)}}%
	}
	\subfloat
	{
		\centering
		\stackunder{\includegraphics[width=20mm]{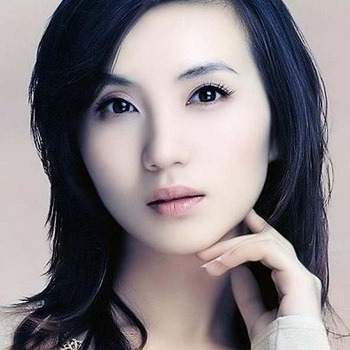}}{\scriptsize{\emph{None}
				(Happy)}}%
	}
	
	\caption{\label{fig:Sample_ExpressionNet}
		Samples of queried images from the web and their annotated tags. The queried expression is written in parentheses.
	}
\end{figure}

The annotators were instructed to select the proper expression category of the face, where the intensity is not important as long as the face depicts the intended emotion. Table~\ref{Tab:NumImages} shows the number of images in each category. Table~\ref{Tab:annotationConfusionMatrix} indicates the percentage of annotated categories for queried emotion terms. As shown, the happy emotion had the highest hit-rate (48\%), and the rest of the emotions had hit-rates less than 20\%. About 15\% of all query results were in the \emph{No-Face} category, as many images from the web contain watermarks, drawings, etc. About 15\% of all queried emotions resulted in neutral faces. Among other expressions, disgust, fear, and contempt had the lowest hit-rate with only 2.7\%, 4\%, and 2.4\% hit-rates, respectively. As one can see, the majority of the returned images from the search engines were happy or neutral faces. The authors believe that this is because people tend to publish their images with positive expressions rather than negative expressions. Figure~\ref{fig:Sample_ExpressionNet} shows a sample image in each category and its intended queries (in parentheses).

\begin{table}[]
	\centering
	\caption{Number of Annotated Images in Each Category}
	\label{Tab:NumImages}
	\vspace{-2mm}
	\begin{tabular}{|l|c|}
		\hline
		\multicolumn{1}{|c|}{\textbf{Expression}} & \textbf{Number} \\ \hline \hline
		Neutral                                   & 80,276           \\ \hline
		Happy                                     & 146,198          \\ \hline
		Sad                                       & 29,487           \\ \hline
		Surprise                                  & 16,288           \\ \hline
		Fear                                      & 8,191            \\ \hline
		Disgust                                   & 5,264            \\ \hline
		Anger                                     & 28,130           \\ \hline
		Contempt                                  & 5,135            \\ \hline
		None     		                          & 35,322           \\ \hline
		Uncertain                                 & 13,163           \\ \hline
		Non-Face	                              & 88,895			 \\ \hline
	\end{tabular}
\end{table}

\begin{table}[]
	\centering
	\caption{Percentage of Annotated Categories for Queried Emotion Terms (\%)}
	\label{Tab:annotationConfusionMatrix}
	\vspace{-2mm}
\begin{tabular}{ccccccccc}
	\cline{3-9}
	& \multicolumn{1}{c|}{}   & \multicolumn{7}{c|}{Query Expression}                                                                                                                                                                                                                          \\ \cline{3-9} 
	& \multicolumn{1}{c|}{}   & \multicolumn{1}{c|}{HA}            & \multicolumn{1}{c|}{SA}            & \multicolumn{1}{c|}{SU}            & \multicolumn{1}{c|}{FE}            & \multicolumn{1}{c|}{DI}          & \multicolumn{1}{c|}{AN}            & \multicolumn{1}{c|}{CO}            \\ \hline
	\multicolumn{1}{|c|}{\multirow{11}{*}{\begin{turn}{+90}Annotated Expression\end{turn}}} & \multicolumn{1}{c|}{NE\scriptsize{*}} & \multicolumn{1}{c|}{17.3}          & \multicolumn{1}{c|}{16.3}          & \multicolumn{1}{c|}{13.9}          & \multicolumn{1}{c|}{17.8}          & \multicolumn{1}{c|}{17.8}        & \multicolumn{1}{c|}{16.1}          & \multicolumn{1}{c|}{20.1}          \\ \cline{2-9} 
	\multicolumn{1}{|c|}{}                                       & \multicolumn{1}{c|}{HA} & \multicolumn{1}{c|}{\textbf{48.9}} & \multicolumn{1}{c|}{\textbf{27.2}} & \multicolumn{1}{c|}{\textbf{30.4}} & \multicolumn{1}{c|}{\textbf{28.6}} & \multicolumn{1}{c|}{\textbf{33}} & \multicolumn{1}{c|}{\textbf{29.5}} & \multicolumn{1}{c|}{\textbf{30.1}} \\ \cline{2-9} 
	\multicolumn{1}{|c|}{}                                       & \multicolumn{1}{c|}{SA} & \multicolumn{1}{c|}{2.6}           & \multicolumn{1}{c|}{15.7}          & \multicolumn{1}{c|}{4.8}           & \multicolumn{1}{c|}{5.8}           & \multicolumn{1}{c|}{4.5}         & \multicolumn{1}{c|}{5.4}           & \multicolumn{1}{c|}{4.6}           \\ \cline{2-9} 
	\multicolumn{1}{|c|}{}                                       & \multicolumn{1}{c|}{SU} & \multicolumn{1}{c|}{2.7}           & \multicolumn{1}{c|}{3.1}           & \multicolumn{1}{c|}{16}            & \multicolumn{1}{c|}{4.4}           & \multicolumn{1}{c|}{3.6}         & \multicolumn{1}{c|}{3.4}           & \multicolumn{1}{c|}{4.1}           \\ \cline{2-9} 
	\multicolumn{1}{|c|}{}                                       & \multicolumn{1}{c|}{FE} & \multicolumn{1}{c|}{0.7}           & \multicolumn{1}{c|}{1.2}           & \multicolumn{1}{c|}{4.2}           & \multicolumn{1}{c|}{4}             & \multicolumn{1}{c|}{1.5}         & \multicolumn{1}{c|}{1.4}           & \multicolumn{1}{c|}{1.3}           \\ \cline{2-9} 
	\multicolumn{1}{|c|}{}                                       & \multicolumn{1}{c|}{DI} & \multicolumn{1}{c|}{0.6}           & \multicolumn{1}{c|}{0.7}           & \multicolumn{1}{c|}{0.7}           & \multicolumn{1}{c|}{0.9}           & \multicolumn{1}{c|}{2.7}         & \multicolumn{1}{c|}{1.1}           & \multicolumn{1}{c|}{1}             \\ \cline{2-9} 
	\multicolumn{1}{|c|}{}                                       & \multicolumn{1}{c|}{AN} & \multicolumn{1}{c|}{2.8}           & \multicolumn{1}{c|}{4.5}           & \multicolumn{1}{c|}{3.8}           & \multicolumn{1}{c|}{5.6}           & \multicolumn{1}{c|}{6}           & \multicolumn{1}{c|}{12.2}          & \multicolumn{1}{c|}{6.1}           \\ \cline{2-9} 
	\multicolumn{1}{|c|}{}                                       & \multicolumn{1}{c|}{CO} & \multicolumn{1}{c|}{1.3}           & \multicolumn{1}{c|}{0.9}           & \multicolumn{1}{c|}{0.4}           & \multicolumn{1}{c|}{1.1}           & \multicolumn{1}{c|}{1.1}         & \multicolumn{1}{c|}{1.2}           & \multicolumn{1}{c|}{2.4}           \\ \cline{2-9} 
	\multicolumn{1}{|c|}{}                                       & \multicolumn{1}{c|}{NO} & \multicolumn{1}{c|}{5.4}           & \multicolumn{1}{c|}{8.7}           & \multicolumn{1}{c|}{4.8}           & \multicolumn{1}{c|}{8.1}           & \multicolumn{1}{c|}{8.8}         & \multicolumn{1}{c|}{9.3}           & \multicolumn{1}{c|}{11.2}          \\ \cline{2-9} 
	\multicolumn{1}{|c|}{}                                       & \multicolumn{1}{c|}{UN} & \multicolumn{1}{c|}{1.3}           & \multicolumn{1}{c|}{3.1}           & \multicolumn{1}{c|}{4.3}           & \multicolumn{1}{c|}{3.1}           & \multicolumn{1}{c|}{4.1}         & \multicolumn{1}{c|}{3.7}           & \multicolumn{1}{c|}{2.7}           \\ \cline{2-9} 
	\multicolumn{1}{|c|}{}                                       & \multicolumn{1}{c|}{NF} & \multicolumn{1}{c|}{16.3}          & \multicolumn{1}{c|}{18.6}          & \multicolumn{1}{c|}{16.7}          & \multicolumn{1}{c|}{20.6}          & \multicolumn{1}{c|}{16.9}        & \multicolumn{1}{c|}{16.8}          & \multicolumn{1}{c|}{16.3}          \\ \hline
	\multicolumn{9}{l}{\begin{tabular}[c]{@{}l@{}}\scriptsize{*} \scriptsize{NE, HA, SA, SU, FE, DI, AN, CO, NO, UN , and NF stand for Neutral,} \\ \scriptsize{Happy, Sad, Surprise, Fear, Anger, Disgust, Contempt, None, Uncertain,} \\ \scriptsize{ and Non-face categories, respectively.}\end{tabular}}
\end{tabular}
\end{table}

\subsubsection{Dimensional (Valence \& Arousal) Annotation}
\label{Sec:DimensionalAnnotation}
The definition of valence and arousal dimensions was adapted from~\cite{russell1980circumplex} and was given to annotators in our tutorial as: ``Valence refers to how positive or negative an event is, and arousal reflects whether an event is exciting/agitating or calm/soothing''. A sample circumplex with estimated positions of several expressions, borrowed from~\cite{paltoglou2013seeing}, was provided in the tutorial as a reference for the annotators. The provided circumplex in the tutorial contained more than 34 complex emotions categories such as suspicious, insulted, impressed, etc., and used to train annotators. The annotators were instructed to consider the intensity of valence and arousal during the annotation. During the annotation process, the annotators were supervised closely and constant necessary feedback was provided when they were uncertain about some images.

To model the dimensional affect of valence and arousal, a 2D Cartesian coordinate system was used where the $x$-axis and $y$-axis represent the valence and arousal, respectively. Similar to Russell's circumplex space model~\cite{russell1980circumplex}, our annotation software did not allow the value of valence and arousal outside of the circumplex. This allows us to convert the Cartesian coordinates to polar coordinates with $0 \leq r \leq 1$ and $0\leq \theta<360$. The annotation software showed the value of valence and arousal to the annotators when they selected a point in the circumplex. This helped the annotators to pick more precise locations of valence and arousal with a higher confidence. 

A predefined estimated region of valence and arousal was defined for each categorical emotion in the annotation software (e.g., for happy emotion the valence is in (0.0, 1.0], and the arousal is in [-0.2, 0.5] ). If the annotators select a value of valence and arousal outside of the selected emotion's region, the software indicates a warning message. The annotators were able to proceed, and they were instructed to do so, if they were confident about the value of valence and arousal.  The images with the warning messages were marked in the database, for further review by the authors. This helped to avoid mistakes in the annotation of the dimensional model of affect.   

\begin{figure}
	\centering
	\includegraphics[width=\columnwidth]{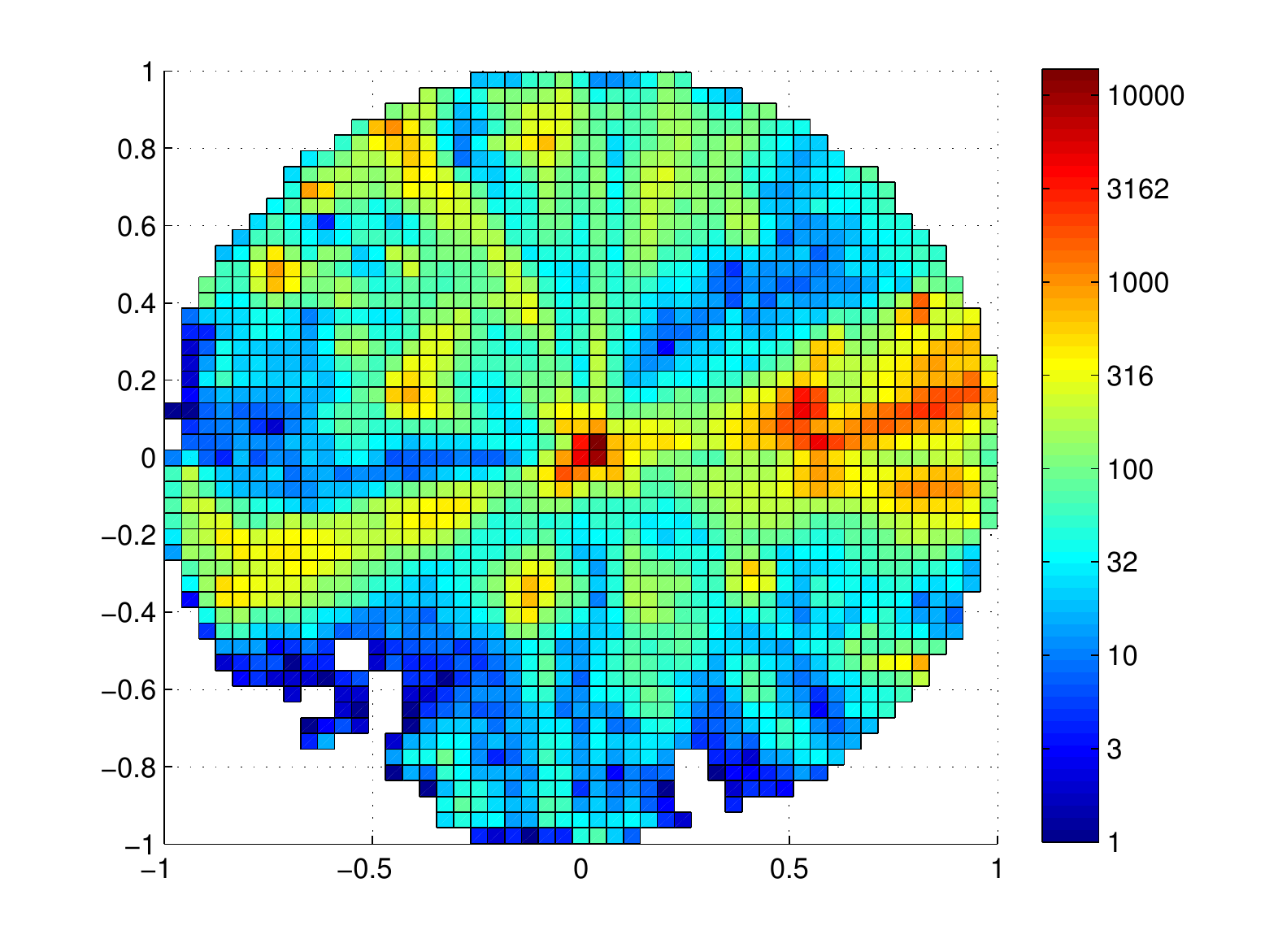}
	\vspace{-9mm}
	\caption{Histogram (number of frames in each range/area) of valence and arousal annotations (Best viewed in color).}
	\label{fig:valenceArousalAnnotationResult}
\end{figure}

Figure~\ref{fig:valenceArousalAnnotationResult} shows the histogram (number of samples in each range/area) of annotated images in a 2D Cartesian coordinate system. As illustrated, there are more samples in the center and the right middle (positive valence and small positive arousal) of the circumplex, which confirms the higher number of Neutral and Happy images in the database compared to other categories in the categorical model. \footnote{A numerical representation of annotated images in each range/area of valence and arousal is provided in the Appendix.}

\subsection{Annotation Agreement}
\label{Sec:AnnotationAgreement}

In order to measure the agreement between the annotators, 36,000 images were annotated by two annotators. The annotations were performed fully blind and independently, i.e., the annotators were not aware of the intended query or other annotator's response. The results showed that the annotators agreed on 60.7\% of the images. Table~\ref{tab:AnnotatorsAgreement} shows the agreement between two annotators for different categories. As it is shown, the annotators highly agreed on the \textit{Happy} and \textit{No Face} categories, and the highest disagreement occurred in the \textit{None} category. Visually inspecting some of the images in the \textit{None} category, the authors believe that the images in this category contain very subtle emotions and they can be easily confused with other categories (the last two example of Fig.~\ref{fig:Sample_ExpressionNet} show images in the \textit{None} category).

\begin{table}[]
\centering
\caption{Annotators' Agreement in Dimensional Model of Affect}
\label{Tab:AnnotatorsPerformance}
\vspace{-2mm}
\begin{tabular}{l|cc|cc|}
\cline{2-5}
                                    & \multicolumn{2}{c|}{Same Category}                         & \multicolumn{2}{c|}{All}                                   \\ \cline{2-5} 
                                    & \multicolumn{1}{l}{Valence} & \multicolumn{1}{l|}{Arousal} & \multicolumn{1}{l}{Valence} & \multicolumn{1}{l|}{Arousal} \\ \hline
\multicolumn{1}{|l|}{\textbf{RMSE}} & 0.190                       & 0.261                        & 0.340                       & 0.362                        \\ \hline
\multicolumn{1}{|l|}{\textbf{CORR}} & 0.951                       & 0.766                        & 0.823                       & 0.567                        \\ \hline
\multicolumn{1}{|l|}{\textbf{SAGR}} & 0.906                       & 0.709                        & 0.815                       & 0.667                        \\ \hline
\multicolumn{1}{|l|}{\textbf{CCC}}  & 0.951                       & 0.746                        & 0.821                       & 0.551                        \\ \hline
\end{tabular}
\end{table}

\begin{table*}[!htbp]
\centering
\caption{Agreement Between Two Annotators in Categorical Model of Affect (\%)}
\label{tab:AnnotatorsAgreement}
\vspace{-2mm}
\begin{tabular}{l|c|c|c|c|c|c|c|c|c|c|c|}
\cline{2-12}
\multicolumn{1}{c|}{}              & Neutral       & Happy         & Sad           & Surprise      & Fear          & Disgust       & Anger         & Contempt      & None		  & Uncertain     & Non-Face       \\ \hline
\multicolumn{1}{|l|}{Neutral}      & \textbf{50.8} & 7.0           & 9.1           & 2.8           & 1.1           & 1.0           & 4.8           & 5.3           & 11.1         & 1.9           & 5.1           \\ \hline
\multicolumn{1}{|l|}{Happy}        & 6.3           & \textbf{79.6} & 0.6           & 1.7           & 0.3           & 0.4           & 0.5           & 3.0           & 4.6          & 1.0           & 2.2           \\ \hline
\multicolumn{1}{|l|}{Sad}          & 11.8          & 0.9           & \textbf{69.7} & 1.2           & 3.4           & 1.3           & 4.0           & 0.3           & 3.5          & 1.2           & 2.6           \\ \hline
\multicolumn{1}{|l|}{Surprise}     & 2.0           & 3.8           & 1.6           & \textbf{66.5} & 14.0          & 0.8           & 1.9           & 0.6           & 4.2          & 1.9           & 2.7           \\ \hline
\multicolumn{1}{|l|}{Fear}         & 3.1           & 1.5           & 3.8           & 15.3          & \textbf{61.1} & 2.5           & 7.2           & 0.0           & 1.9          & 0.4           & 3.3           \\ \hline
\multicolumn{1}{|l|}{Disgust}      & 1.5           & 0.8           & 3.6           & 1.2           & 3.5           & \textbf{67.6} & 13.1          & 1.7           & 2.7          & 2.3           & 2.1           \\ \hline
\multicolumn{1}{|l|}{Anger}        & 8.1           & 1.2           & 7.5           & 1.7           & 2.9           & 4.4           & \textbf{62.3} & 1.3           & 5.5          & 1.9           & 3.3           \\ \hline
\multicolumn{1}{|l|}{Contempt}     & 10.2          & 7.5           & 2.1           & 0.5           & 0.5           & 4.4           & 2.1           & \textbf{66.9} & 3.7          & 1.5           & 0.6           \\ \hline
\multicolumn{1}{|l|}{None} 		   & \textbf{22.6} & 12.0          & 14.5          & 8.0           & 6.0           & 2.3           & 16.9          & 1.3           & 9.6          & 4.3           & 2.6           \\ \hline
\multicolumn{1}{|l|}{Uncertain}    & 13.5          & 12.1          & 7.8           & 7.3           & 4.0           & 4.5           & 6.2           & 2.6           & 12.3         & \textbf{20.6} & 8.9           \\ \hline
\multicolumn{1}{|l|}{Non-Face}      & 3.7           & 3.8           & 1.7           & 1.1           & 0.9           & 0.4           & 1.7           & 0.4           & 1.2          & 1.4           & \textbf{83.9} \\ \hline
\end{tabular}
\end{table*}

Table~\ref{Tab:AnnotatorsPerformance} shows various evaluation metrics between the two annotators in the continuous dimensional model of affect. These metrics are defined in Sec.~\ref{Sec:EvaluationMetrics}. We calculated these metrics in two scenarios: 1) the annotators agreed on the category of the image; 2) on all images that are annotated by two annotators. As Table~\ref{Tab:AnnotatorsPerformance} shows, when the annotators agreed on the category of the image, the annotations have a high correlation and sign agreement (SAGR). According to Table~\ref{tab:AnnotatorsAgreement}, this occurred on only 60.7\% images. However, there is less correlation and SAGR on overall images, since the annotators had a different perception of emotions expressed in the images. It can also be seen that the annotators agreed on valence more than arousal. The authors believe that this is because the perception of valence (how positive or negative the emotion is) is easier and less subjective than arousal (how excited or calm the subject is) especially in still images. Comparing the metrics in the existing dimensional databases (shown in Table~\ref{Tab:ExisitngAlgorithms}) with the agreement of human labelers on \textit{AffectNet}, suggest that \textit{AffectNet} is a very challenging database and even human annotations have more RMSE than automated methods on existing databases.


\section{Baseline}
\label{sec:baseline}
In this section, two baselines are proposed to classify images in the categorical model and predict the value of valence and arousal in the continuous domain of dimensional model. Since deep Convolutional Neural Networks (CNNs) have been a successful approach to learn appropriate feature abstractions directly from the image and there are many samples in \textit{AffectNet} necessary to train CNNs, we proposed two simple CNN baselines for both categorical and dimensional models. We also compared the proposed baselines with conventional approaches (Support Vector Machines~\cite{cortes1995support} and Support Vector Regressions~\cite{smola1997support}) learned from hand-crafted features (HOG). In the following sections, we first introduce our training, validation and test sets, and then show the performance of each proposed baselines.

\subsection{Test, Validation, and Training Sets}

\textbf{Test set: }The subset of the annotated images that are annotated by two annotators is reserved for the test set. To determine the value of valence and arousal in the test set, since there are two responses for one image in the continuous domain, one of the annotations is picked randomly. To select the category of image in the categorical model, if there was a disagreement, a favor was given to the intended query, i.e., if one of the annotators labeled the image as the intended query, the image was labeled with the intended query in the test set. This happened in 29.5\% of the images with disagreement between the annotators. On the rest of the images with disagreement, one of the annotations was assigned to the image randomly. Since the test set is a random sampling of all images, it is heavily imbalanced. In other words, there are more than 11,000 images with happy expression while it contains only 1,000 images with contemptuous expression. 

\textbf{Validation set: } Five hundred samples of each category is selected randomly as a validation set. The validation set is used for hyper-parameter tuning, and since it is balanced, there is no need for any skew normalization. 

\textbf{Training set:} The rest of images are considered as training examples. The training examples, as shown in Table~\ref{Tab:NumImages}, are heavily imbalanced.

\subsection{Categorical Model Baseline}

Facial expression data is usually highly skewed. This form of imbalance is commonly referred to as \emph{intrinsic} variation, i.e., it is a direct result of the nature of expressions in the real world. This happens in both the categorical and dimensional models of affect. For instance, Caridakis \textit{et al.}~\cite{caridakis2008user} reported that a bias toward quadrant 1 (positive arousal, positive valence) exists in the SAL database. The problem of learning from imbalanced data sets has two challenges. First, training data with an imbalanced distribution often causes learning algorithms to perform poorly on the minority class~\cite{he2009learning}. Second, the imbalance in the test/validation data distribution can affect the performance metrics dramatically. Jeni \textit{et al.}~\cite{jeni2013facing} studied the influence of skew on imbalanced validation set. The study showed that with exception of area under the ROC curve (AUC), all other studied evaluation metrics, i.e., Accuracy, F1-score, Cohen’s kappa~\cite{cohen1960}, Krippendorf’s Alpha~\cite{krippendorff1970estimating}, and area under Precision-Recall curve (AUC-PR) are affected by skewed distributions dramatically. While AUC is unaffected by skew, precision-recall curves suggested that AUC may mask poor performance. To avoid or minimize skew-biased estimates of performance, the study suggested to report both skew-normalized scores and the original evaluation. 

We used AlexNet~\cite{krizhevsky2012imagenet} architecture as our deep CNN baseline. AlexNet consists of five convolution layers, followed by max-pooling and normalization layers, and three fully-connected layers. To train our baseline with an imbalanced training set, four approaches are studied in this paper as \textit{Imbalanced learning}, \textit{Down-Sampling}, \textit{Up-Sampling}, and \textit{Weighted-Loss}. The imbalanced learning approach was trained with the imbalanced training set without any change in the skew of the dataset. To train the down-sampling approach, we selected a maximum of 15,000 samples from each class. Since there are less than 15,000 samples for some classes such as Disgust, Contempt, and Fear, the resulting training set is semi-balanced. To train the up-sampling approach, we heavily up-sampled the under-represented classes by replicating their samples so that all classes had the same number of samples as the class with maximum samples, i.e., Happy class. 
 
The weighted-loss approach weighted the loss function for each of the classes by their relative proportion in the training dataset. In other words, the loss function heavily penalizes the networks for misclassifying examples from under-represented classes, while penalizing networks less for misclassifying examples from well-represented classes. The entropy loss formulation for a training example $(X, l)$ is defined as:

\begin{table*}[!htbp]
\centering
\caption{F1-Scores of four different approaches of training AlexNet}
\label{Tab:F1-scores}
\vspace{-2mm}
\setlength\tabcolsep{4.2pt} 
\begin{tabular}{lcccccccccccccccc}
\cline{2-17} 
\multicolumn{1}{c?}{}          & \multicolumn{4}{c?}{Imbalanced}                                       & \multicolumn{4}{c?}{Down-Sampling}                                  & \multicolumn{4}{c?}{Up-Sampling}                                    & \multicolumn{4}{c|}{Weighted-Loss}                                  \\ \cline{2-17}
\multicolumn{1}{l?}{}          & \multicolumn{2}{c|}{Top-1}         & \multicolumn{2}{c?}{Top-2}       & \multicolumn{2}{c|}{Top-1}       & \multicolumn{2}{c?}{Top-2}       & \multicolumn{2}{c|}{Top-1}       & \multicolumn{2}{c?}{Top-2}       & \multicolumn{2}{c|}{Top-1}       & \multicolumn{2}{c|}{Top-2}       \\ \cline{2-17} 
\multicolumn{1}{l?}{}          & Orig* & \multicolumn{1}{c|}{Norm*} & Orig & \multicolumn{1}{c?}{Norm} & Orig & \multicolumn{1}{c|}{Norm} & Orig & \multicolumn{1}{c?}{Norm} & Orig & \multicolumn{1}{c|}{Norm} & Orig & \multicolumn{1}{c?}{Norm} & Orig & \multicolumn{1}{c|}{Norm} & Orig & \multicolumn{1}{c|}{Norm} \\ \hline
\multicolumn{1}{|l?}{Neutral}  & 0.63  & \multicolumn{1}{c|}{0.49}  & 0.82 & \multicolumn{1}{c?}{0.66} & 0.58 & \multicolumn{1}{c|}{0.49} & 0.78 & \multicolumn{1}{c?}{0.70} & 0.61 & \multicolumn{1}{c|}{0.50} & 0.81 & \multicolumn{1}{c?}{0.64} & 0.57 & \multicolumn{1}{c|}{0.52} & 0.81 & \multicolumn{1}{c|}{0.77} \\ \hline
\multicolumn{1}{|l?}{Happy}    & 0.88  & \multicolumn{1}{c|}{0.65}  & 0.95 & \multicolumn{1}{c?}{0.80} & 0.85 & \multicolumn{1}{c|}{0.68} & 0.92 & \multicolumn{1}{c?}{0.85} & 0.85 & \multicolumn{1}{c|}{0.71} & 0.95 & \multicolumn{1}{c?}{0.80} & 0.82 & \multicolumn{1}{c|}{0.73} & 0.92 & \multicolumn{1}{c|}{0.88} \\ \hline
\multicolumn{1}{|l?}{Sad}      & 0.63  & \multicolumn{1}{c|}{0.60}  & 0.84 & \multicolumn{1}{c?}{0.81} & 0.64 & \multicolumn{1}{c|}{0.60} & 0.81 & \multicolumn{1}{c?}{0.78} & 0.6  & \multicolumn{1}{c|}{0.57} & 0.81 & \multicolumn{1}{c?}{0.77} & 0.63 & \multicolumn{1}{c|}{0.61} & 0.83 & \multicolumn{1}{c|}{0.81} \\ \hline
\multicolumn{1}{|l?}{Surprise} & 0.61  & \multicolumn{1}{c|}{0.64}  & 0.84 & \multicolumn{1}{c?}{0.86} & 0.53 & \multicolumn{1}{c|}{0.63} & 0.75 & \multicolumn{1}{c?}{0.83} & 0.57 & \multicolumn{1}{c|}{0.66} & 0.80 & \multicolumn{1}{c?}{0.81} & 0.51 & \multicolumn{1}{c|}{0.63} & 0.77 & \multicolumn{1}{c|}{0.86} \\ \hline
\multicolumn{1}{|l?}{Fear}     & 0.52  & \multicolumn{1}{c|}{0.54}  & 0.78 & \multicolumn{1}{c?}{0.79} & 0.54 & \multicolumn{1}{c|}{0.57} & 0.80 & \multicolumn{1}{c?}{0.82} & 0.56 & \multicolumn{1}{c|}{0.58} & 0.75 & \multicolumn{1}{c?}{0.76} & 0.56 & \multicolumn{1}{c|}{0.66} & 0.79 & \multicolumn{1}{c|}{0.86} \\ \hline
\multicolumn{1}{|l?}{Disgust}  & 0.52  & \multicolumn{1}{c|}{0.55}  & 0.76 & \multicolumn{1}{c?}{0.78} & 0.53 & \multicolumn{1}{c|}{0.64} & 0.74 & \multicolumn{1}{c?}{0.81} & 0.53 & \multicolumn{1}{c|}{0.59} & 0.70 & \multicolumn{1}{c?}{0.72} & 0.48 & \multicolumn{1}{c|}{0.66} & 0.69 & \multicolumn{1}{c|}{0.83} \\ \hline
\multicolumn{1}{|l?}{Anger}    & 0.65  & \multicolumn{1}{c|}{0.59}  & 0.83 & \multicolumn{1}{c?}{0.80} & 0.62 & \multicolumn{1}{c|}{0.60} & 0.79 & \multicolumn{1}{c?}{0.78} & 0.63 & \multicolumn{1}{c|}{0.59} & 0.81 & \multicolumn{1}{c?}{0.77} & 0.60 & \multicolumn{1}{c|}{0.60} & 0.81 & \multicolumn{1}{c|}{0.81} \\ \hline
\multicolumn{1}{|l?}{Contempt} & 0.08  & \multicolumn{1}{c|}{0.08}  & 0.49 & \multicolumn{1}{c?}{0.49} & 0.22 & \multicolumn{1}{c|}{0.32} & 0.60 & \multicolumn{1}{c?}{0.70} & 0.15 & \multicolumn{1}{c|}{0.18} & 0.42 & \multicolumn{1}{c?}{0.42} & 0.27 & \multicolumn{1}{c|}{0.59} & 0.58 & \multicolumn{1}{c|}{0.79} \\ \hline
\multicolumn{17}{l}{\scriptsize{*Orig and Norm stand for \textbf{Orig}inal and skew-\textbf{Norma}lized, respectively.}}                                                                                                                                                                                                                               
\end{tabular}
\vspace{-3mm}
\end{table*}

\begin{equation}
\label{eq:infoGainLoss}
E = - \sum_{i=1}^{K}{H_{l,i} log(\hat{p_i})}
\end{equation}
where $H_{l,i}$ denotes row $l$ penalization factor of class $i$, $K$ is the number of classes, and $\hat{p_i}$ is the 
predictive softmax with values $[0,1]$ indicating the predicted probability of each class as:
\begin{equation}
\hat{p_i} = \frac{exp(x_i)}{\sum_{j=1}^{K}{exp(x_j)}}
\end{equation}

Equation~\eqref{eq:infoGainLoss} can be re-written as:
\begin{equation}
\begin{aligned}
E ={} 	& - \sum_{i}{H_{l,i} log(\frac{exp(x_i)}{\sum_j{exp(x_j)}})} \\
& = - \sum_{i}{ H_{l,i} x_i} + \sum_{i}{H_{l,i} log (\sum_{j} exp(x_j))}\\
& = log(\sum_{j}{exp(x_j)}) \sum_{i}{H_{l,i}} - \sum_{i}{ H_{l,i} x_i} 
\end{aligned}
\end{equation}

The derivate with respect to the prediction $x_k$ is:
\begin{equation}
\begin{aligned}
\frac{\partial E}{\partial x_k} ={} 	& \frac{\partial}{\partial x_k} [log(\sum_{j}{exp(x_j)}) \sum_{i}{H_{l,i}}] - 
\frac{\partial}{\partial x_k} [\sum_{i}{ H_{l,i} x_i}]\\
& = (\sum_{i}{H_{l,i}}) \frac{1}{\sum_{j}{exp(x_j)}} \frac{\partial}{\partial x_k} \sum_{j}{exp(x_j)} - H_{l,k}\\
& = (\sum_{i}{H_{l,i}}) \frac{exp(x_k)}{\sum_{j}{exp(x_j)}} - H_{l,k}\\
& = (\sum_{i}{H_{l,i}}) \hat{p_k} - H_{l,k} 
\end{aligned}
\end{equation}

When $H = I$, the identity, the proposed weighted-loss approach gives the traditional cross-entropy loss function. We used the implemented Infogain loss in Caffe~\cite{jia2014caffe} for this purpose. For simplicity, we used a diagonal matrix defined as:
\begin{equation}
\label{eq:H_Matrix}
H_{ij} = 
\begin{cases}
\frac{f_i}{f_{min}},				& \text{if } i = j\\
0,                   		& \text{otherwise}
\end{cases}
\end{equation}
where $f_i$ is the number of samples of the $i^{\text{th}}$ class and $f_{min}$ is the number of samples in the most under-represented class, i.e., Disgust class in this situation.

Before training the network, the faces were cropped and resized to 256$\times$256 pixels. No facial registration was performed at this baseline. To augment the data, five crops of 224$\times$224 and their horizontal flips were extracted from the four corners and the center of the image at random during the training phase. The networks were trained for 20 epochs using a batch size of 256. The base learning rate was set to 0.01, and decreased step-wise by a factor of 0.1 every 10,000 iterations. We used a momentum of 0.9.

Table~\ref{Tab:F1-scores} shows the top-1 and top-2 F1-Scores for the imbalanced learning, down-sampling, up-sampling, and weighted-loss approaches on the test set. Since the test set is imbalanced, both the skew-normalized and the original scores are reported. The skew normalization is performed by random under-sampling of the classes in the test set. This process is repeated 200 times, and the skew-normalized score is the average of the score on multiple trials. As it is shown, the weighted-loss approach performed better than other approaches in the skew-normalized fashion. The improvement is significant in under-represented classes, i.e., Contempt, Fear, and Disgust. The imbalanced approach performed worst in the Contempt and Disgust categories since there were a few training samples of these classes compared with other classes. The up-sampling approach also did not classify the Contempt and Disgust categories well, since the training samples of these classes were heavily up-sampled (almost 20 times), and the network was over-fitted to these samples. Hence the network lost its generalization and performed poorly on these classes of the test set. 

The confusion matrix of the weighted-loss approaches is shown in Table~\ref{Tab:ConfusionMatrixWeightedLoss}. The weighted-loss approach classified the samples of Contempt and Disgust categories with an acceptable accuracy but did not perform well in Happy and Neutral. This is because the network was not penalized enough for misclassifying examples from these classes. We believe that a better formulation of the weight matrix $H$ based on the number of samples in the mini-batches or other data-driven approaches can improve the recognition of well-represented classes.

\begin{table}[!htbp]
\centering
\caption{Confusion Matrix of Weighted-Loss Approach on the Test Set}
\label{Tab:ConfusionMatrixWeightedLoss}
\vspace{-2mm}
\setlength\tabcolsep{5.5pt} 
\begin{tabular}{lc|c|c|c|c|c|c|c|c|}
\cline{3-10}
                                              & \multicolumn{1}{l|}{} & \multicolumn{8}{c|}{Predicted}                                                                                                \\ \cline{3-10} 
                                              &                       & \textbf{NE}   & \textbf{HA}   & \textbf{SA}   & \textbf{SU}   & \textbf{FE}   & \textbf{DI}   & \textbf{AN}   & \textbf{CO}   \\ \hline
\multicolumn{1}{|l|}{\multirow{8}{*}{\begin{turn}{+90}Actual\end{turn}}} & \textbf{NE}           & \textbf{53.3} & 2.8           & 9.8           & 8.7           & 1.7           & 2.5           & 10.4          & 10.9          \\ \cline{2-10} 
\multicolumn{1}{|l|}{}                        & \textbf{HA}           & 4.5           & \textbf{72.8} & 1.1           & 6.0           & 0.6           & 1.7           & 1.0           & 12.2          \\ \cline{2-10} 
\multicolumn{1}{|l|}{}                        & \textbf{SA}           & 13.0          & 1.3           & \textbf{61.7} & 3.6           & 5.8           & 4.4           & 9.2           & 1.2           \\ \cline{2-10} 
\multicolumn{1}{|l|}{}                        & \textbf{SU}           & 3.4           & 1.2           & 1.7           & \textbf{69.9} & 18.9          & 1.7           & 2.8           & 0.5           \\ \cline{2-10} 
\multicolumn{1}{|l|}{}                        & \textbf{FE}           & 1.5           & 1.5           & 4.6           & 13.5          & \textbf{70.4} & 4.2           & 4.3           & 0.2           \\ \cline{2-10} 
\multicolumn{1}{|l|}{}                        & \textbf{DI}           & 2.0           & 2.2           & 5.8           & 3.3           & 6.2           & \textbf{68.6} & 10.6          & 1.3           \\ \cline{2-10} 
\multicolumn{1}{|l|}{}                        & \textbf{AN}           & 6.2           & 1.2           & 5.0           & 3.2           & 5.8           & 11.1          & \textbf{65.8} & 1.9           \\ \cline{2-10} 
\multicolumn{1}{|l|}{}                        & \textbf{CO}           & 16.2          & 13.1          & 3.5           & 3.1           & 0.5           & 4.3           & 5.7           & \textbf{53.8} \\ \hline
\end{tabular}
\end{table}

Table~\ref{Tab:EvaluationResult} shows accuracy, F1-score, Cohen’s kappa, Krippendorf’s Alpha, area under the ROC curve (AUC), and area under the Precision-Recall curve (AUC-PR) on the test sets. Except for the accuracy, all the metrics are calculated in a binary-class manner where the positive class contains the samples labeled by the given category, and the negative class contains the rest. The reported result in Table~\ref{Tab:EvaluationResult} is the average of these metrics over eight classes. The accuracy is defined in a multi-class manner in which the number of correct predictions is divided by the total number of samples in the test set. The skew-normalization is performed by balancing the distribution of classes in the test set using random under-sampling and averaging over 200 trials. Since the validation set is balanced, there is no need for skew-normalization. 

\begin{table*}[!htbp]
\centering
\caption{Evaluation Metrics and Comparison of CNN baselines, SVM and MS Cognitive on Categorical Model of Affect.}
\label{Tab:EvaluationResult}
\vspace{-2mm}
\begin{tabular}{l?cc|cc|cc|cc?cc?cc|}
	\cline{2-13}
										 & \multicolumn{8}{c?}{\textbf{CNN Baselines}}                                                                                                                                      & \multicolumn{2}{c?}{\multirow{2}{*}{\textbf{SVM}}} & \multicolumn{2}{c|}{\multirow{2}{*}{\textbf{MS Cognitive}}} \\ \cline{2-9}
										 & \multicolumn{2}{c|}{\textbf{Imbalanced}} & \multicolumn{2}{c|}{\textbf{Down-Sampling}} & \multicolumn{2}{c|}{\textbf{Up-Sampling}} & \multicolumn{2}{c?}{\textbf{Weighted-Loss}} & \multicolumn{2}{c?}{}                              & \multicolumn{2}{c|}{}                                    \\ \cline{2-13} 
										 & Orig				   & Norm               & Orig                 & Norm                 & Orig                & Norm                & Orig                 & Norm                 & Orig                     & Norm                    & Orig                        & Norm                       \\ \hline
	\multicolumn{1}{|l?}{\textbf{Accuracy}}   & 0.72                & 0.54               & 0.68                 & 0.58                 & 0.68                & 0.57                & 0.64                 & 0.63                 & 0.60                     & 0.37                    & 0.68                        & 0.48                       \\ \hline
	\multicolumn{1}{|l?}{\textbf{F$_1$-Score}}    & 0.57                & 0.52               & 0.56                 & 0.57                 & 0.56                & 0.55                & 0.55                 & 0.62                 & 0.37                     & 0.31                    & 0.51                        & 0.45                       \\ \hline
	\multicolumn{1}{|l?}{\textbf{Kappa}} & 0.53                & 0.46               & 0.51                 & 0.51                 & 0.52                & 0.49                & 0.5                  & 0.57                 & 0.32                     & 0.25                    & 0.46                        & 0.40                       \\ \hline
	\multicolumn{1}{|l?}{\textbf{Alpha}} & 0.52                & 0.45               & 0.51                 & 0.51                 & 0.51                & 0.48                & 0.5                  & 0.57                 & 0.31                     & 0.22                    & 0.46                        & 0.37                       \\ \hline
	\multicolumn{1}{|l?}{\textbf{AUC}}   & 0.85                & 0.80               & 0.82                 & 0.85                 & 0.82                & 0.84                & 0.86                 & 0.86                 & 0.77                        & 0.70                       & 0.83                        & 0.77                       \\ \hline
	\multicolumn{1}{|l?}{\textbf{AUCPR}} & 0.56                & 0.55               & 0.54                 & 0.57                 & 0.55                & 0.56                & 0.58                 & 0.64                 & 0.39                        & 0.37                       & 0.52                        & 0.50                       \\ \hline
\end{tabular}
\end{table*}

We compared the performance of CNN baseline with a Support Vector Machine (SVM)~\cite{cortes1995support}. To train SVM, the faces in the images were cropped and resized to 256$\times$256 pixels. HOG~\cite{dalal2005histograms} features were extracted with the cell size of 8. We applied PCA retaining 95\% of the variance to reduce the HOG features dimensionality from 36,864 to 6,697 features. We used a linear kernel SVM in Liblinear package~\cite{fan2008liblinear} (which is optimized for large-scale linear classification and regression). Table~\ref{Tab:EvaluationResult} shows the evaluation metrics of SVM. 
Reported AUC and AUCPR values for SVM are calculated using the LibLinear's resulting decision values. We calculated the scores of predictions  using a posterior-probability transformation sigmoid function.
Comparing the performance of SVM with the CNN baselines on AffectNet, indicates that CNN models perform better than conventional SVM and HOG features in all metrics. 

We also compared the baseline with an available off-the-shelf expression recognition system (Microsoft Cognitive Services emotion API~\cite{MicrosofCognitiveServices}). The MS cognitive system had an excellent performance on Neutral and Happy categories with an accuracy of 0.94 and 0.85, respectively. However, it performed poorly on other classes with an accuracy of 0.25, 0.27 and 0.04 in the Fear, Disgust and Contempt categories. Table~\ref{Tab:EvaluationResult} shows the evaluation metrics on the MS cognitive system. Comparing the performance of the MS cognitive with the simple baselines on AffectNet indicates that \textit{AffectNet} is a challenging database and a great resource to further improve the performance of facial expression recognition systems. 

\begin{figure}
\centering
\subfloat
{
	\centering
	\stackunder{\includegraphics[width=20mm]{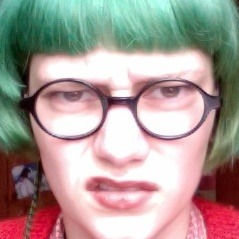}}{\scriptsize{\emph{Angry} (Disgust)}}%
}
\subfloat
{
	\centering
	\stackunder{\includegraphics[width=20mm]{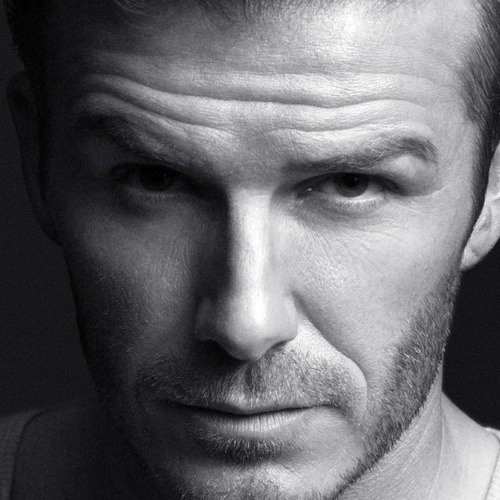}}{\scriptsize{\emph{Disgust} (Angry)}}%
}
\subfloat
{
	\centering
	\stackunder{\includegraphics[width=20mm]{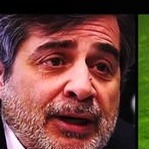}}{\scriptsize{\emph{Fear} (Sad)}}%
}
\subfloat
{
	\centering
	\stackunder{\includegraphics[width=20mm]{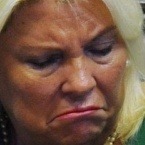}}{\scriptsize{\emph{Angry} (Sad)}}%
}
\vspace{-0.2cm}
\subfloat
{
	\centering
	\stackunder{\includegraphics[width=20mm]{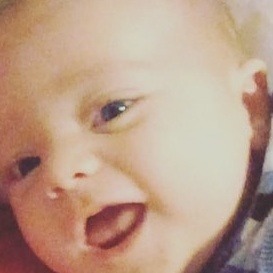}}{\scriptsize{\emph{Happy} (Surprise)}}%
}
\subfloat
{
	\centering
	\stackunder{\includegraphics[width=20mm]{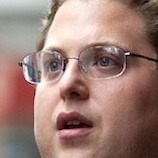}}{\scriptsize{\emph{Fear} (Surprise)}}%
}
\subfloat
{
	\centering
	\stackunder{\includegraphics[width=20mm]{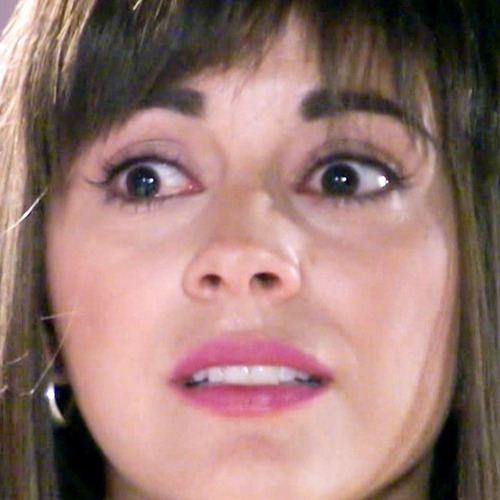}}{\scriptsize{\emph{Surprise} (Fear)}}%
}
\subfloat
{
	\centering
	\stackunder{\includegraphics[width=20mm]{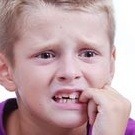}}{\scriptsize{\emph{Angry} (Fear)}}%
}
\vspace{-0.2cm}
\subfloat
{
	\centering
	\stackunder{\includegraphics[width=20mm]{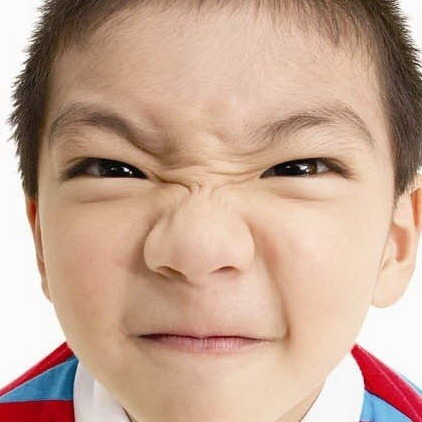}}{\scriptsize{\emph{Angry} (Disgust)}}%
}
\subfloat
{
	\centering
	\stackunder{\includegraphics[width=20mm]{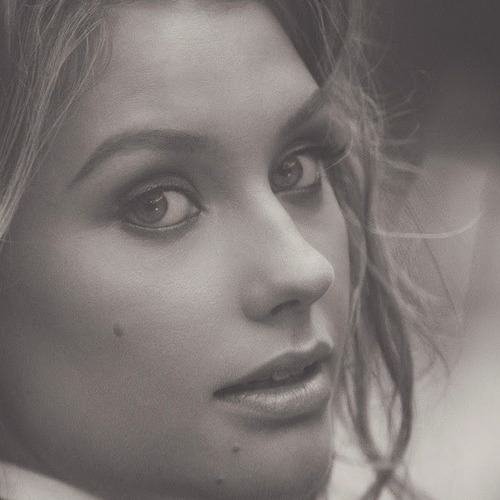}}{\scriptsize{\emph{Happy} (Neutral)}}%
}
\subfloat
{
	\centering
	\stackunder{\includegraphics[width=20mm]{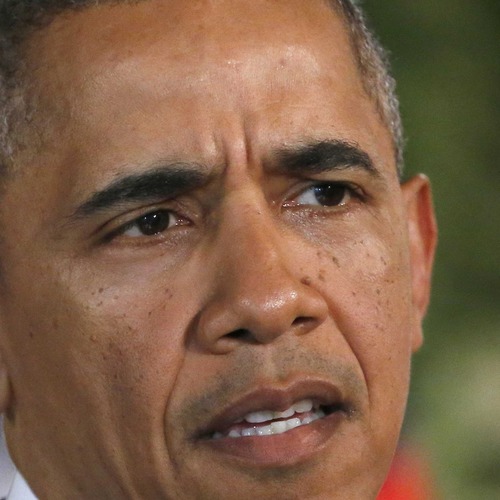}}{\scriptsize{\emph{Sad} (Angry)}}%
}
\subfloat
{
	\centering
	\stackunder{\includegraphics[width=20mm]{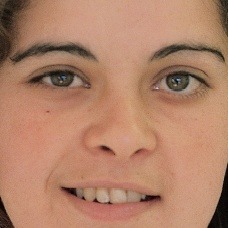}}{\scriptsize{\emph{Happy} (Contempt)}}%
}
\caption{\label{fig:Sample_MissClassified}
Samples of miss-classified images. Their corresponding ground-truth is given in parentheses.
}
\end{figure}

Figure~\ref{fig:Sample_MissClassified} shows nine samples of randomly selected misclassified images of the weighted-loss approach and their corresponding ground-truth. As the figure shows, it is really difficult to assign some of the emotions to a single category. Some of the faces have partial similarities in facial features to the misclassified images, such as nose wrinkled in disgust, or eyebrows raised in surprise. This emphasizes the fact that classifying facial expressions in the wild is a challenging task and, as mentioned before, even human annotators agreed on only 60.7\%  of the images.

\subsection{Dimensional Model (Valence and Arousal) Baseline}


Predicting dimensional model in the continuous domain is a real-valued regression problem. We used AlexNet~\cite{krizhevsky2012imagenet} architecture as our deep CNN baseline to predict the value of valence and arousal. Particularly, two separate AlexNets were trained where the last fully-connected layer was replaced with a linear regression layer containing only one neuron. The output of the neuron predicted the value of valence/arousal in continuous domain [-1,1]. A Euclidean (L2) loss was used to measure the distance between the predicted value ($\hat{y}_n$) and actual value of valence/arousal ($y_n$) as:
\begin{equation}
E = \frac{1}{2N}\sum_{n=1}^{N}{||\hat{y}_n-y_n||_2^2}
\end{equation}

\begin{figure}
	\centering
	\includegraphics[width=3.2in]{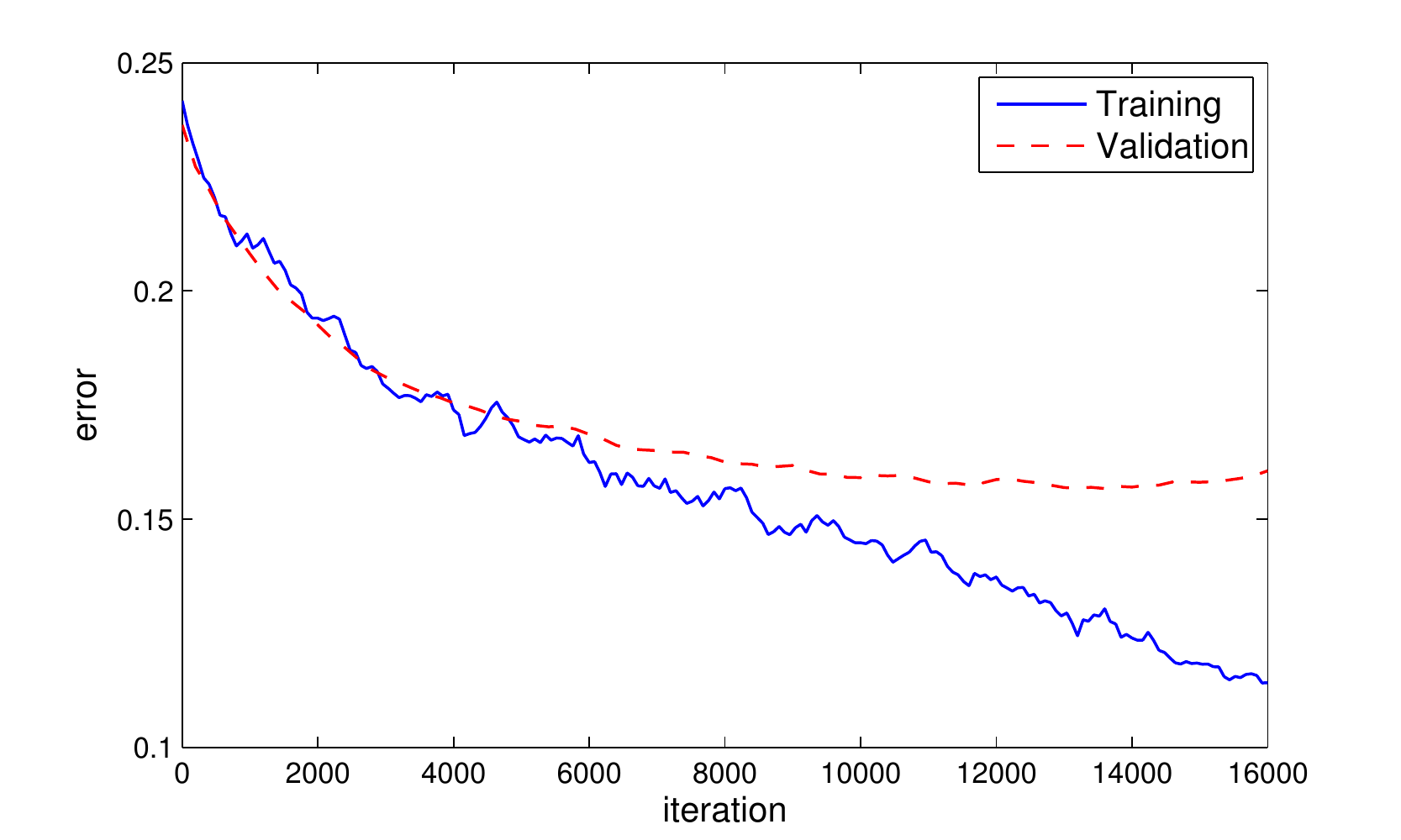}
	\vspace{-6mm}
	\caption{Euclidean error of training valence and arousal.}
	\label{fig:valenceArousalloss}
\end{figure}

The faces were cropped and resized to 256$\times$256 pixels. The base learning rate was fixed and set to 0.001 during the training process. We used a momentum of 0.9. Training was continued until a plateau was reached in the Euclidean error of the validation set (approximately 16 epochs with a mini-batch size of 256). Figure~\ref{fig:valenceArousalloss} shows the value of training and validation losses over 16K iterations (about 16 epochs).

We also compared Support Vector Regression (SVR)~\cite{smola1997support} with our DNN baseline for predicting valence and arousal in \textit{AffectNet}. In our experiments, first, the faces in the images were cropped and resized to 256$\times$256 pixels. Histogram of Oriented Gradient (HOG)~\cite{dalal2005histograms} features were extracted with the cell size of 8. Afterward, we applied PCA retaining 95\% of the variance of these features to reduce the dimensionality. Two separate SVRs were trained to predict the value of valence and arousal. Liblinear~\cite{fan2008liblinear} package was used to implement SVR baseline.

Table~\ref{Tab:ValenceArousalBaseline} shows the performances of the proposed baseline and SVR on the test set. As shown, the CNN baseline can predict the value of valence and arousal better than SVR. This is because the high variety of samples in \textit{AffectNet} allows the CNN to extract more discriminative features than hand-crafted HOG, and therefore it learned a better representation of dimensional affect.
 
\begin{table}[]
\centering
\caption{Baselines' Performances of Predicting Valence and Arousal on Test Set}
\label{Tab:ValenceArousalBaseline}
\vspace{-2mm}
\begin{tabular}{l|cc|cc|}
\cline{2-5}
                                    & \multicolumn{2}{c|}{CNN (AlexNet)}                         & \multicolumn{2}{c|}{SVR}                                   \\ \cline{2-5} 
                                    & \multicolumn{1}{l}{Valence} & \multicolumn{1}{l|}{Arousal} & \multicolumn{1}{l}{Valence} & \multicolumn{1}{l|}{Arousal} \\ \hline
\multicolumn{1}{|l|}{\textbf{RMSE}} & 0.394                       & 0.402                        & 0.494                       & 0.400                        \\ \hline
\multicolumn{1}{|l|}{\textbf{CORR}} & 0.602                       & 0.539                        & 0.429                       & 0.360                        \\ \hline
\multicolumn{1}{|l|}{\textbf{SAGR}} & 0.728                       & 0.670                        & 0.619                       & 0.748                        \\ \hline
\multicolumn{1}{|l|}{\textbf{CCC}}  & 0.541                       & 0.450                        & 0.340                       & 0.199                        \\ \hline
\end{tabular}
\end{table}

\begin{figure}
	\centering
	\includegraphics[width=\columnwidth]{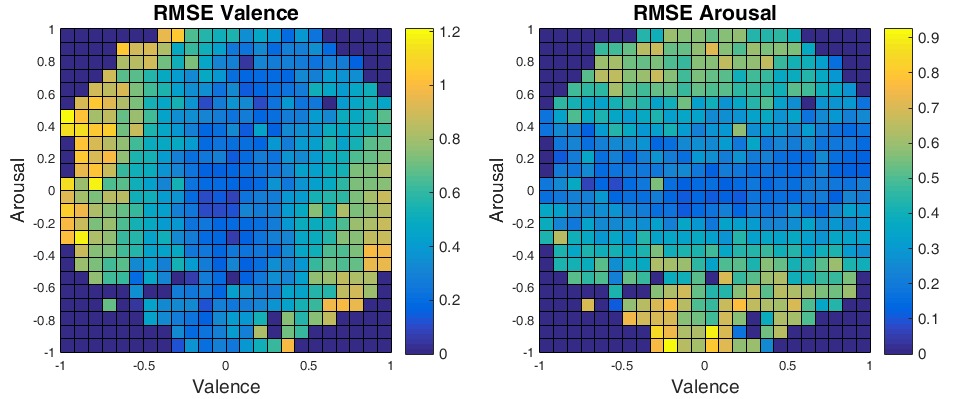}
	\vspace{-6mm}
	\caption{RMSE of predicted valence and arousal using AlexNet and Euclidean (L2) loss (Best viewed in color).}
	\label{fig:valenceArousalerrorPlot}
\end{figure}

The RMSE of CNN baseline (AlexNet) between the predicted valence and arousal and the ground-truth are shown in Fig.~\ref{fig:valenceArousalerrorPlot}. As illustrated, the CNN baseline has a lower error rate in the center of circumplex. In particular, predicting low-valence mid-arousal and low-arousal mid-valence areas were more challenging. These areas correspond to the expressions of contempt, bored, and sleepy. It should be mentioned that predicting valence and arousal in the wild is a challenging task, and as discussed in Sec.~\ref{Sec:AnnotationAgreement}, the disagreement between two human annotators has RMSE=0.367 and RMSE=0.481 for valence and arousal, respectively.

\section{Conclusion}
\label{sec:conclusion}

The analysis of human facial behavior is a very complex and challenging problem. The majority of the techniques for automated facial affect analysis are mainly based on machine learning methodologies, and their performance highly depends on the amount and diversity of annotated training samples. Recently, databases of facial expression and affect in the wild received much attention. However, existing databases of facial affect in the wild only cover one model of affect, have a limited number of subjects, or contain few samples of certain emotions.

The Internet is a vast source of facial images, most of which are captured in uncontrolled conditions. These images are often taken in the wild under natural conditions. In this paper, we introduced a new publicly available database of a facial \textbf{Affect} from the Inter\textbf{Net} (called \textit{AffectNet}) by querying different search engines using emotion related tags in six different languages. \textit{AffectNet} contains more than 1M images with faces and extracted landmark points. Twelve human experts manually annotated 450,000 of these images in both the categorical and dimensional (valence and arousal) models and tagged the images that have any occlusion on the face.

The agreement level of human labelers on a subset of \textit{AffectNet} showed that expression recognition and predicting valence and arousal in the wild is a challenging task. The two annotators agreed on 60.7\% of the category of facial expressions, and there was a large disagreement on the value of valence and arousal (RMSE=0.34 and 0.36) between the two annotators.

Two simple deep neural network baselines were examined to classify the facial expression images and predict the value of valence and arousal in the continuous domain of dimensional model. Evaluation metrics showed that simple deep neural network baselines trained on \textit{AffectNet} can perform better than conventional machine learning methods and available off-the-shelf expression recognition systems. \textit{AffectNet} is by far the largest database of facial expression, valence and arousal in the wild, enabling further progress in the automatic understanding of facial behavior in both categorical and continuous dimensional space. The interested investigators can study categorical and dimensional models in the same corpus, and possibly co-train them to improve the performance of their affective computing systems. It is highly anticipated that the availability of this database for the research community, along with the recent advances in deep neural networks, can improve the performance of automated affective computing systems in recognizing facial expressions and predicting valence and arousal.


%

\ifCLASSOPTIONcompsoc
  \section*{Acknowledgments}
This work is partially supported by the NSF grants IIS-1111568 and CNS-1427872. We gratefully acknowledge the support of NVIDIA Corporation with the donation of the Tesla K40 GPUs used for this research.
\else
  \section*{Acknowledgment}
This work is partially supported by the NSF grants IIS-1111568 and CNS-1427872. We gratefully acknowledge the support of NVIDIA Corporation with the donation of the Tesla K40 GPUs used for this research.  
\fi

\ifCLASSOPTIONcaptionsoff
  \newpage
\fi

\begin{IEEEbiography}[{\includegraphics[width=1in,height=1.25in,clip,keepaspectratio]{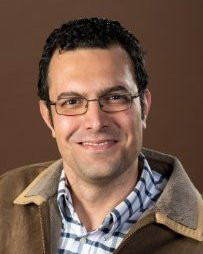}}]{Ali Mollahosseini}
received the BSc degree in computer software engineering from the Iran University of Science and Technology, Iran, in 2006, and the MSc degree in computer engineering - artificial intelligence from AmirKabir University, Iran, in 2010. He is currently working toward the Ph.D. degree and is a graduate research assistant in the Department of Electrical and Computer Engineering at the University of Denver. His research interests include deep neural networks for the analysis of facial expression, developing humanoid social robots and computer vision.
  
\end{IEEEbiography}
\vspace{-1cm}
\begin{IEEEbiography}[{\includegraphics[width=1in,height=1.25in,clip,keepaspectratio]{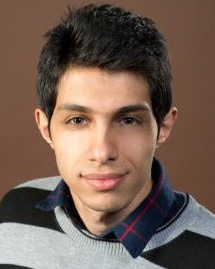}}]{Behzad Hasani}
	received the BSc degree in computer hardware engineering from Khaje Nasir Toosi University of Technology, Tehran, Iran, in 2013, and the MSc degree in computer engineering - artificial intelligence from Iran University of Science and Technology, Tehran, Iran, in 2015. He is currently pursuing his Ph.D. degree in electrical \& computer engineering and is a graduate research assistant in the Department of Electrical and Computer Engineering at the University of Denver. His research interests include Computer Vision, Machine Learning, and Deep Neural Networks, especially on facial expression analysis.
\end{IEEEbiography}
\vspace{-1cm}
\begin{IEEEbiography}[{\includegraphics[width=1in,height=1.25in,clip,keepaspectratio]{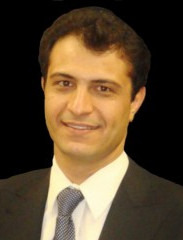}}]{Mohammad H. Mahoor}
	received the BS degree in electronics from the Abadan Institute of	Technology, Iran, in 1996, the MS degree in	biomedical engineering from the Sharif University of Technology, Iran, in 1998, and the Ph.D. degree in electrical and computer engineering from the University of Miami, Florida, in 2007. He is an Associate Professor of Electrical and Computer Engineering at DU. He does research in the area of computer vision and machine learning including visual object recognition, object tracking and pose estimation, motion estimation, 3D reconstruction, and human-robot interaction (HRI) such as humanoid social robots for interaction and intervention with children with special needs (e.g., autism) and elderly with depression and dementia. He has received over \$3M of research funding from state and federal agencies including the National Science Foundation. He is a Senior Member of IEEE and has published about 100 conference and journal papers. 
\end{IEEEbiography}


\vfill


\newpage

\appendices

\onecolumn
\section{}

\begin{table}[H]
	\centering
	\caption{Samples of Annotated Categories for Queried Emotion Terms}
	\label{Tab:annotationConfusionMatrix_appendix}
	\vspace{-2mm}
\begin{tabular}{ccccccccc}
	\cline{3-9}
	& \multicolumn{1}{c|}{}   & \multicolumn{7}{c|}{Queried Expression}                                                                                                                                                                                                                          \\ \cline{3-9} 
	& \multicolumn{1}{c|}{}   & \multicolumn{1}{c|}{Happy}            & \multicolumn{1}{c|}{Sad}            & \multicolumn{1}{c|}{Surprise}            & \multicolumn{1}{c|}{Fear}            & \multicolumn{1}{c|}{Disgust}          & \multicolumn{1}{c|}{Anger}            & \multicolumn{1}{c|}{Contempt}            \\ \hline
	\multicolumn{1}{|c|}{\multirow{11}{*}{\begin{turn}{+90}Annotated Expression\end{turn}}} & \multicolumn{1}{c|}{\raisebox{0.5cm}{Neutral}} & \multicolumn{1}{c|}{ 
			\includegraphics[width=1.2cm]{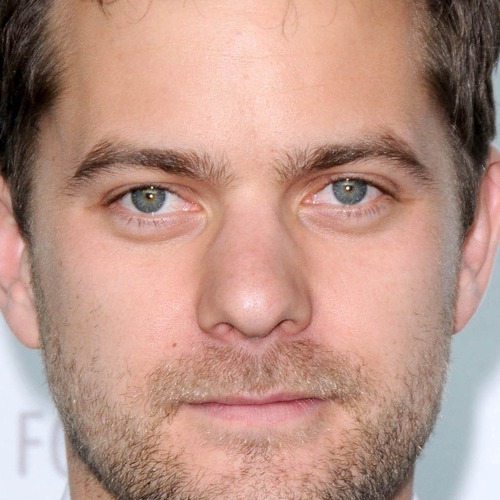}
	 }          & \multicolumn{1}{c|}{\includegraphics[width=1.2cm]{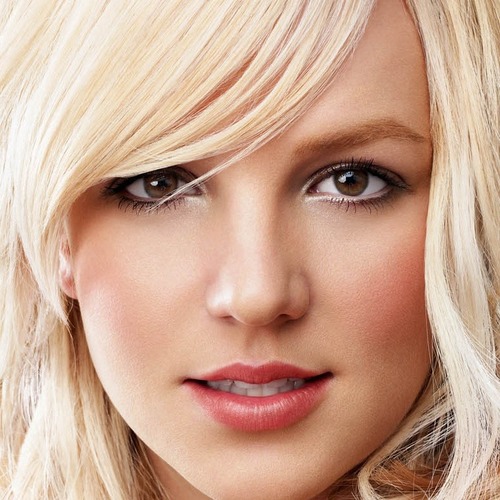}}          & \multicolumn{1}{c|}{\includegraphics[width=1.2cm]{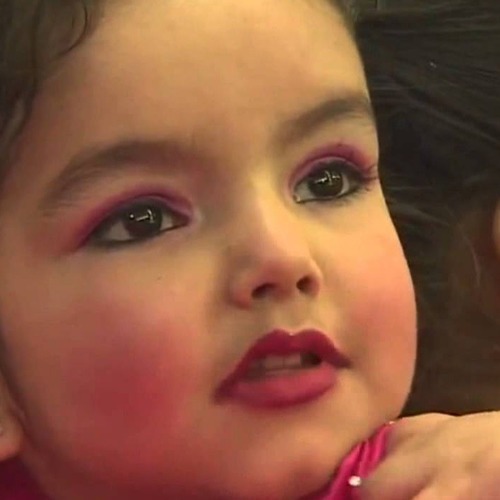}}          & \multicolumn{1}{c|}{\includegraphics[width=1.2cm]{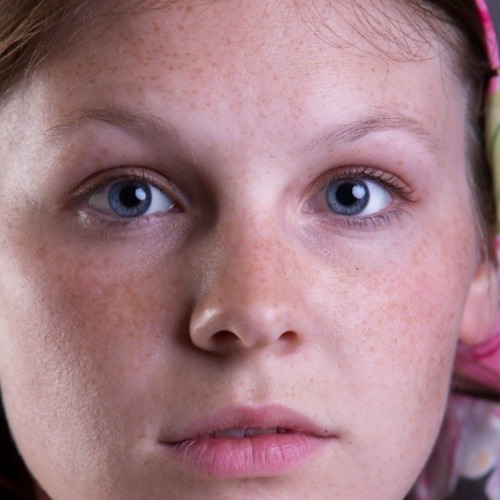}}          & \multicolumn{1}{c|}{\includegraphics[width=1.2cm]{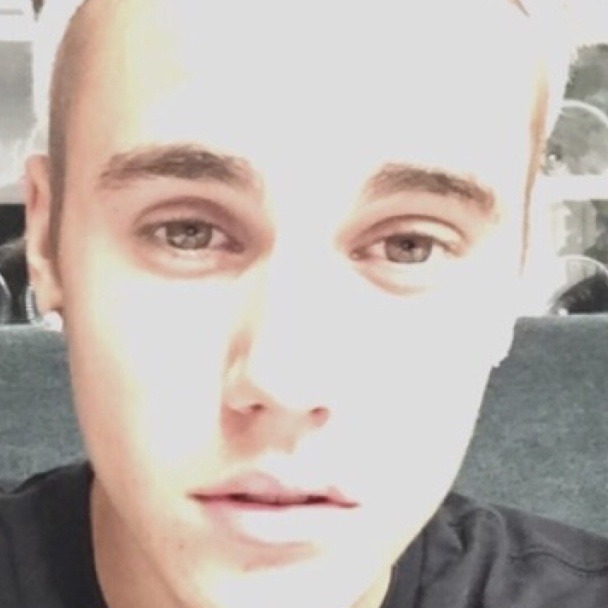}}        & \multicolumn{1}{c|}{\includegraphics[width=1.2cm]{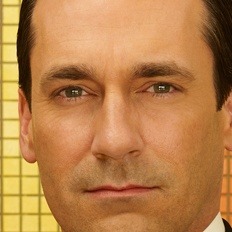}}          & \multicolumn{1}{c|}{\includegraphics[width=1.2cm]{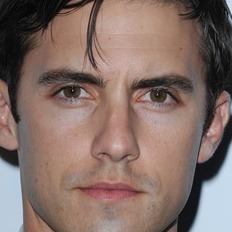}}          \\ \cline{2-9} 
	\multicolumn{1}{|c|}{}                                       & \multicolumn{1}{c|}{\raisebox{0.5cm}{Happy}} & \multicolumn{1}{c|}{\includegraphics[width=1.2cm]{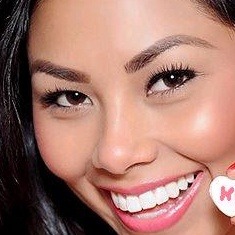}} & \multicolumn{1}{c|}{\textbf{\includegraphics[width=1.2cm]{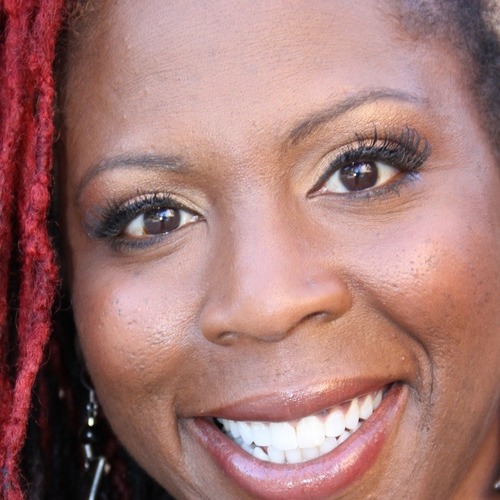}}} & \multicolumn{1}{c|}{\includegraphics[width=1.2cm]{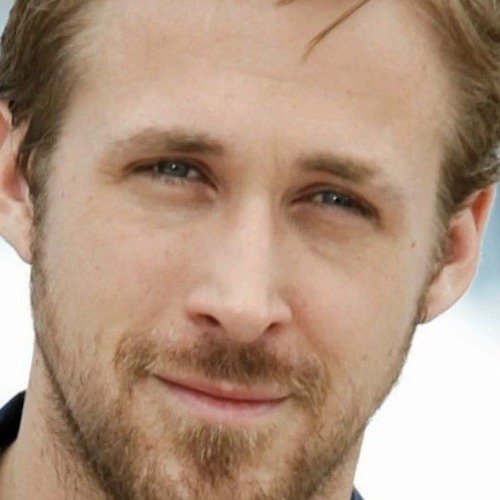}} & \multicolumn{1}{c|}{\includegraphics[width=1.2cm]{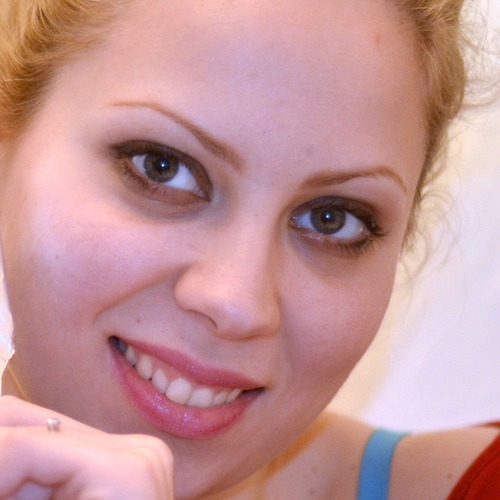}} & \multicolumn{1}{c|}{\includegraphics[width=1.2cm]{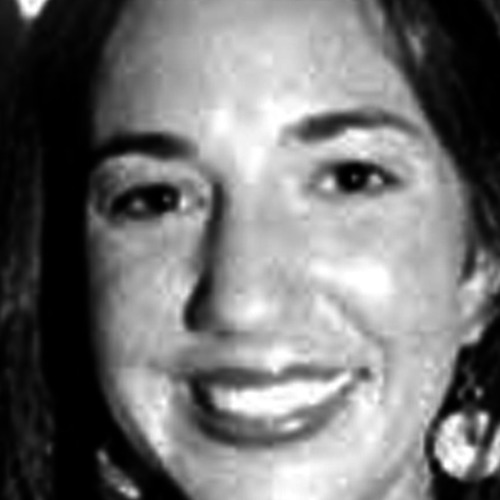}} & \multicolumn{1}{c|}{\includegraphics[width=1.2cm]{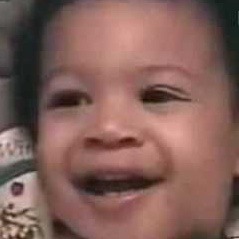}} & \multicolumn{1}{c|}{\includegraphics[width=1.2cm]{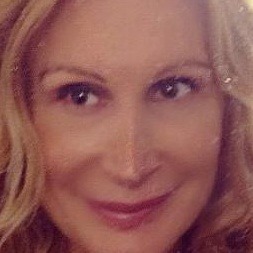}} \\ \cline{2-9} 
	\multicolumn{1}{|c|}{}                                       & \multicolumn{1}{c|}{\raisebox{0.5cm}{Sad}} & \multicolumn{1}{c|}{\includegraphics[width=1.2cm]{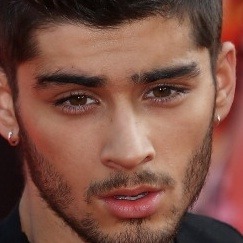}}           & \multicolumn{1}{c|}{\includegraphics[width=1.2cm]{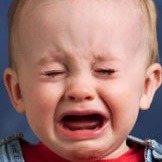}}          & \multicolumn{1}{c|}{\includegraphics[width=1.2cm]{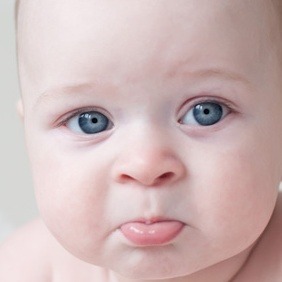}}           & \multicolumn{1}{c|}{\includegraphics[width=1.2cm]{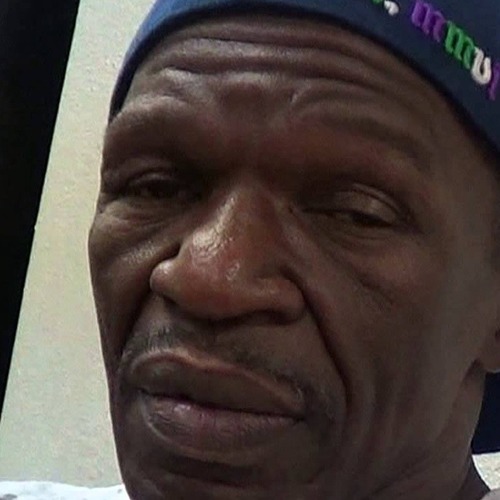} }          & \multicolumn{1}{c|}{\includegraphics[width=1.2cm]{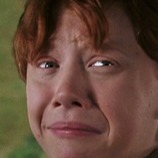}}         & \multicolumn{1}{c|}{\includegraphics[width=1.2cm]{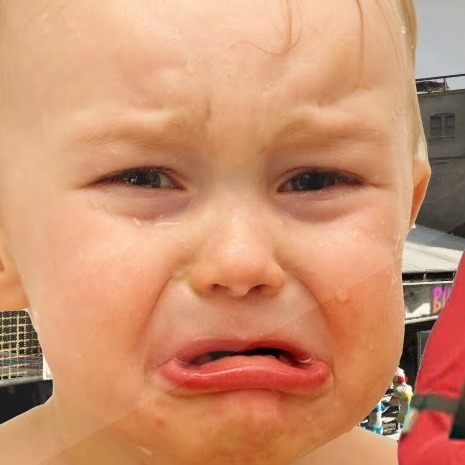}}           & \multicolumn{1}{c|}{\includegraphics[width=1.2cm]{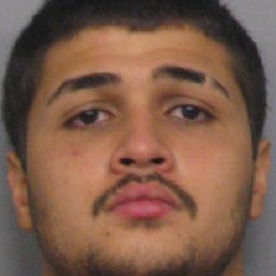}}           \\ \cline{2-9} 
	\multicolumn{1}{|c|}{}                                       & \multicolumn{1}{c|}{\raisebox{0.5cm}{Surprise}} & \multicolumn{1}{c|}{\includegraphics[width=1.2cm]{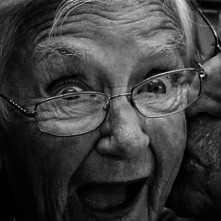}}           & \multicolumn{1}{c|}{\includegraphics[width=1.2cm]{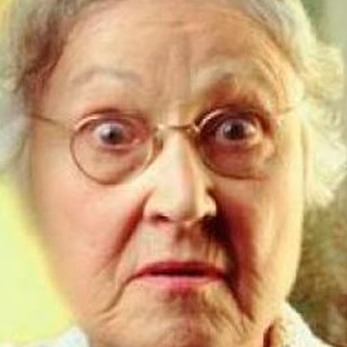}}           & \multicolumn{1}{c|}{\includegraphics[width=1.2cm]{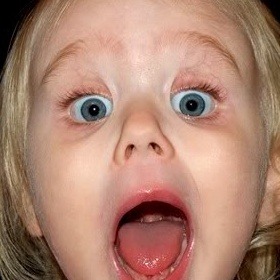}}            & \multicolumn{1}{c|}{\includegraphics[width=1.2cm]{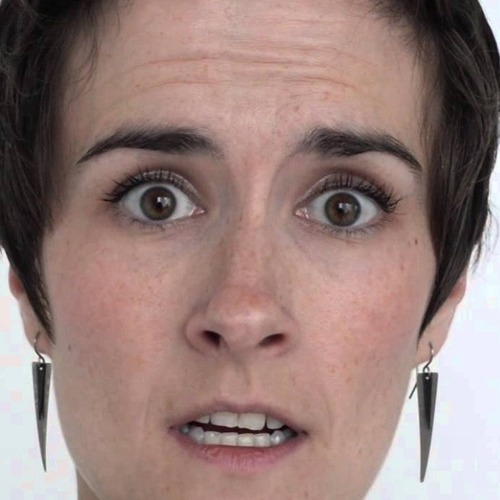} }          & \multicolumn{1}{c|}{\includegraphics[width=1.2cm]{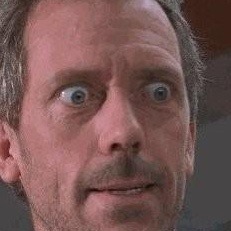} }        & \multicolumn{1}{c|}{\includegraphics[width=1.2cm]{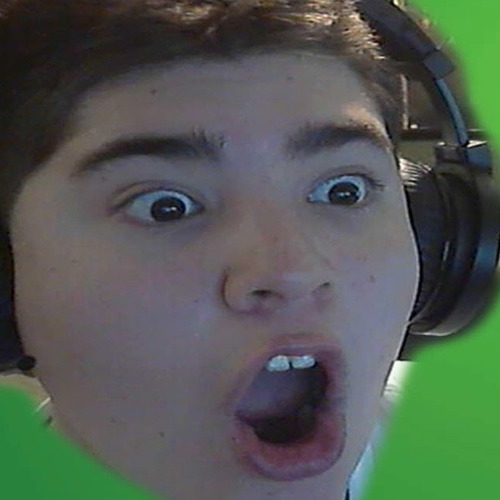}}           & \multicolumn{1}{c|}{\includegraphics[width=1.2cm]{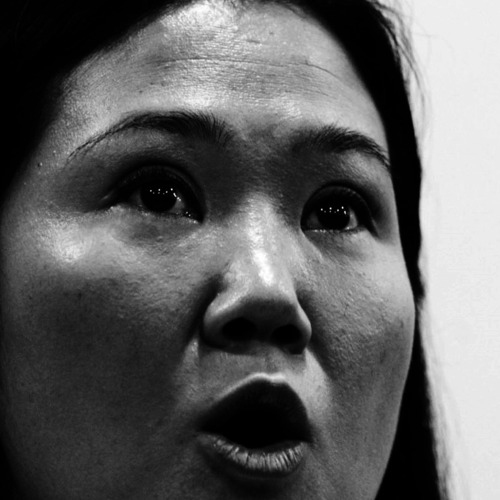}}           \\ \cline{2-9} 
	\multicolumn{1}{|c|}{}                                       & \multicolumn{1}{c|}{\raisebox{0.5cm}{Fear}} & \multicolumn{1}{c|}{\includegraphics[width=1.2cm]{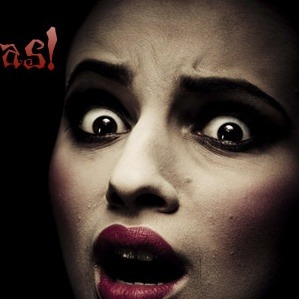}}           & \multicolumn{1}{c|}{\includegraphics[width=1.2cm]{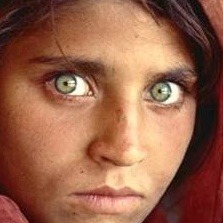}}           & \multicolumn{1}{c|}{\includegraphics[width=1.2cm]{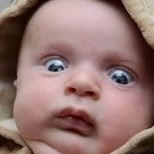}}           & \multicolumn{1}{c|}{\includegraphics[width=1.2cm]{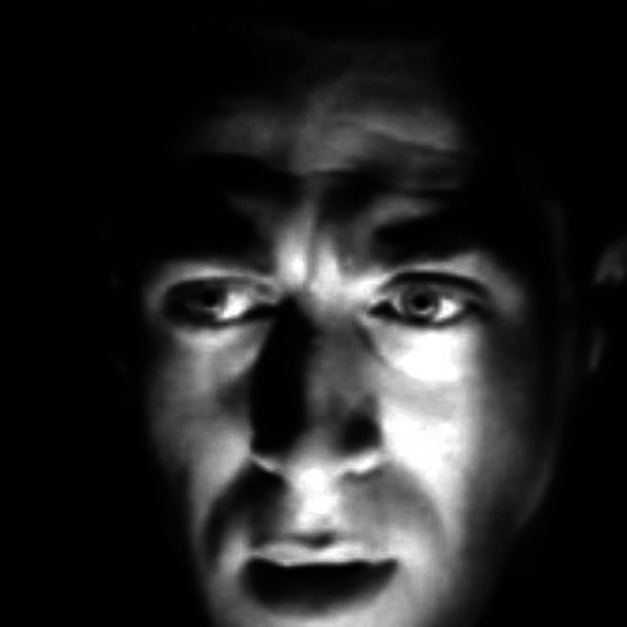}}             & \multicolumn{1}{c|}{\includegraphics[width=1.2cm]{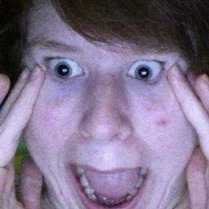}}         & \multicolumn{1}{c|}{\includegraphics[width=1.2cm]{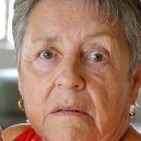}}           & \multicolumn{1}{c|}{\includegraphics[width=1.2cm]{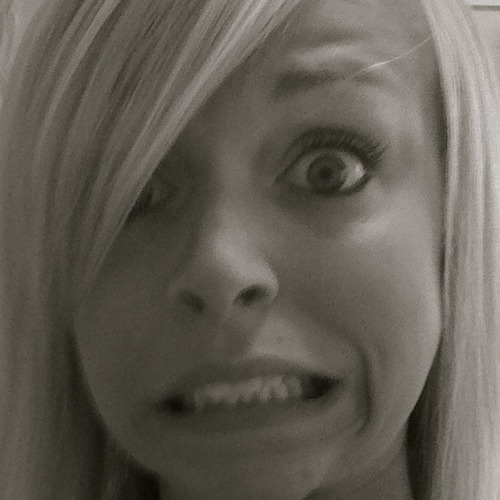}}           \\ \cline{2-9} 
	\multicolumn{1}{|c|}{}                                       & \multicolumn{1}{c|}{\raisebox{0.5cm}{Disgust}} & \multicolumn{1}{c|}{\includegraphics[width=1.2cm]{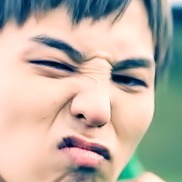}}           & \multicolumn{1}{c|}{\includegraphics[width=1.2cm]{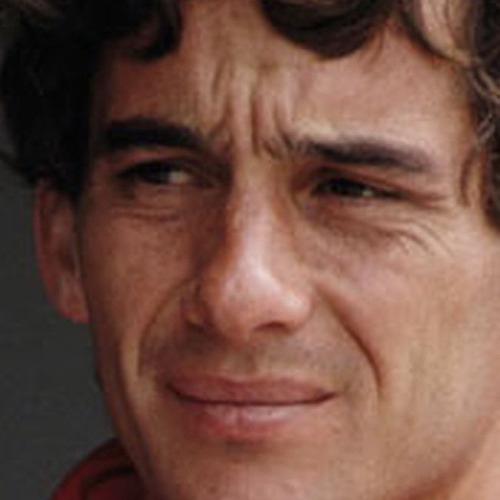}}           & \multicolumn{1}{c|}{\includegraphics[width=1.2cm]{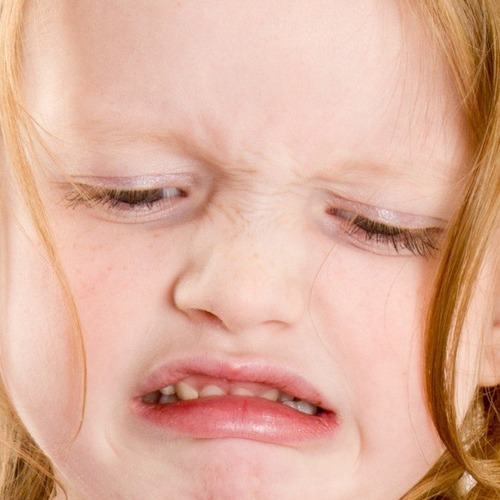}}           & \multicolumn{1}{c|}{\includegraphics[width=1.2cm]{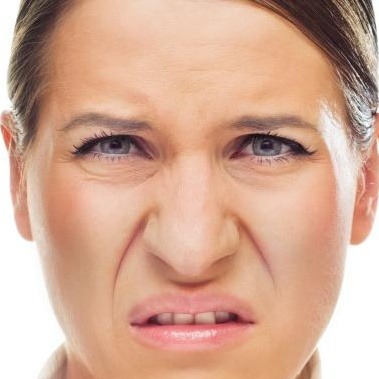}}           & \multicolumn{1}{c|}{\includegraphics[width=1.2cm]{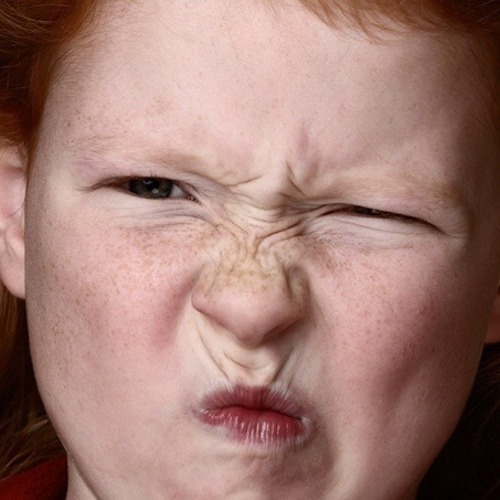}}         & \multicolumn{1}{c|}{\includegraphics[width=1.2cm]{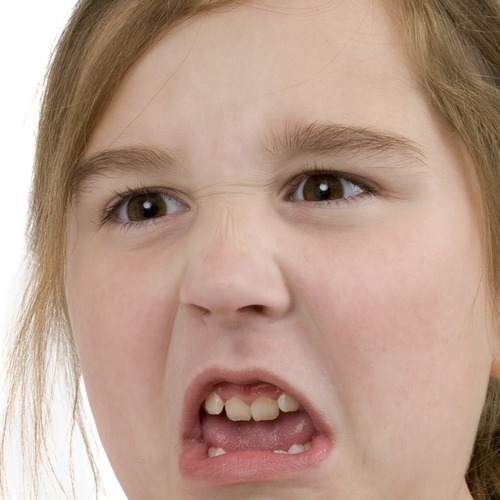}}           & \multicolumn{1}{c|}{\includegraphics[width=1.2cm]{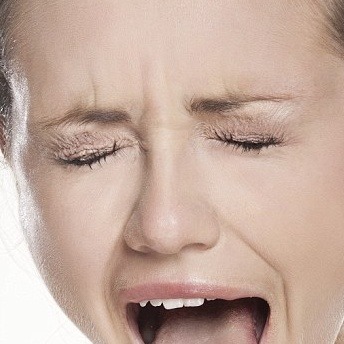}}             \\ \cline{2-9} 
	\multicolumn{1}{|c|}{}                                       & \multicolumn{1}{c|}{\raisebox{0.5cm}{Anger}} & \multicolumn{1}{c|}{\includegraphics[width=1.2cm]{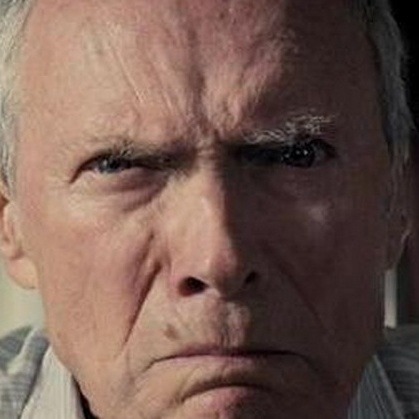}}           & \multicolumn{1}{c|}{\includegraphics[width=1.2cm]{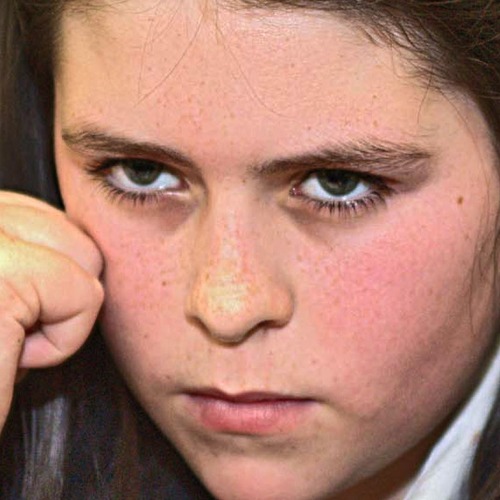}}           & \multicolumn{1}{c|}{\includegraphics[width=1.2cm]{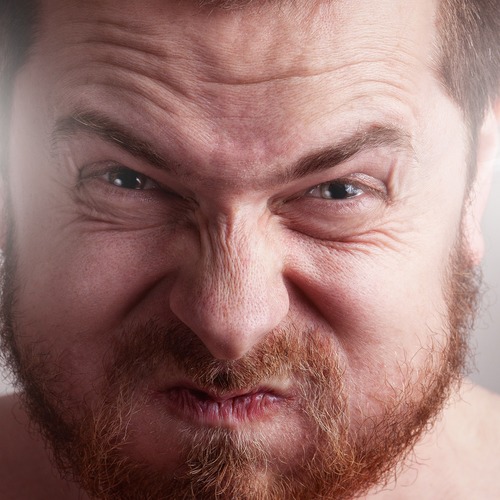}}           & \multicolumn{1}{c|}{\includegraphics[width=1.2cm]{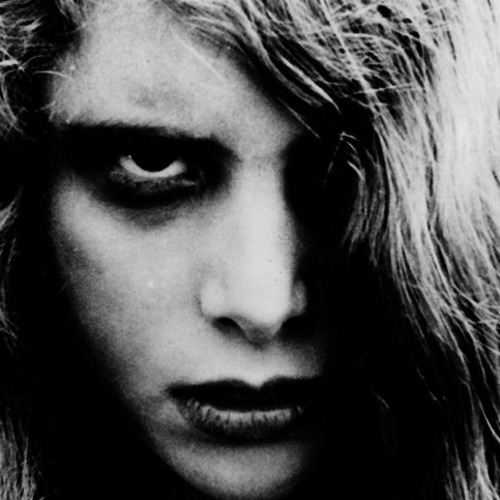}}           & \multicolumn{1}{c|}{\includegraphics[width=1.2cm]{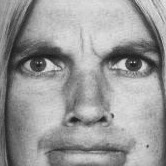}}           & \multicolumn{1}{c|}{\includegraphics[width=1.2cm]{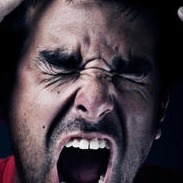}}          & \multicolumn{1}{c|}{\includegraphics[width=1.2cm]{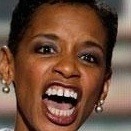}}           \\ \cline{2-9} 
	\multicolumn{1}{|c|}{}                                       & \multicolumn{1}{c|}{\raisebox{0.5cm}{Contempt}} & \multicolumn{1}{c|}{\includegraphics[width=1.2cm]{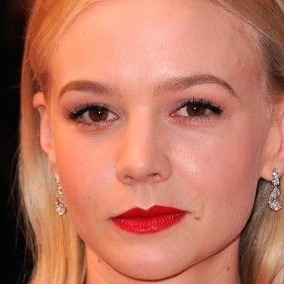}}           & \multicolumn{1}{c|}{\includegraphics[width=1.2cm]{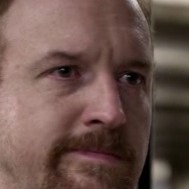}}          & \multicolumn{1}{c|}{\includegraphics[width=1.2cm]{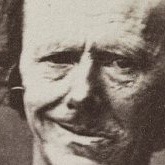}}           & \multicolumn{1}{c|}{\includegraphics[width=1.2cm]{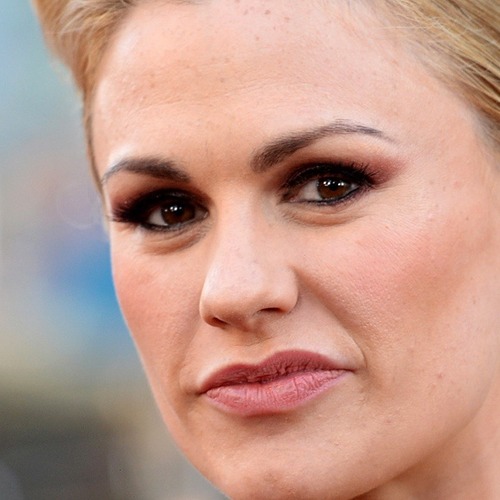}}           & \multicolumn{1}{c|}{\includegraphics[width=1.2cm]{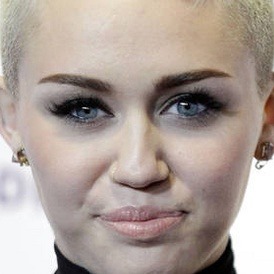}}         & \multicolumn{1}{c|}{\includegraphics[width=1.2cm]{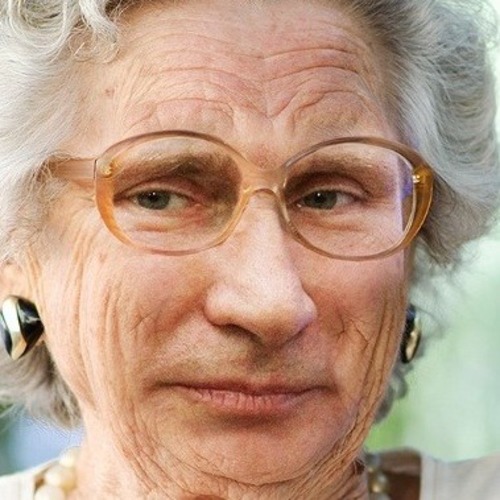}}           & \multicolumn{1}{c|}{\includegraphics[width=1.2cm]{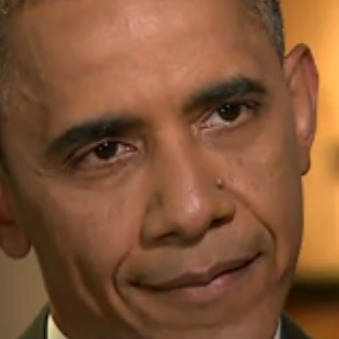}}           \\ \cline{2-9} 
	\multicolumn{1}{|c|}{}                                       & \multicolumn{1}{c|}{\raisebox{0.5cm}{None}} & \multicolumn{1}{c|}{\includegraphics[width=1.2cm]{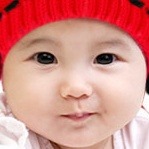}}           & \multicolumn{1}{c|}{\includegraphics[width=1.2cm]{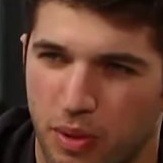} }          & \multicolumn{1}{c|}{\includegraphics[width=1.2cm]{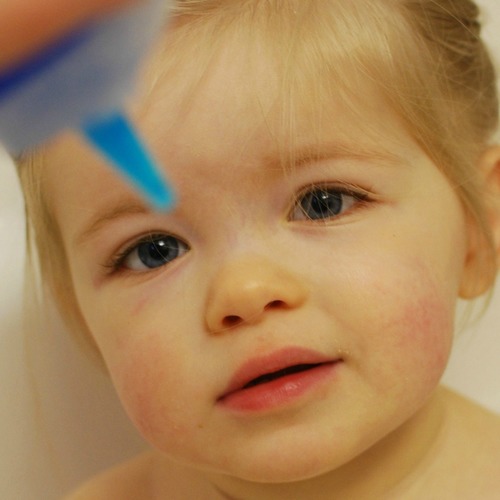} }          & \multicolumn{1}{c|}{\includegraphics[width=1.2cm]{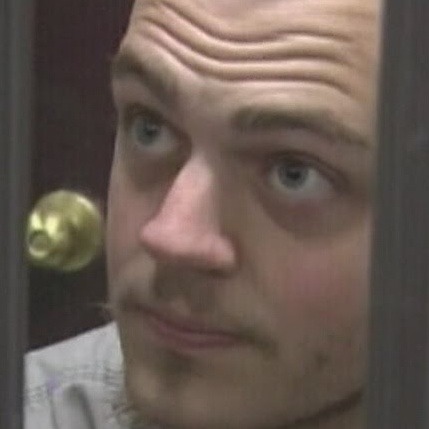}}           & \multicolumn{1}{c|}{\includegraphics[width=1.2cm]{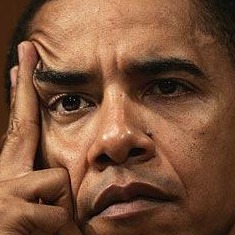}}         & \multicolumn{1}{c|}{\includegraphics[width=1.2cm]{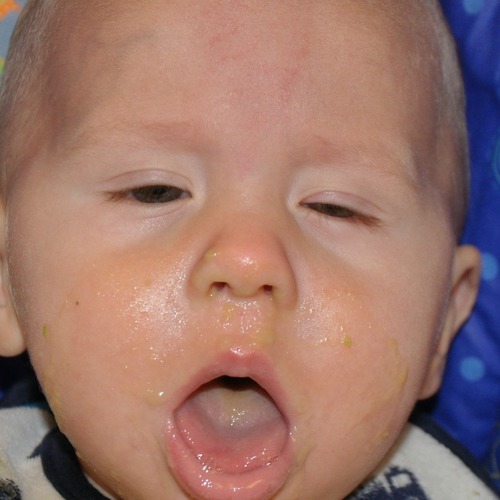}}           & \multicolumn{1}{c|}{\includegraphics[width=1.2cm]{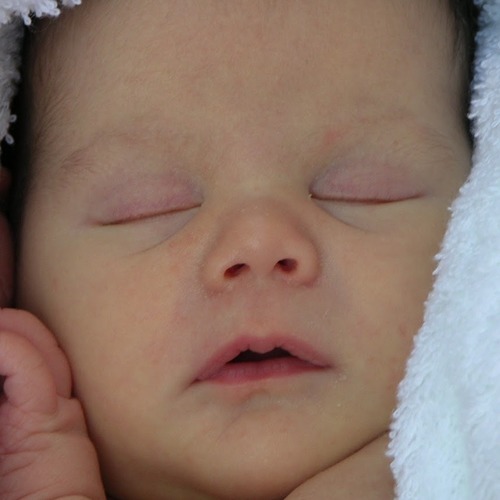}}          \\ \cline{2-9} 
	\multicolumn{1}{|c|}{}                                       & \multicolumn{1}{c|}{\raisebox{0.5cm}{Uncertain}} & \multicolumn{1}{c|}{\includegraphics[width=1.2cm]{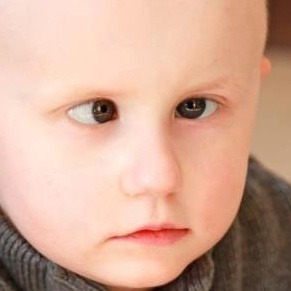}}           & \multicolumn{1}{c|}{\includegraphics[width=1.2cm]{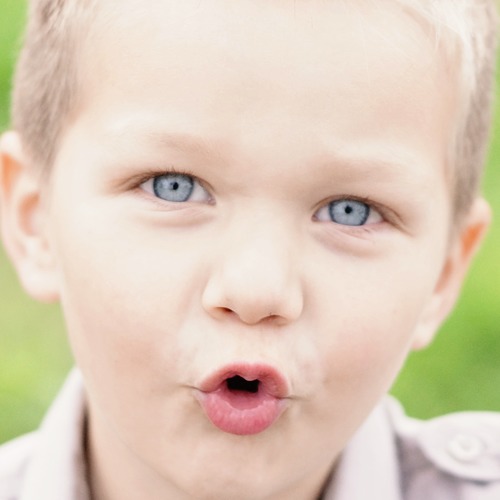}}           & \multicolumn{1}{c|}{\includegraphics[width=1.2cm]{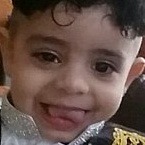}}           & \multicolumn{1}{c|}{\includegraphics[width=1.2cm]{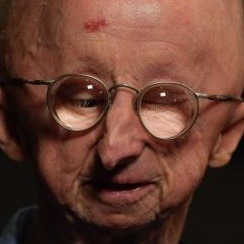}}           & \multicolumn{1}{c|}{\includegraphics[width=1.2cm]{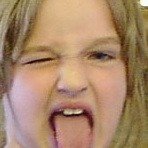}}         & \multicolumn{1}{c|}{\includegraphics[width=1.2cm]{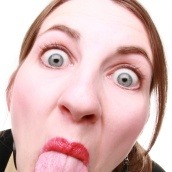}}           & \multicolumn{1}{c|}{\includegraphics[width=1.2cm]{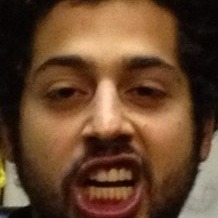}}           \\ \cline{2-9} 
	\multicolumn{1}{|c|}{}                                       & \multicolumn{1}{c|}{\raisebox{0.5cm}{Non-Face}} & \multicolumn{1}{c|}{\includegraphics[width=1.2cm]{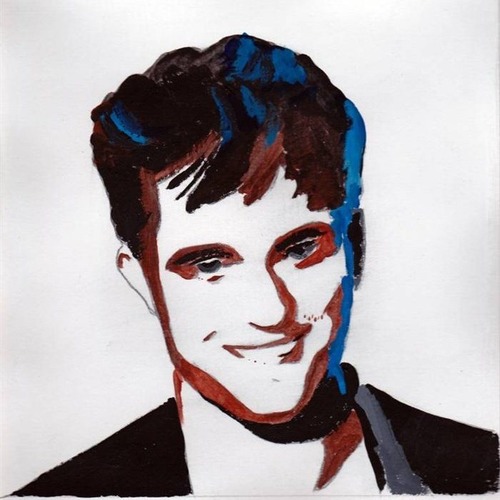}}          & \multicolumn{1}{c|}{\includegraphics[width=1.2cm]{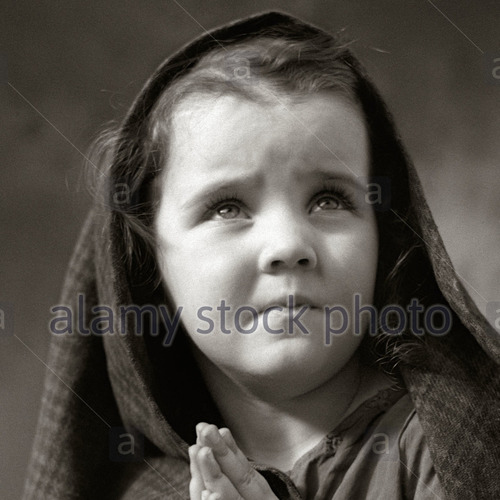}}          & \multicolumn{1}{c|}{\includegraphics[width=1.2cm]{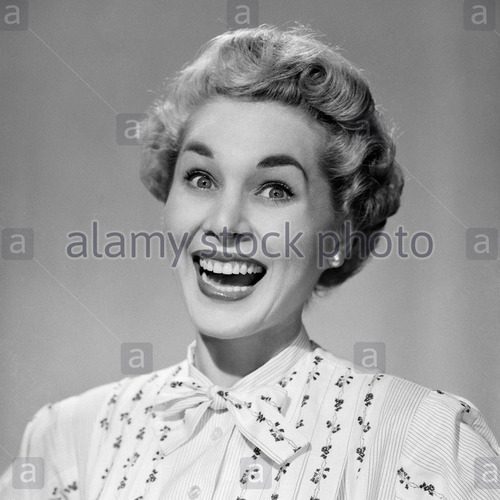}}          & \multicolumn{1}{c|}{\includegraphics[width=1.2cm]{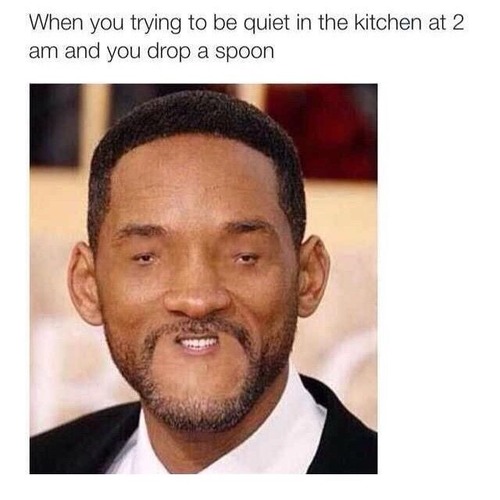}}          & \multicolumn{1}{c|}{\includegraphics[width=1.2cm]{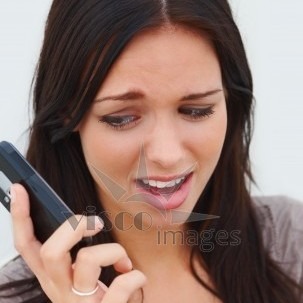}}        & \multicolumn{1}{c|}{\includegraphics[width=1.2cm]{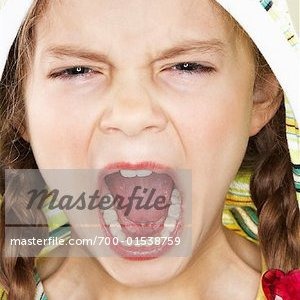}}          & \multicolumn{1}{c|}{\includegraphics[width=1.2cm]{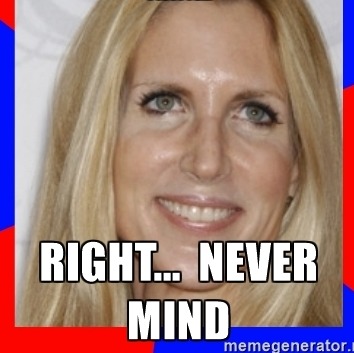}}          \\ \hline

\end{tabular}
\end{table}

\newpage

\begin{table}[]
	\centering
	\caption{Samples of Annotated Images by Two Annotators (Randomly selected)}
	\label{tab:AnnotatorsAgreement_appendix}
	\vspace{-2mm}
	\setlength\tabcolsep{3.2pt} 
\begin{tabular}{ll|c|c|c|c|c|c|c|c|c|c|c|}

\cline{3-13}

                                             &                       & \multicolumn{11}{c|}{Annotator 1}                                                                                                                                                              \\ \cline{3-13}

		& \multicolumn{1}{c|}{} &  Neutral       & Happy         & Sad           & Surprise      & Fear          & Disgust       & Anger         & Contempt      & None		  & Uncertain     & Non-Face       \\ \hline
		\multicolumn{1}{|l|}{\multirow{11}{*}{{\begin{turn}{+90}Annotator 2\end{turn}}}} & \raisebox{0.5cm}{Neutral}      & 
		\includegraphics[width=1.2cm]{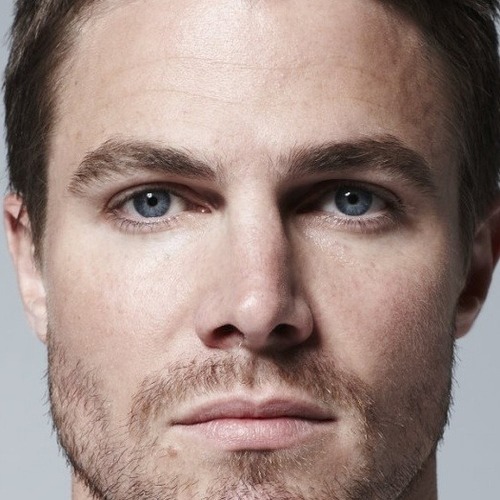} & \includegraphics[width=1.2cm]{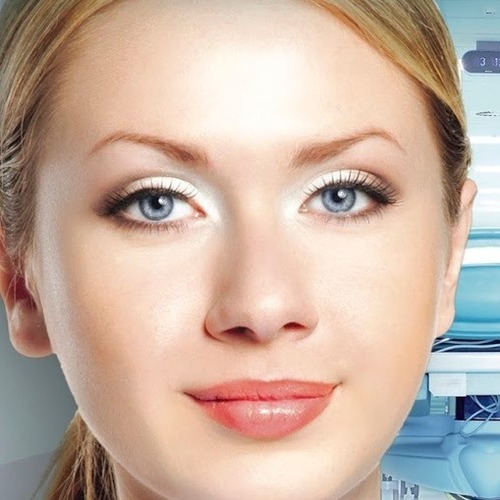}           & \includegraphics[width=1.2cm]{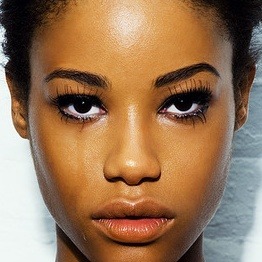}           & \includegraphics[width=1.2cm]{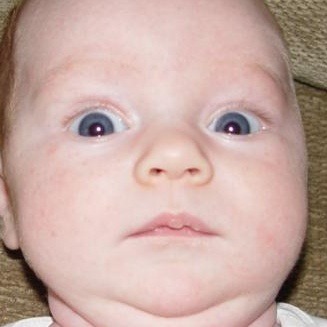}           & \includegraphics[width=1.2cm]{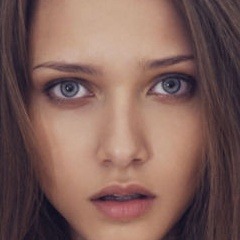}          & \includegraphics[width=1.2cm]{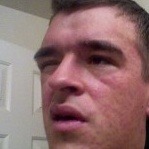}           & \includegraphics[width=1.2cm]{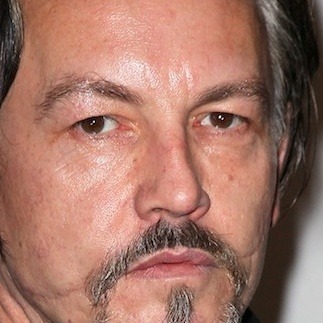}           & \includegraphics[width=1.2cm]{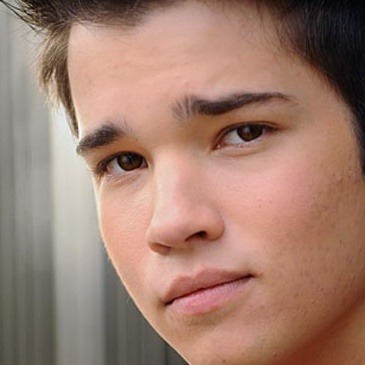}           & \includegraphics[width=1.2cm]{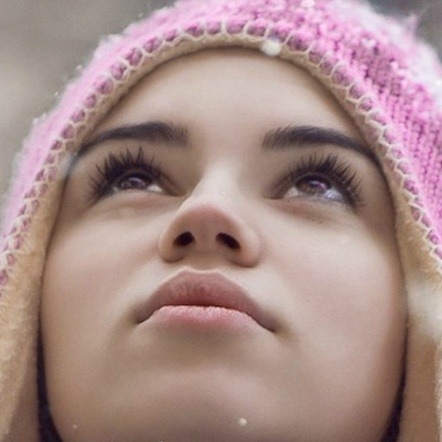}         & \includegraphics[width=1.2cm]{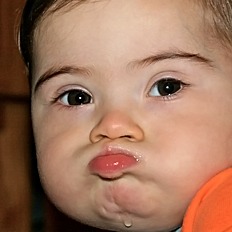}           & \includegraphics[width=1.2cm]{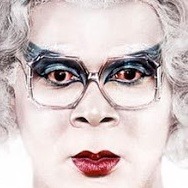}           \\ \cline{2-13}

		\multicolumn{1}{|l|}{}                       & \raisebox{0.5cm}{Happy}        & 
		\includegraphics[width=1.2cm]{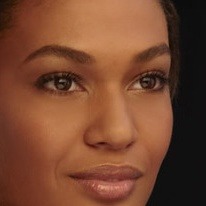}           & \includegraphics[width=1.2cm]{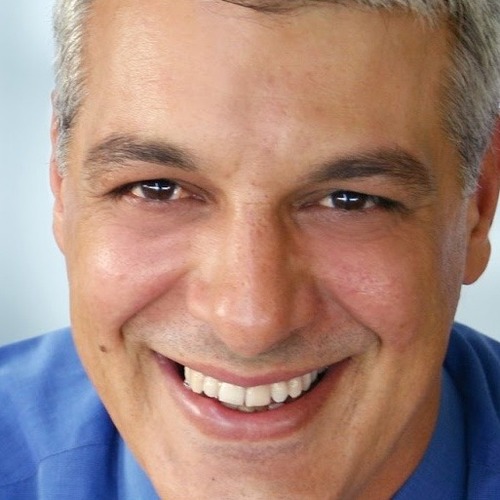} & \includegraphics[width=1.2cm]{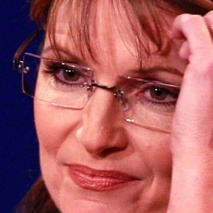}           & \includegraphics[width=1.2cm]{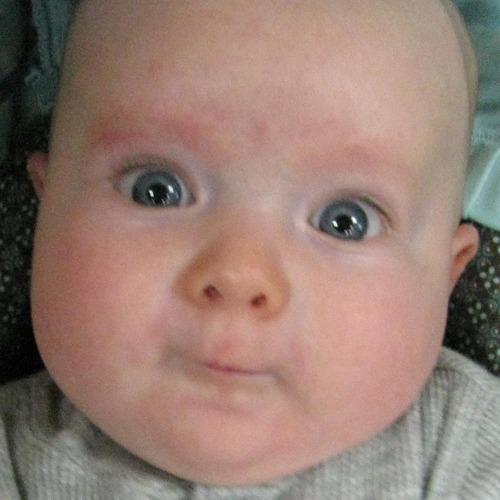}           & \includegraphics[width=1.2cm]{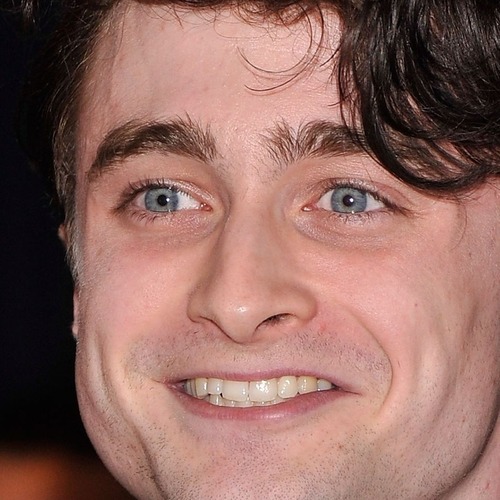}           & \includegraphics[width=1.2cm]{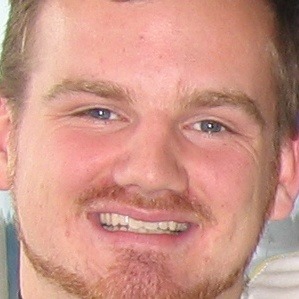}           & \includegraphics[width=1.2cm]{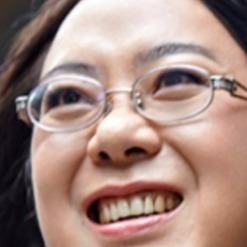}          & \includegraphics[width=1.2cm]{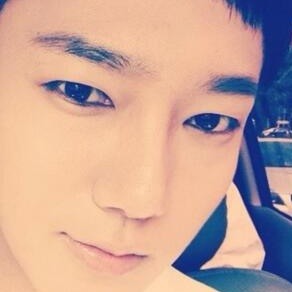}           & \includegraphics[width=1.2cm]{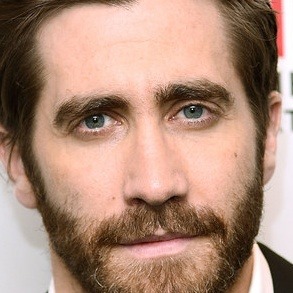}          & \includegraphics[width=1.2cm]{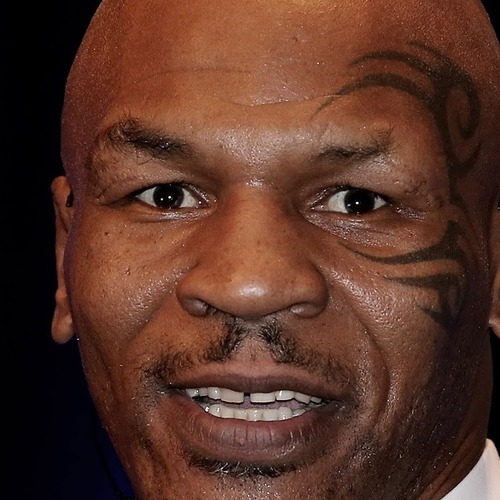}           & \includegraphics[width=1.2cm]{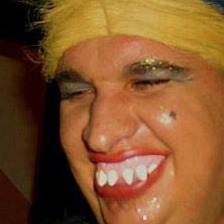}           \\ \cline{2-13}
		
		\multicolumn{1}{|l|}{}                       &\raisebox{0.5cm}{Sad}          & 
		\includegraphics[width=1.2cm]{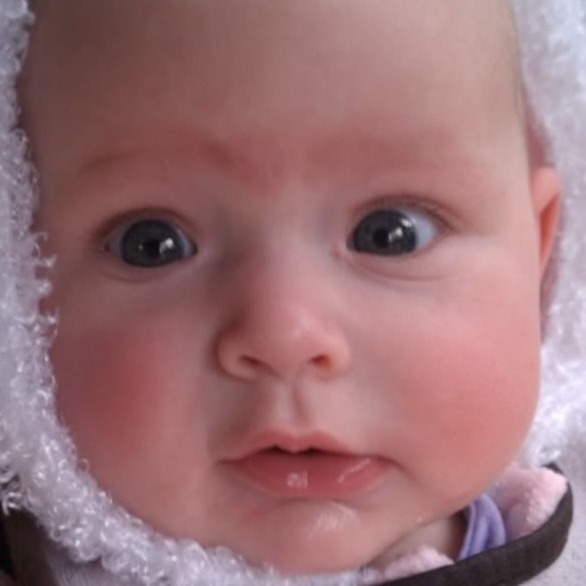}          & \includegraphics[width=1.2cm]{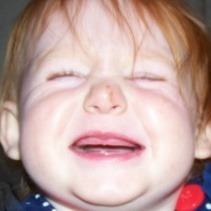}           & \includegraphics[width=1.2cm]{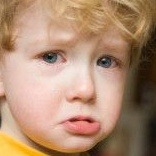} & \includegraphics[width=1.2cm]{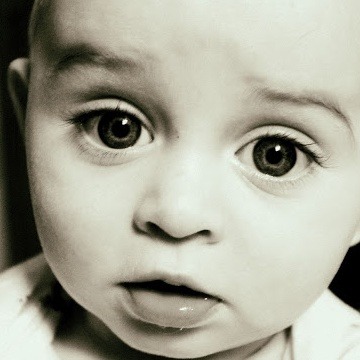}           & \includegraphics[width=1.2cm]{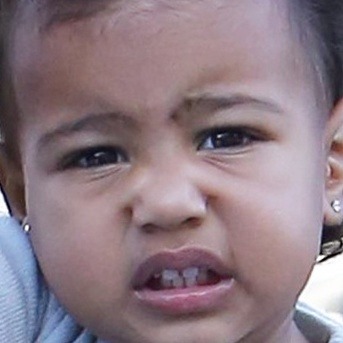}           & \includegraphics[width=1.2cm]{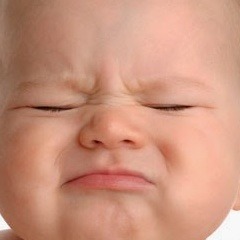}           & \includegraphics[width=1.2cm]{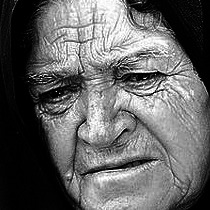}           & \includegraphics[width=1.2cm]{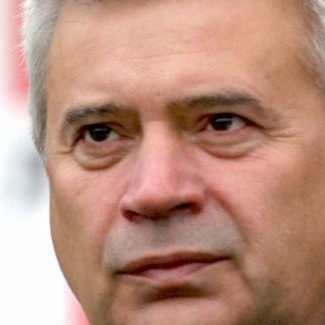}           & \includegraphics[width=1.2cm]{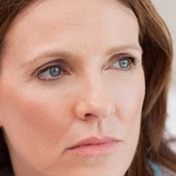}          & \includegraphics[width=1.2cm]{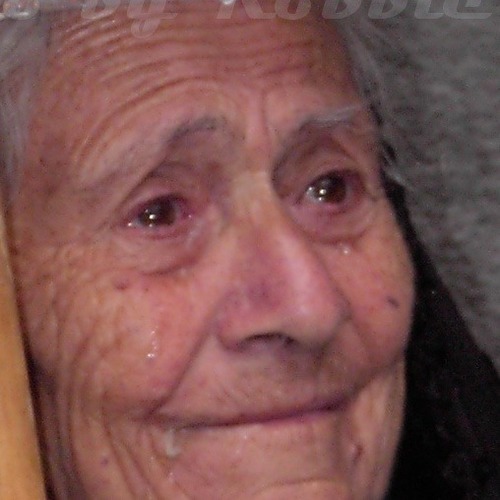}           & \includegraphics[width=1.2cm]{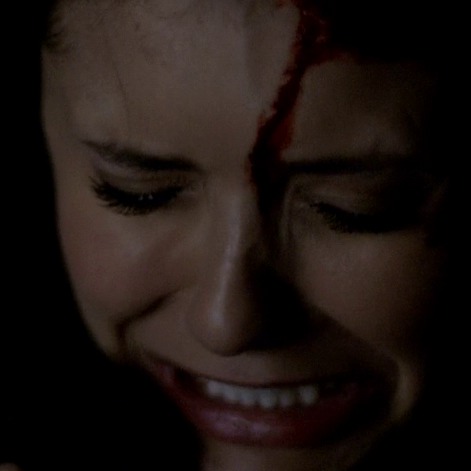}           \\ \cline{2-13}
		
		\multicolumn{1}{|l|}{}                       &\raisebox{0.5cm}{Surprise}     & 
		\includegraphics[width=1.2cm]{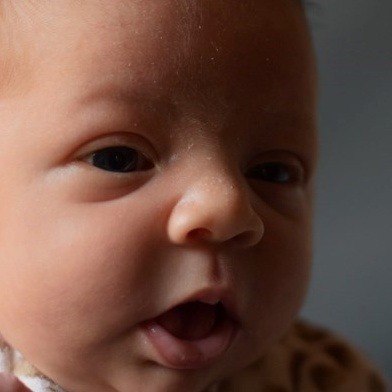}           & 		\includegraphics[width=1.2cm]{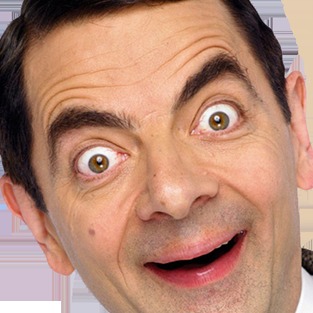}           & \includegraphics[width=1.2cm]{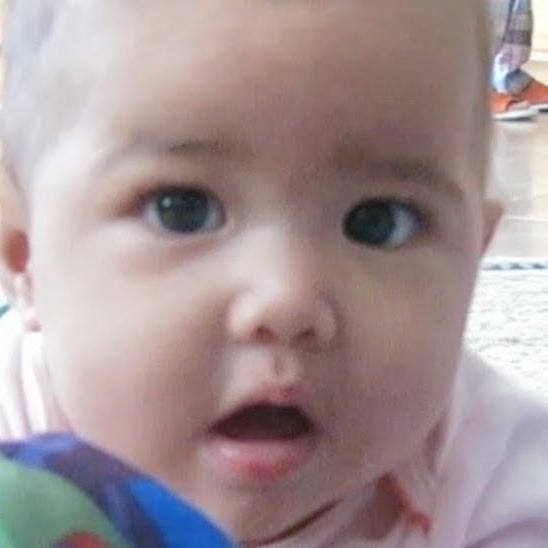}           & \includegraphics[width=1.2cm]{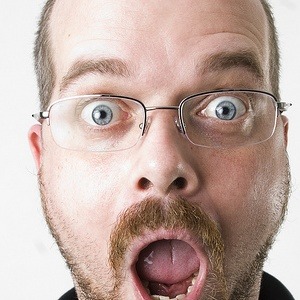} & \includegraphics[width=1.2cm]{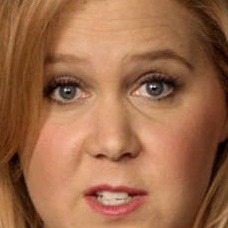}          & \includegraphics[width=1.2cm]{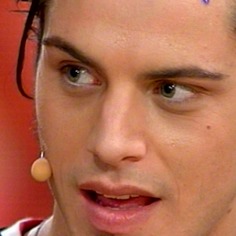}           & \includegraphics[width=1.2cm]{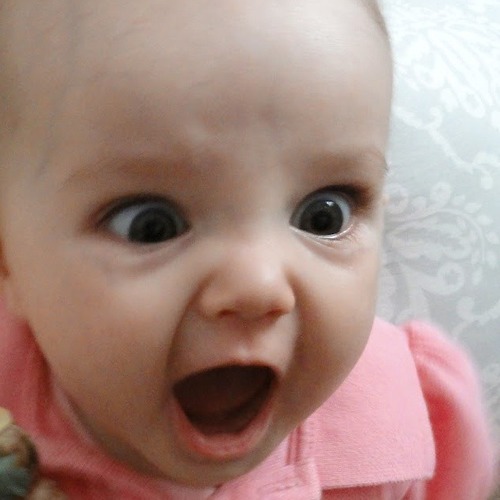}           & \includegraphics[width=1.2cm]{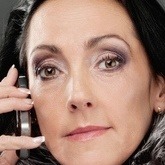}           & \includegraphics[width=1.2cm]{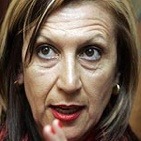}          & \includegraphics[width=1.2cm]{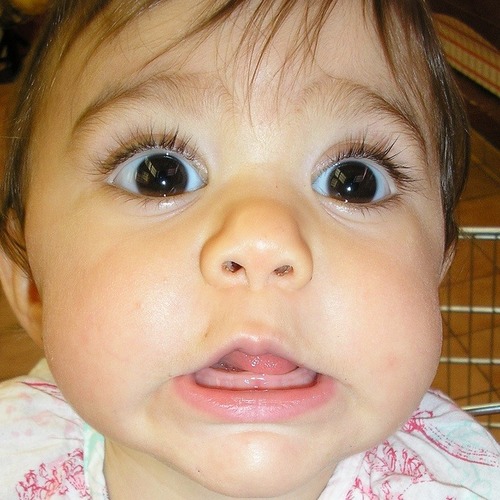}           & \includegraphics[width=1.2cm]{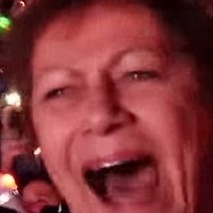}          \\ \cline{2-13}
		
		\multicolumn{1}{|l|}{}                       &\raisebox{0.5cm}{Fear}         & \includegraphics[width=1.2cm]{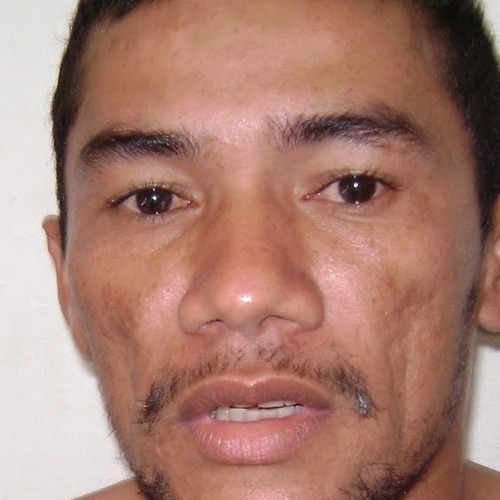}           & \includegraphics[width=1.2cm]{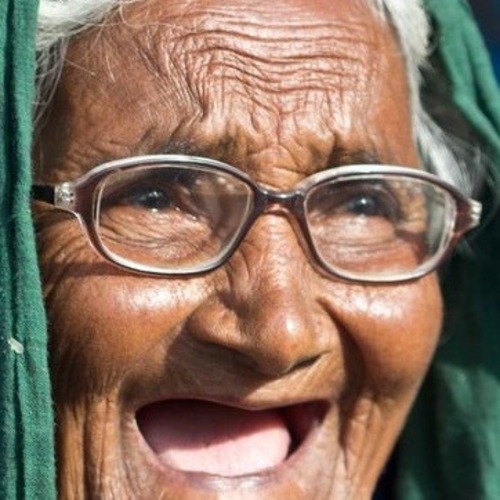}           & \includegraphics[width=1.2cm]{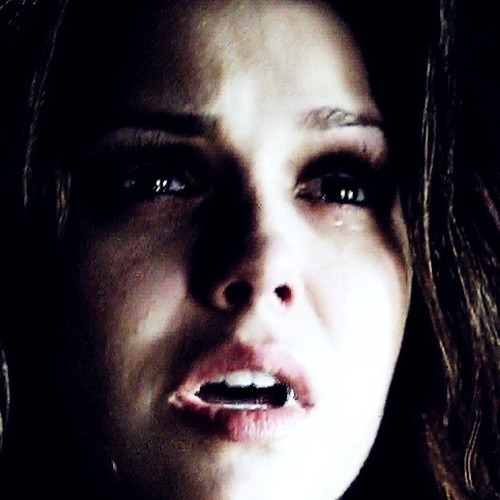}           & \includegraphics[width=1.2cm]{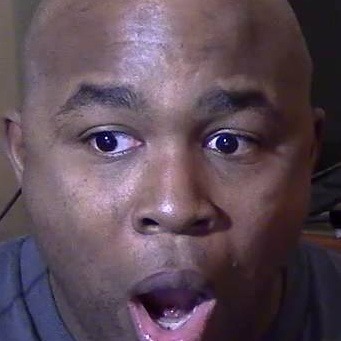}          & \includegraphics[width=1.2cm]{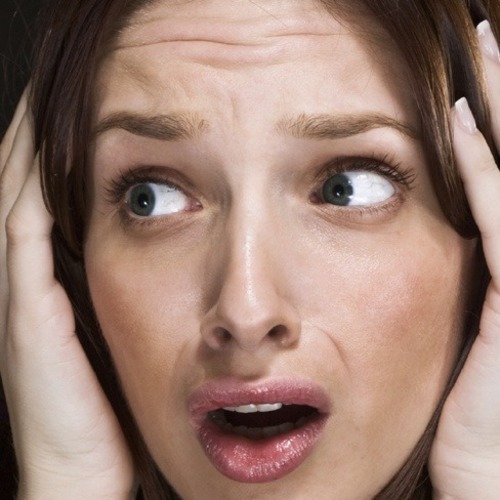} & \includegraphics[width=1.2cm]{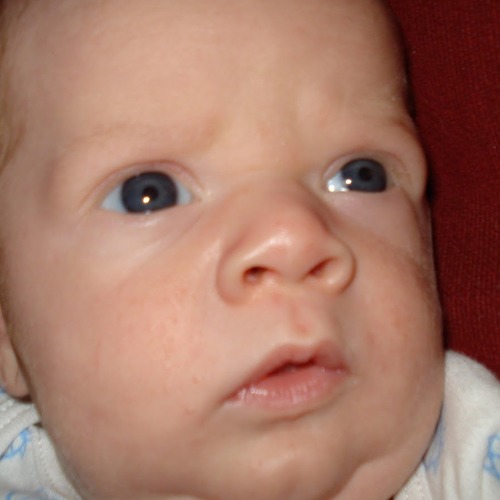}           & \includegraphics[width=1.2cm]{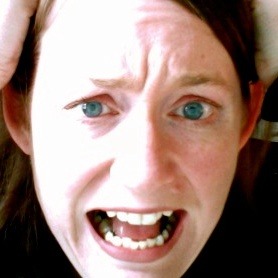}           & 
		\includegraphics[width=1.2cm]{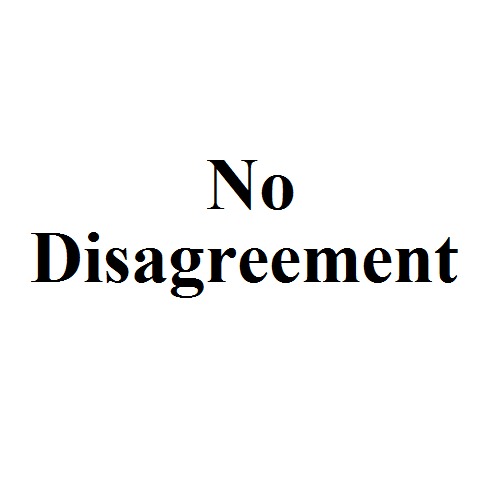}          & 
		\includegraphics[width=1.2cm]{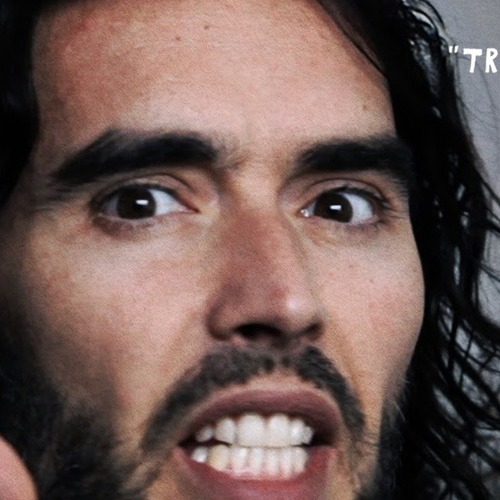}          & \includegraphics[width=1.2cm]{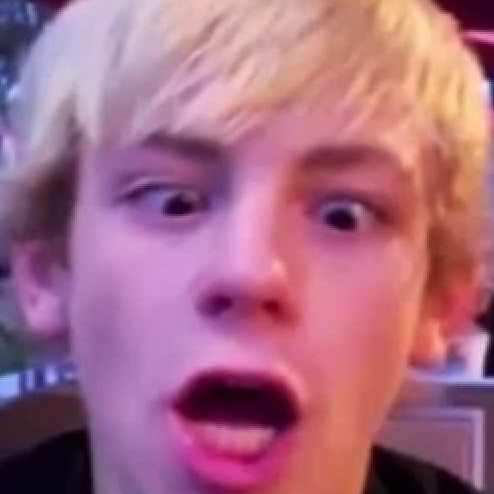}           & \includegraphics[width=1.2cm]{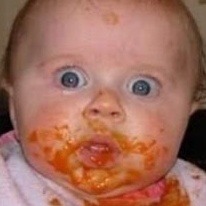}           \\ \cline{2-13}
		
		\multicolumn{1}{|l|}{}                       &\raisebox{0.5cm}{Disgust}      & 
		\includegraphics[width=1.2cm]{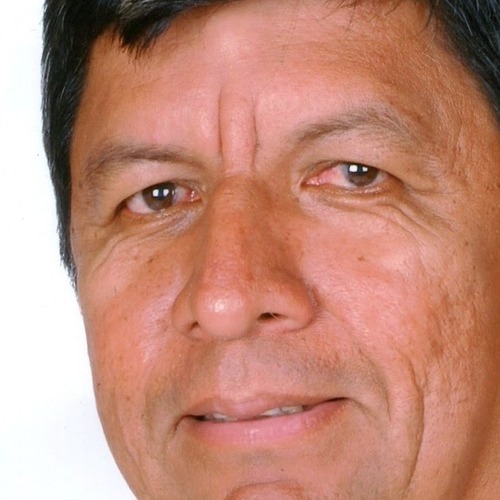}           & \includegraphics[width=1.2cm]{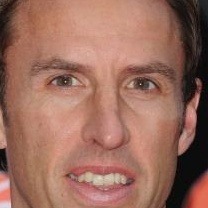}           & \includegraphics[width=1.2cm]{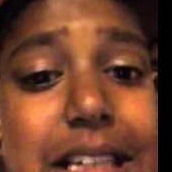}           & \includegraphics[width=1.2cm]{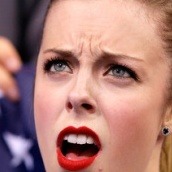}           & \includegraphics[width=1.2cm]{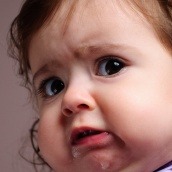}           & \includegraphics[width=1.2cm]{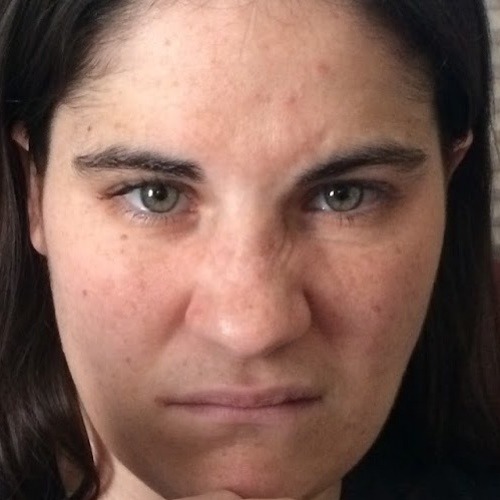} & \includegraphics[width=1.2cm]{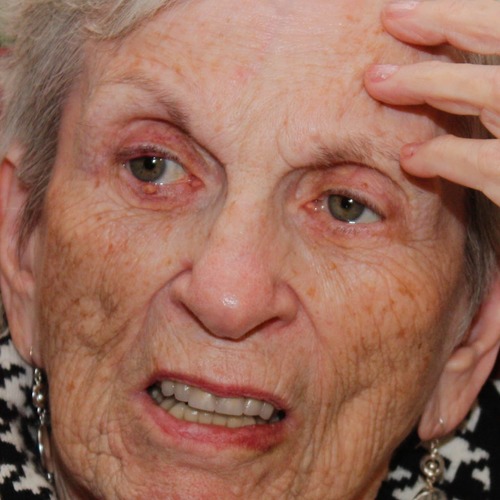}          & \includegraphics[width=1.2cm]{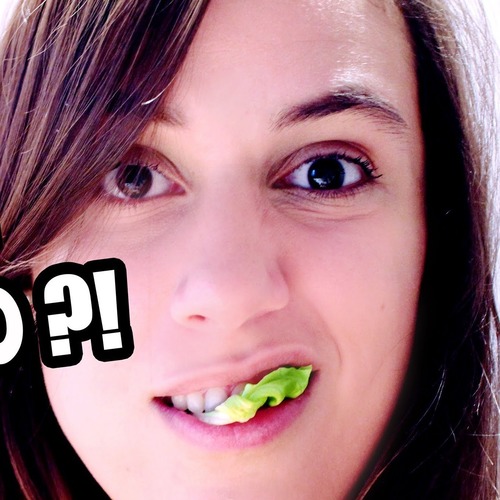}          & \includegraphics[width=1.2cm]{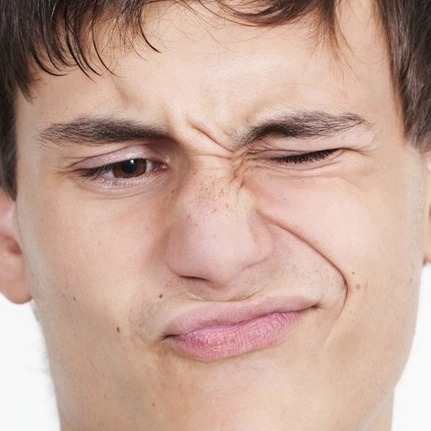}          & \includegraphics[width=1.2cm]{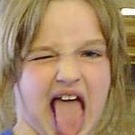}           & \includegraphics[width=1.2cm]{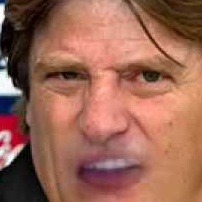}           \\ \cline{2-13}
		
		\multicolumn{1}{|l|}{}                       & \raisebox{0.5cm}{Anger}        & 
		\includegraphics[width=1.2cm]{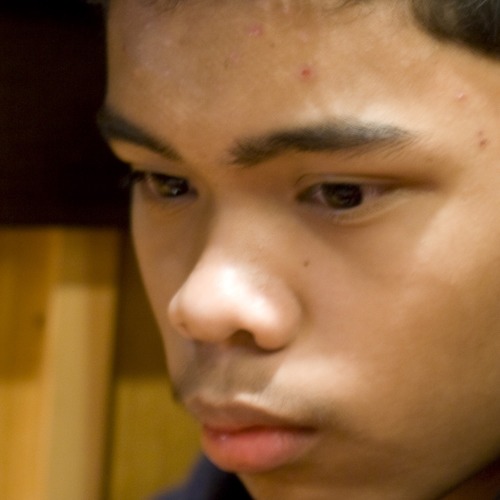}           & \includegraphics[width=1.2cm]{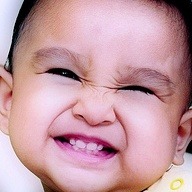}           & \includegraphics[width=1.2cm]{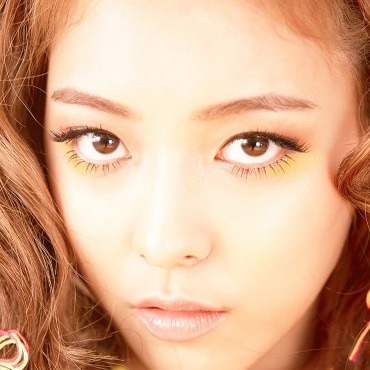}           & \includegraphics[width=1.2cm]{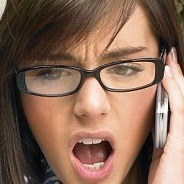}           & \includegraphics[width=1.2cm]{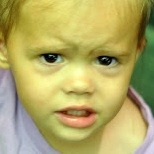}           & \includegraphics[width=1.2cm]{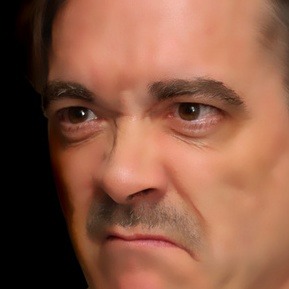}           & \includegraphics[width=1.2cm]{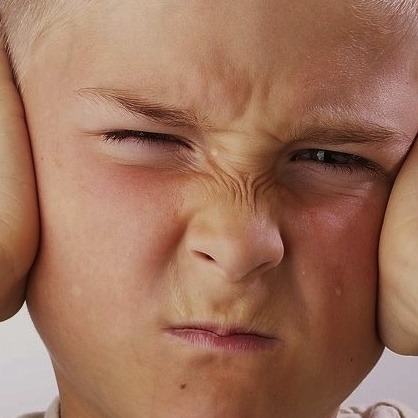} & \includegraphics[width=1.2cm]{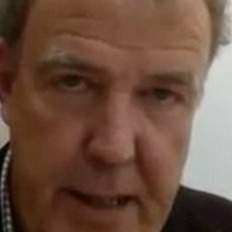}           & \includegraphics[width=1.2cm]{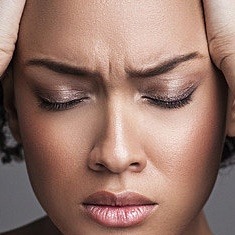}          & \includegraphics[width=1.2cm]{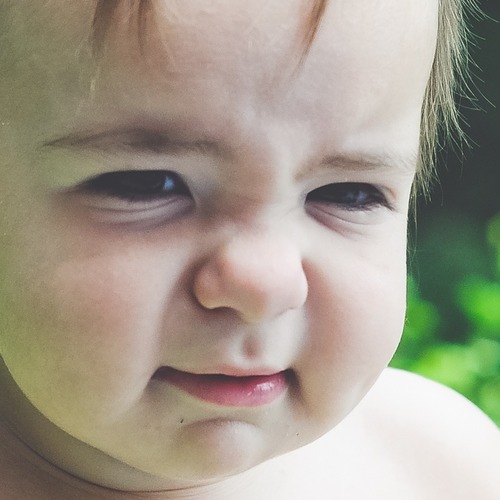}           & \includegraphics[width=1.2cm]{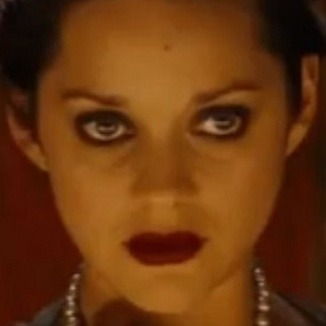}             \\ \cline{2-13}
		
		\multicolumn{1}{|l|}{}                       &\raisebox{0.5cm}{Contempt}     & 	\includegraphics[width=1.2cm]{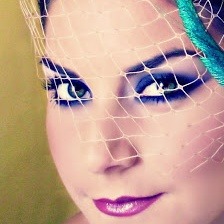}          & \includegraphics[width=1.2cm]{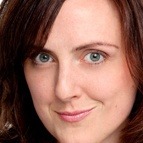}           & \includegraphics[width=1.2cm]{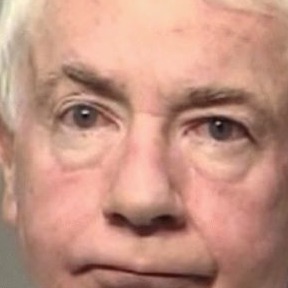}           & \includegraphics[width=1.2cm]{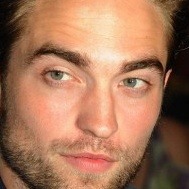}           & \includegraphics[width=1.2cm]{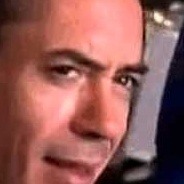}           & \includegraphics[width=1.2cm]{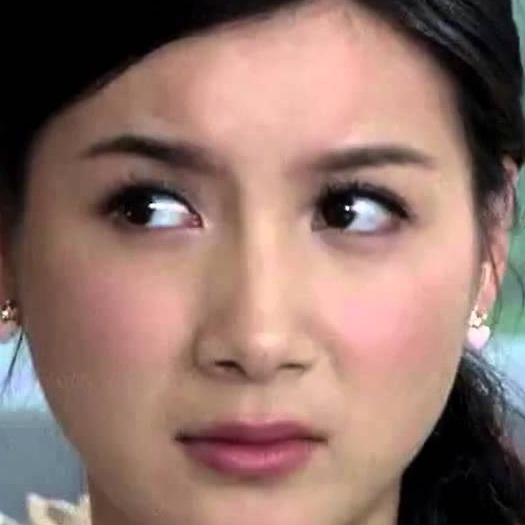}           & \includegraphics[width=1.2cm]{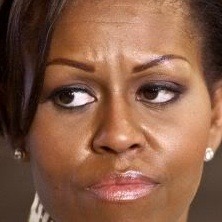}           & \includegraphics[width=1.2cm]{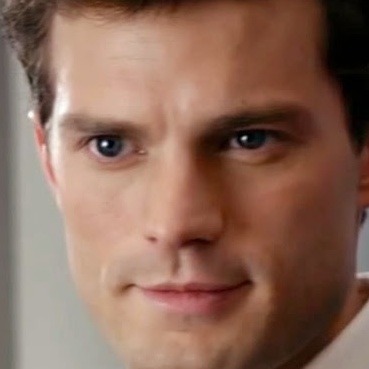} & \includegraphics[width=1.2cm]{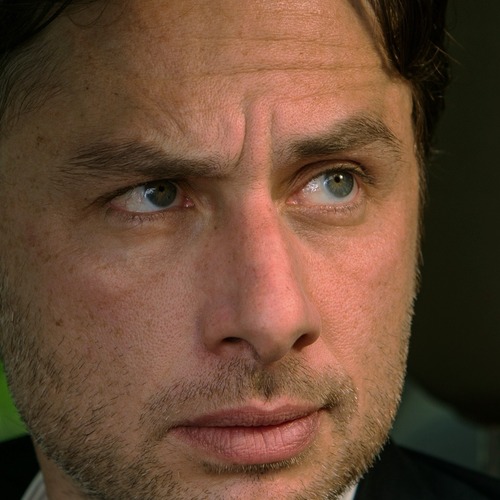}          & \includegraphics[width=1.2cm]{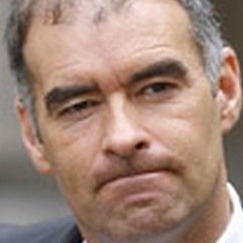}           & \includegraphics[width=1.2cm]{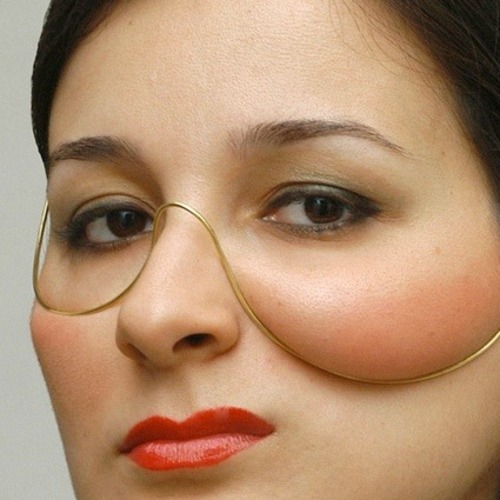}           \\ \cline{2-13}
		
		\multicolumn{1}{|l|}{}                       &\raisebox{0.5cm}{None} 		   & 
		\includegraphics[width=1.2cm]{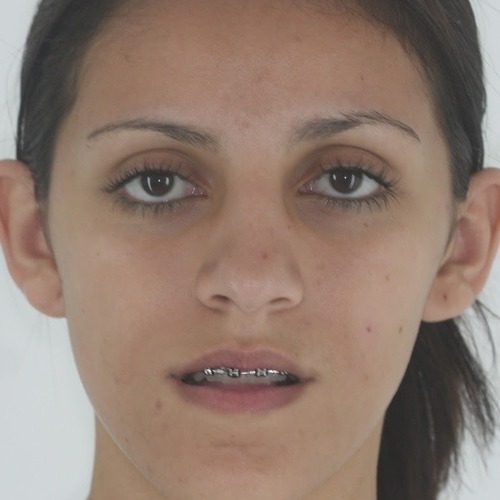} & \includegraphics[width=1.2cm]{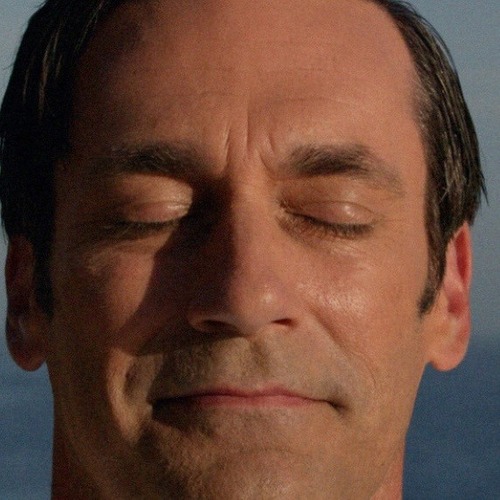}          & \includegraphics[width=1.2cm]{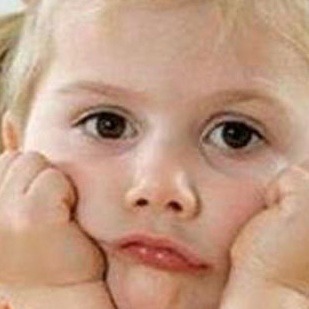}          & \includegraphics[width=1.2cm]{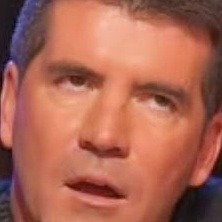}           & \includegraphics[width=1.2cm]{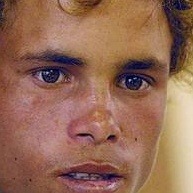}           & \includegraphics[width=1.2cm]{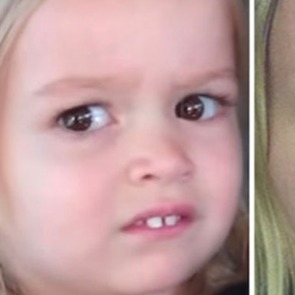}           & \includegraphics[width=1.2cm]{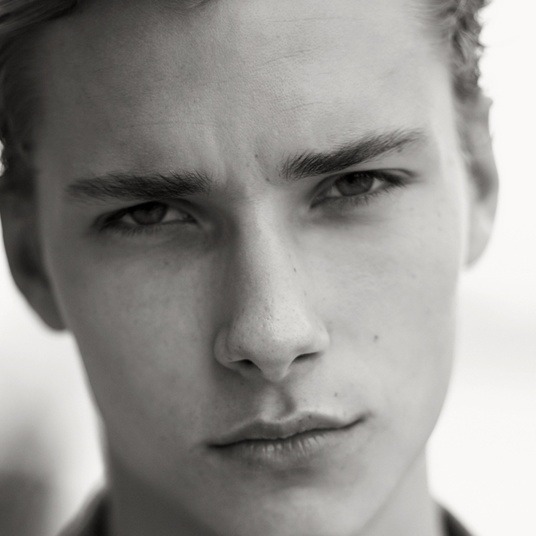}          & \includegraphics[width=1.2cm]{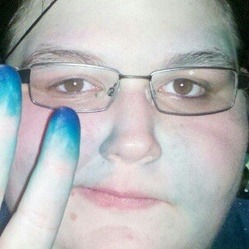}           & \includegraphics[width=1.2cm]{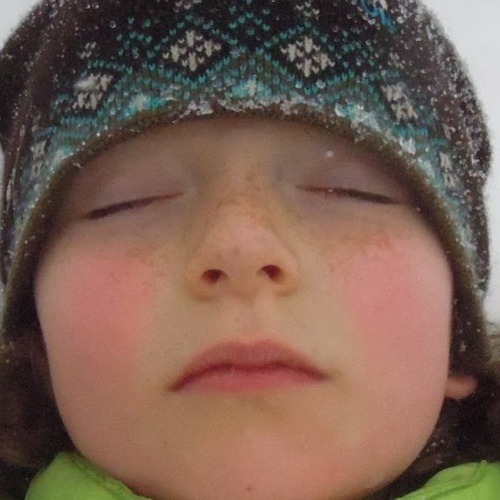}          & \includegraphics[width=1.2cm]{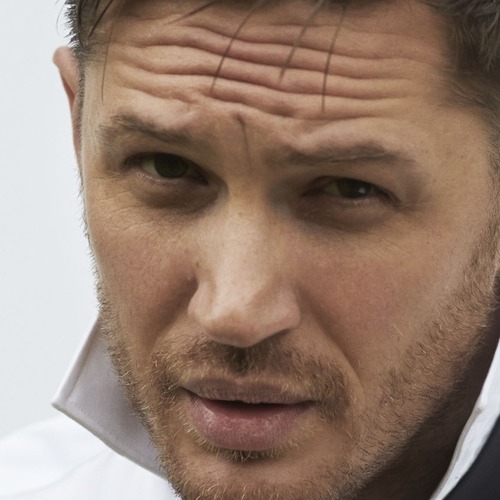}           & \includegraphics[width=1.2cm]{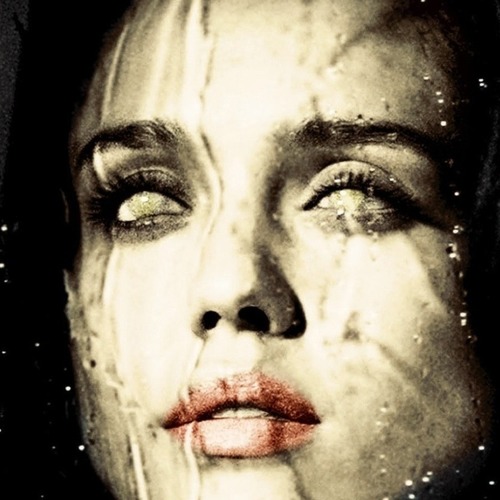}           \\ \cline{2-13}
		
		\multicolumn{1}{|l|}{}                       &\raisebox{0.5cm}{Uncertain}    & 
		\includegraphics[width=1.2cm]{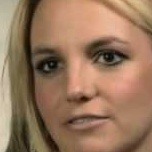}          & 		\includegraphics[width=1.2cm]{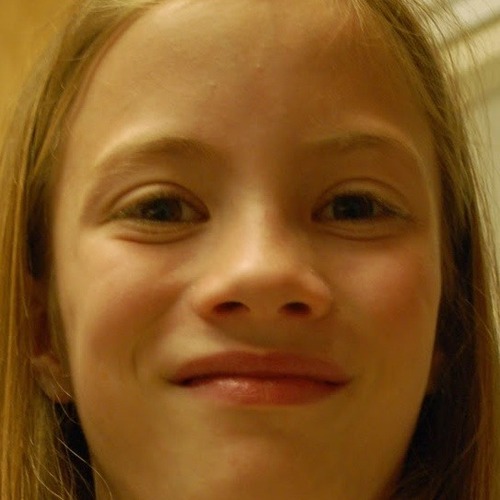}          & \includegraphics[width=1.2cm]{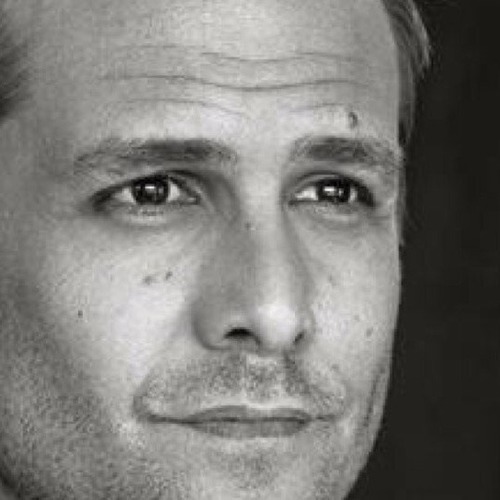}           & \includegraphics[width=1.2cm]{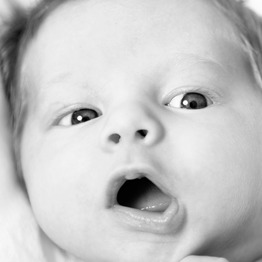}           & \includegraphics[width=1.2cm]{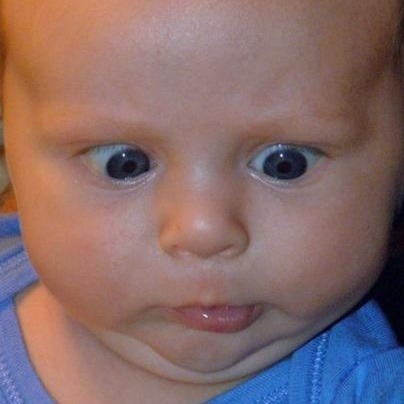}           & \includegraphics[width=1.2cm]{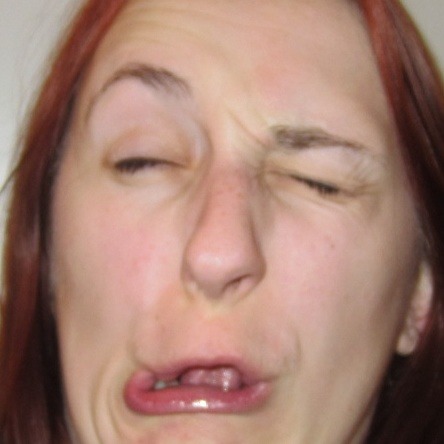}           & \includegraphics[width=1.2cm]{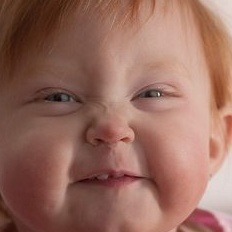}           & \includegraphics[width=1.2cm]{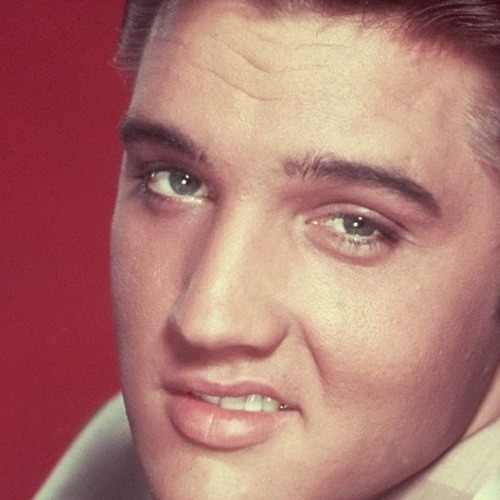}           & \includegraphics[width=1.2cm]{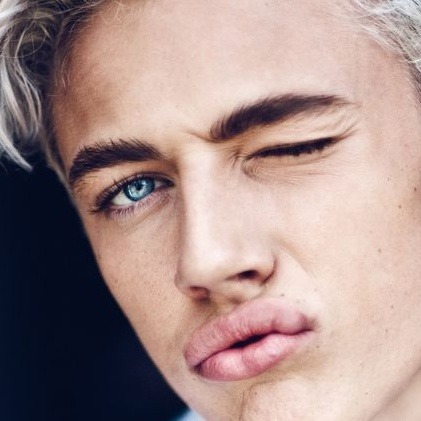}         & \includegraphics[width=1.2cm]{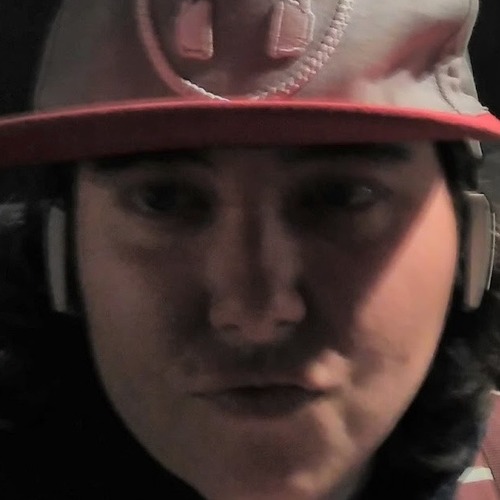} & \includegraphics[width=1.2cm]{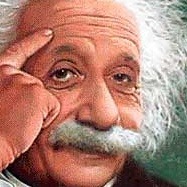}           \\ \cline{2-13}
		
		\multicolumn{1}{|l|}{}                       &\raisebox{0.5cm}{Non-Face}      & 
		\includegraphics[width=1.2cm]{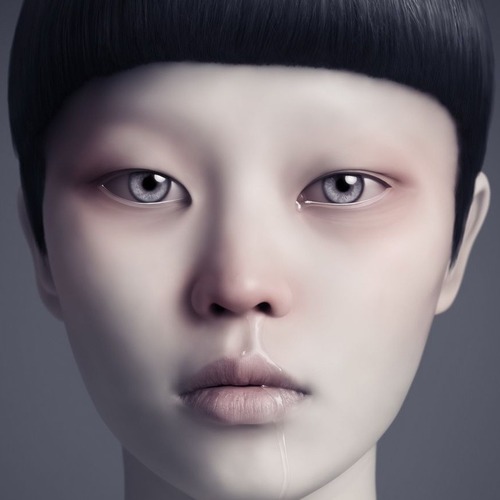}           & 		\includegraphics[width=1.2cm]{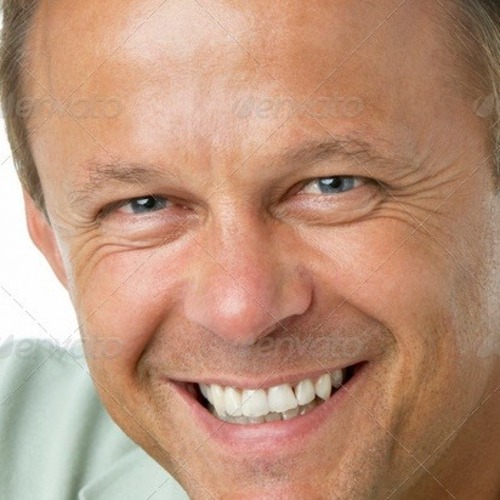}           & \includegraphics[width=1.2cm]{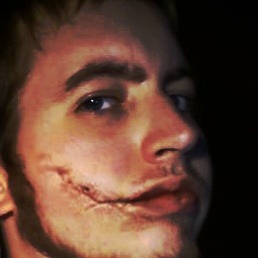}           & \includegraphics[width=1.2cm]{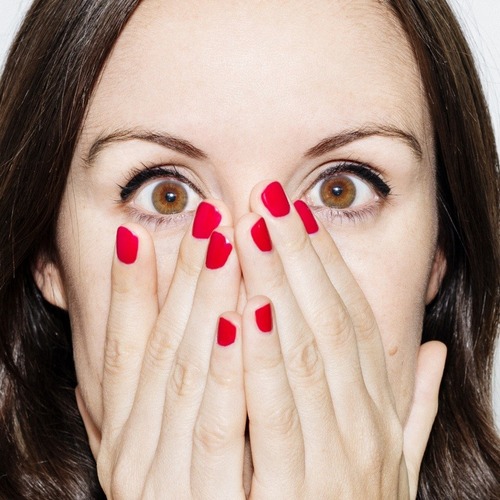}           & \includegraphics[width=1.2cm]{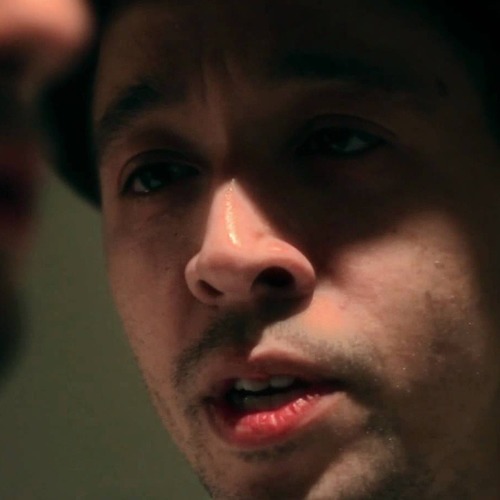}           & \includegraphics[width=1.2cm]{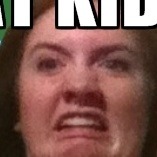}           & \includegraphics[width=1.2cm]{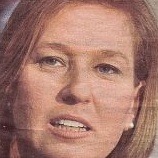}           & \includegraphics[width=1.2cm]{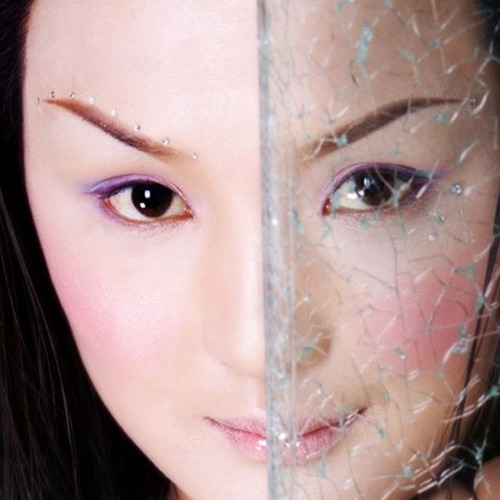}           & \includegraphics[width=1.2cm]{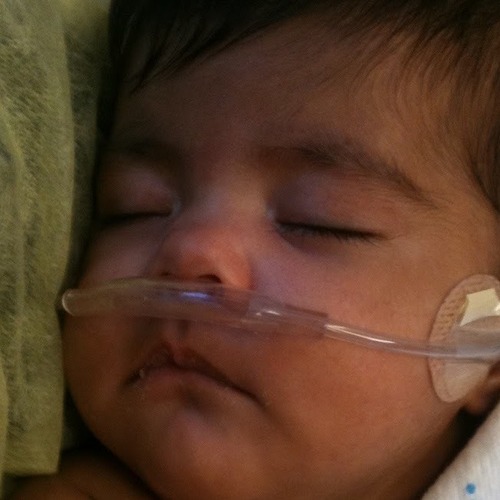}          & \includegraphics[width=1.2cm]{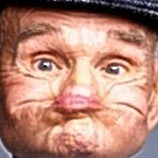}           & \includegraphics[width=1.2cm]{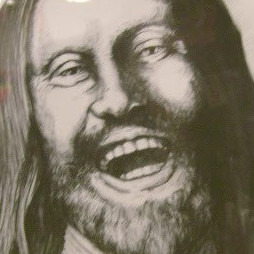} \\ \hline
	\end{tabular}
\end{table}

\begin{table}[H]
	\centering
	
	\caption{Agreement percentage between Two Annotators in Categorical Model of Affect (\%)}
	\label{Tab:AnnotatorsAgreement_appendix}
	\begin{tabular}{ccccccccccccc}
		\cline{2-13}
		\multicolumn{1}{c|}{}     & \multicolumn{1}{c|}{A1*}   & \multicolumn{1}{c|}{A2}   & \multicolumn{1}{c|}{A3}   & \multicolumn{1}{c|}{A4}   & \multicolumn{1}{c|}{A5}   & \multicolumn{1}{c|}{A6}   & \multicolumn{1}{c|}{A7}   & \multicolumn{1}{c|}{A8}   & \multicolumn{1}{c|}{A9}   & \multicolumn{1}{c|}{A10}  & \multicolumn{1}{c|}{A11}  & \multicolumn{1}{c|}{A12}  \\ \hline
		\multicolumn{1}{|c|}{A1}  & \multicolumn{1}{c|}{0.0**} & \multicolumn{1}{c|}{69}   & \multicolumn{1}{c|}{70}   & \multicolumn{1}{c|}{68}   & \multicolumn{1}{c|}{0}    & \multicolumn{1}{c|}{0}    & \multicolumn{1}{c|}{0}    & \multicolumn{1}{c|}{0}    & \multicolumn{1}{c|}{0}    & \multicolumn{1}{c|}{0}    & \multicolumn{1}{c|}{0}    & \multicolumn{1}{c|}{0}    \\ \hline
		\multicolumn{1}{|c|}{A2}  & \multicolumn{1}{c|}{69}    & \multicolumn{1}{c|}{0}    & \multicolumn{1}{c|}{64.9} & \multicolumn{1}{c|}{68.3} & \multicolumn{1}{c|}{0}    & \multicolumn{1}{c|}{0}    & \multicolumn{1}{c|}{0}    & \multicolumn{1}{c|}{64.7} & \multicolumn{1}{c|}{0}    & \multicolumn{1}{c|}{0}    & \multicolumn{1}{c|}{0}    & \multicolumn{1}{c|}{0}    \\ \hline
		\multicolumn{1}{|c|}{A3}  & \multicolumn{1}{c|}{70}    & \multicolumn{1}{c|}{64.9} & \multicolumn{1}{c|}{0}    & \multicolumn{1}{c|}{70.6} & \multicolumn{1}{c|}{67.4} & \multicolumn{1}{c|}{69.9} & \multicolumn{1}{c|}{63}   & \multicolumn{1}{c|}{62.3} & \multicolumn{1}{c|}{0}    & \multicolumn{1}{c|}{48.1} & \multicolumn{1}{c|}{0}    & \multicolumn{1}{c|}{0}    \\ \hline
		\multicolumn{1}{|c|}{A4}  & \multicolumn{1}{c|}{68}    & \multicolumn{1}{c|}{68.3} & \multicolumn{1}{c|}{70.6} & \multicolumn{1}{c|}{0}    & \multicolumn{1}{c|}{70.4} & \multicolumn{1}{c|}{70.8} & \multicolumn{1}{c|}{64.3} & \multicolumn{1}{c|}{67.5} & \multicolumn{1}{c|}{0}    & \multicolumn{1}{c|}{27.5} & \multicolumn{1}{c|}{0}    & \multicolumn{1}{c|}{0}    \\ \hline
		\multicolumn{1}{|c|}{A5}  & \multicolumn{1}{c|}{0}     & \multicolumn{1}{c|}{0}    & \multicolumn{1}{c|}{67.4} & \multicolumn{1}{c|}{70.4} & \multicolumn{1}{c|}{0}    & \multicolumn{1}{c|}{70.6} & \multicolumn{1}{c|}{0}    & \multicolumn{1}{c|}{0}    & \multicolumn{1}{c|}{0}    & \multicolumn{1}{c|}{0}    & \multicolumn{1}{c|}{0}    & \multicolumn{1}{c|}{0}    \\ \hline
		\multicolumn{1}{|c|}{A6}  & \multicolumn{1}{c|}{0}     & \multicolumn{1}{c|}{0}    & \multicolumn{1}{c|}{69.9} & \multicolumn{1}{c|}{70.8} & \multicolumn{1}{c|}{70.6} & \multicolumn{1}{c|}{0}    & \multicolumn{1}{c|}{0}    & \multicolumn{1}{c|}{0}    & \multicolumn{1}{c|}{0}    & \multicolumn{1}{c|}{0}    & \multicolumn{1}{c|}{0}    & \multicolumn{1}{c|}{0}    \\ \hline
		\multicolumn{1}{|c|}{A7}  & \multicolumn{1}{c|}{0}     & \multicolumn{1}{c|}{0}    & \multicolumn{1}{c|}{63}   & \multicolumn{1}{c|}{64.3} & \multicolumn{1}{c|}{0}    & \multicolumn{1}{c|}{0}    & \multicolumn{1}{c|}{0}    & \multicolumn{1}{c|}{0}    & \multicolumn{1}{c|}{0}    & \multicolumn{1}{c|}{75.8} & \multicolumn{1}{c|}{0}    & \multicolumn{1}{c|}{0}    \\ \hline
		\multicolumn{1}{|c|}{A8}  & \multicolumn{1}{c|}{0}     & \multicolumn{1}{c|}{64.7} & \multicolumn{1}{c|}{62.3} & \multicolumn{1}{c|}{67.5} & \multicolumn{1}{c|}{0}    & \multicolumn{1}{c|}{0}    & \multicolumn{1}{c|}{0}    & \multicolumn{1}{c|}{0}    & \multicolumn{1}{c|}{51.1} & \multicolumn{1}{c|}{0}    & \multicolumn{1}{c|}{0}    & \multicolumn{1}{c|}{0}    \\ \hline
		\multicolumn{1}{|c|}{A9}  & \multicolumn{1}{c|}{0}     & \multicolumn{1}{c|}{0}    & \multicolumn{1}{c|}{0}    & \multicolumn{1}{c|}{0}    & \multicolumn{1}{c|}{0}    & \multicolumn{1}{c|}{0}    & \multicolumn{1}{c|}{0}    & \multicolumn{1}{c|}{51.1} & \multicolumn{1}{c|}{0}    & \multicolumn{1}{c|}{0}    & \multicolumn{1}{c|}{54.4} & \multicolumn{1}{c|}{0}    \\ \hline
		\multicolumn{1}{|c|}{A10} & \multicolumn{1}{c|}{0}     & \multicolumn{1}{c|}{0}    & \multicolumn{1}{c|}{48.1} & \multicolumn{1}{c|}{27.5} & \multicolumn{1}{c|}{0}    & \multicolumn{1}{c|}{0}    & \multicolumn{1}{c|}{75.8} & \multicolumn{1}{c|}{0}    & \multicolumn{1}{c|}{0}    & \multicolumn{1}{c|}{87.5} & \multicolumn{1}{c|}{0}    & \multicolumn{1}{c|}{61.9} \\ \hline
		\multicolumn{1}{|c|}{A11} & \multicolumn{1}{c|}{0}     & \multicolumn{1}{c|}{0}    & \multicolumn{1}{c|}{0}    & \multicolumn{1}{c|}{0}    & \multicolumn{1}{c|}{0}    & \multicolumn{1}{c|}{0}    & \multicolumn{1}{c|}{0}    & \multicolumn{1}{c|}{0}    & \multicolumn{1}{c|}{54.4} & \multicolumn{1}{c|}{0}    & \multicolumn{1}{c|}{0}    & \multicolumn{1}{c|}{0}    \\ \hline
		\multicolumn{1}{|c|}{A12} & \multicolumn{1}{c|}{0}     & \multicolumn{1}{c|}{0}    & \multicolumn{1}{c|}{0}    & \multicolumn{1}{c|}{0}    & \multicolumn{1}{c|}{0}    & \multicolumn{1}{c|}{0}    & \multicolumn{1}{c|}{0}    & \multicolumn{1}{c|}{0}    & \multicolumn{1}{c|}{0}    & \multicolumn{1}{c|}{61.9} & \multicolumn{1}{c|}{0}    & \multicolumn{1}{c|}{0}    \\ \hline
		\multicolumn{13}{l}{\begin{tabular}[c]{@{}l@{}} \scriptsize{* A1 to A12 indicate Annotators 1 to 12}\\ \scriptsize{** Zero means that there were no common images between the two annotators}\end{tabular}}                                                                                                                                                                                        
	\end{tabular}
	
\end{table}


\begin{figure}[H]
	\centering
	\includegraphics[width=\textwidth-1cm]{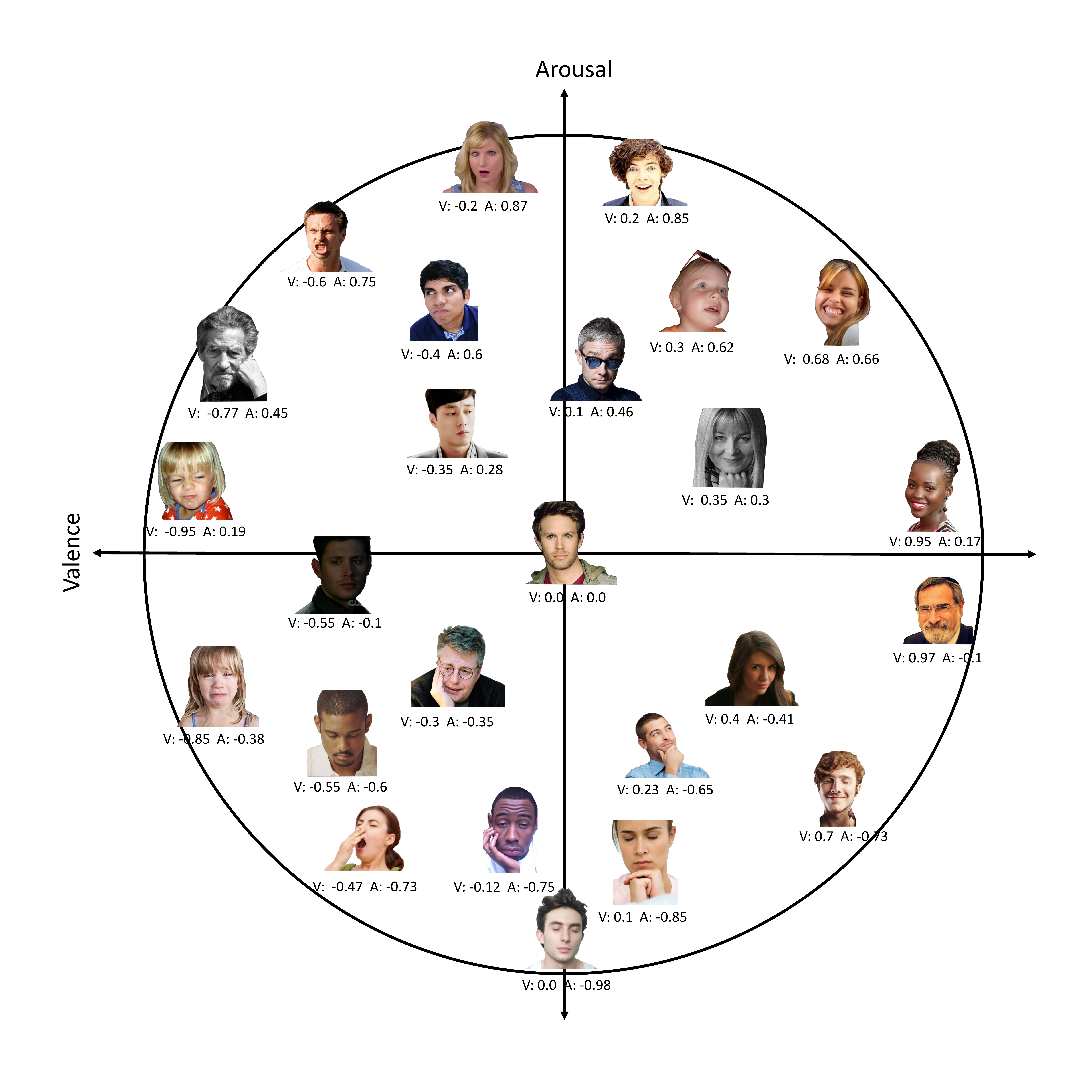}
	\vspace{-12mm}
	\caption{Sample images in Valence Arousal circumplex with their corresponding Valence and Arousal values (V: Valence, A: Arousal).}
	
	\label{fig:sampleImageCircumplex_appendix}
\end{figure}

\begin{table}[H]
	\centering
	
	\caption{Number of annotated images in each range/area of valence and arousal}
	\label{Tab:NumberOfValAro_appendix}
	\begin{tabular}{c|c|c|c|c|c|c|c|c|c|c|c|}
		\cline{3-12}
		\multicolumn{2}{c|}{\multirow{2}{*}{}}                          & \multicolumn{10}{c|}{Valence}                                                                                                                  \\ \cline{3-12} 
		\multicolumn{2}{c|}{}                                           & {[}-1,-.8{]} & {[}-.8,-.6{]} & {[}-.6,-.4{]} & {[}-.4,-.2{]} & {[}-.2,0{]} & {[}0,.2{]} & {[}.2,.4{]} & {[}.4,.6{]} & {[}.6,.8{]} & {[}.8,1{]} \\ \hline
		\multicolumn{1}{|c|}{\multirow{10}{*}{\begin{turn}{+90}Arousal\end{turn}}} & {[}.8,1{]}    & 0            & 0             & 21            & 674           & 1021        & 521        & 60          & 57          & 0           & 0          \\ \cline{2-12} 
		\multicolumn{1}{|c|}{}                          & {[}.6,.8{]}   & 0            & 74            & 161           & 561           & 706         & 1006       & 432         & 738         & 530         & 0          \\ \cline{2-12} 
		\multicolumn{1}{|c|}{}                          & {[}.4,.6{]}   & 638          & 720           & 312           & 505           & 2689        & 1905       & 1228        & 992         & 3891        & 957        \\ \cline{2-12} 
		\multicolumn{1}{|c|}{}                          & {[}.2,.4{]}   & 6770         & 9283          & 3884          & 2473          & 5530        & 2296       & 3506        & 1824        & 2667        & 1125       \\ \cline{2-12} 
		\multicolumn{1}{|c|}{}                          & {[}0,.2{]}    & 3331         & 1286          & 2971          & 4854          & 14083       & 15300      & 4104        & 9998        & 13842       & 9884       \\ \cline{2-12} 
		\multicolumn{1}{|c|}{}                          & {[}-.2,0{]}   & 395          & 577           & 5422          & 3675          & 9024        & 23201      & 6237        & 42219       & 23281       & 21040      \\ \cline{2-12} 
		\multicolumn{1}{|c|}{}                          & {[}-.4,-.2{]} & 787          & 1364          & 3700          & 6344          & 2804        & 1745       & 821         & 5241        & 10619       & 9934       \\ \cline{2-12} 
		\multicolumn{1}{|c|}{}                          & {[}-.6,-.4{]} & 610          & 7800          & 2645          & 3571          & 2042        & 2517       & 1993        & 467         & 1271        & 921        \\ \cline{2-12} 
		\multicolumn{1}{|c|}{}                          & {[}-.8,-.6{]} & 0            & 3537          & 8004          & 4374          & 5066        & 3379       & 4169        & 944         & 873         & 0          \\ \cline{2-12} 
		\multicolumn{1}{|c|}{}                          & {[}-1,-.8{]}  & 0            & 0             & 4123          & 1759          & 4836        & 1845       & 1672        & 739         & 0           & 0          \\ \hline
	\end{tabular}
	
\end{table}

\begin{table}[H]
\centering
\caption{Evaluation Metrics and Comparison of CNN baselines, SVM and MS Cognitive on Categorical Model of Affect on the Validation Set.}
\label{tab:baseLineMetricOnValidationSet_appendix}
\begin{tabular}{l|c|c|c|c|c|c|}
\cline{2-7}
\multirow{2}{*}{}                & \multicolumn{4}{c|}{CNN Baselines}                       & \multirow{2}{*}{SVM} & \multirow{2}{*}{MS Cognitive} \\ \cline{2-5}
                                 & Imbalanced & Down-Sampling & Up-Sampling & Weighted-Loss &                      &                               \\ \hline
\multicolumn{1}{|l|}{Accuracy}   & 0.40       & 0.50          & 0.47        & 0.58          & 0.30                 & 0.37                          \\ \hline
\multicolumn{1}{|l|}{F\_1-Score} & 0.34       & 0.49          & 0.44        & 0.58          & 0.24                 & 0.33                          \\ \hline
\multicolumn{1}{|l|}{Kappa}      & 0.32       & 0.42          & 0.38        & 0.51          & 0.18                 & 0.27                          \\ \hline
\multicolumn{1}{|l|}{Alpha}      & 0.39       & 0.42          & 0.37        & 0.51          & 0.13                 & 0.23                          \\ \hline
\multicolumn{1}{|l|}{AUCPR}      & 0.42       & 0.48          & 0.44        & 0.56          & 0.30                 & 0.38                          \\ \hline
\multicolumn{1}{|l|}{AUC}        & 0.74       & 0.47          & 0.75        & 0.82          & 0.68                 & 0.70                          \\ \hline
\end{tabular}
\end{table}

\begin{table}[H]
\centering
\caption{Baselines' Performances of Predicting Valence and Arousal on the Validation Set}
\label{tab:baseLinePerformanceValidationValenceArousal_appendix}
\begin{tabular}{l|cc|cc|}
\cline{2-5}
                           & \multicolumn{2}{c|}{CNN (AlexNet)}                         & \multicolumn{2}{c|}{SVR}                                   \\ \cline{2-5} 
                           & \multicolumn{1}{l}{Valence} & \multicolumn{1}{l|}{Arousal} & \multicolumn{1}{l}{Valence} & \multicolumn{1}{l|}{Arousal} \\ \hline
\multicolumn{1}{|l|}{RMSE} & 0.37                        & 0.41                         & 0.55                        & 0.42                         \\ \hline
\multicolumn{1}{|l|}{CORR} & 0.66                        & 0.54                         & 0.35                        & 0.31                         \\ \hline
\multicolumn{1}{|l|}{SAGR} & 0.74                        & 0.65                         & 0.57                        & 0.68                         \\ \hline
\multicolumn{1}{|l|}{CCC}  & 0.60                        & 0.34                         & 0.30                        & 0.18                         \\ \hline
\end{tabular}
\end{table}

\end{document}